\documentclass[onecolumn]{article}
\usepackage{colm2024_conference}
\usepackage{scrextend}
\usepackage{booktabs}
\usepackage{longtable}
\usepackage{graphicx}
\usepackage{enumitem}
\usepackage{algorithm}
\usepackage{algpseudocode}
\usepackage{natbib}
\usepackage{makecell}
\usepackage{array}
\usepackage{amsmath}
\usepackage{amssymb}
\usepackage{amsfonts}
\usepackage{multirow}
\usepackage{caption}
\usepackage{subcaption}
\usepackage{tabularx}
\usepackage{colortbl}
\usepackage{float}
\usepackage{placeins}
\usepackage{hyperref}
\usepackage[utf8]{inputenc}
\usepackage[T1]{fontenc}
\DeclareUnicodeCharacter{00B0}{\ensuremath{^\circ}}
\DeclareUnicodeCharacter{00B3}{\ensuremath{^3}}
\DeclareUnicodeCharacter{00C5}{\AA{}}
\DeclareUnicodeCharacter{03B1}{\ensuremath{\alpha}}
\DeclareUnicodeCharacter{03B2}{\ensuremath{\beta}}
\DeclareUnicodeCharacter{03B3}{\ensuremath{\gamma}}
\usepackage{xspace}
\usepackage{url}
\usepackage{nicefrac}
\usepackage{textcomp}
\usepackage{siunitx}
\usepackage{listings}
\usepackage{adjustbox}
\usepackage{framed}
\usepackage{CJKutf8}

\renewcommand{\arraystretch}{1.18}
\algrenewcommand\alglinenumber[1]{\scriptsize #1:}

\newcommand{\model}{S1-Omni\xspace}
\newcommand{\sonevl}{S1-VL-32B\xspace}
\newcommand{\sonecorpus}{S1-Omni-Corpus\xspace}

\definecolor{CaseBlue}{HTML}{2457A7}
\definecolor{CaseBorder}{HTML}{B8CDF5}
\definecolor{CaseFill}{HTML}{F3F7FD}
\definecolor{CaseMuted}{HTML}{52657A}
\definecolor{CaseFailure}{HTML}{9C3D32}
\definecolor{CaseFailureBorder}{HTML}{E5B7B0}
\definecolor{CaseFailureFill}{HTML}{FFF7F5}

\makeatletter
\newcommand{\CaseNeedspace}[1]{%
  \par
  \begingroup
  \dimen@=\pagegoal
  \advance\dimen@ by -\pagetotal
  \ifdim\dimen@<#1\relax
    \newpage
  \fi
  \endgroup
}
\makeatother

\newenvironment{CaseBlock}[1]{%
  \CaseNeedspace{8\baselineskip}%
  \def\FrameCommand##1{\fcolorbox{CaseBorder}{CaseFill}{##1}}%
  \MakeFramed{\advance\hsize-\width\FrameRestore}%
  \noindent\textcolor{CaseBlue}{\textbf{#1}}\par\smallskip
  \small
  \normalcolor
  \setlength{\parskip}{0.45em}%
  \widowpenalty=10000
  \clubpenalty=10000
  \displaywidowpenalty=10000
}{%
  \endMakeFramed
  \normalcolor
  \normalsize
}

\newenvironment{CaseBlockImmediate}[1]{%
  \def\FrameCommand##1{\fcolorbox{CaseBorder}{CaseFill}{##1}}%
  \MakeFramed{\advance\hsize-\width\FrameRestore}%
  \noindent\textcolor{CaseBlue}{\textbf{#1}}\par\smallskip
  \small
  \color{black}
  \setlength{\parskip}{0.45em}%
  \widowpenalty=10000
  \clubpenalty=10000
  \displaywidowpenalty=10000
}{%
  \endMakeFramed
  \color{black}
  \normalsize
}

\newenvironment{CaseBlockCompact}[1]{%
  \CaseNeedspace{8\baselineskip}%
  \def\FrameCommand##1{\fcolorbox{CaseBorder}{CaseFill}{##1}}%
  \MakeFramed{\advance\hsize-\width\FrameRestore}%
  \noindent\textcolor{CaseBlue}{\textbf{#1}}\par\smallskip
  \footnotesize
  \color{black}
  \setlength{\parskip}{0.25em}%
  \widowpenalty=10000
  \clubpenalty=10000
  \displaywidowpenalty=10000
}{%
  \endMakeFramed
  \color{black}
  \normalsize
}

\newcommand{\CaseProtocolMarker}[1]{%
  \CaseNeedspace{3\baselineskip}%
  \par\smallskip
  \noindent{\color{CaseBlue}\bfseries\ttfamily\detokenize{#1}}\par\smallskip
}

\newcommand{\CaseNativeOutputHeading}{%
  \CaseNeedspace{5\baselineskip}%
  \par\medskip
  \begingroup
  \noindent\colorbox{CaseFill}{%
    \parbox{\dimexpr\linewidth-2\fboxsep\relax}{%
      \textcolor{CaseBlue}{\bfseries Decoder Output}%
    }%
  }\par\smallskip
  \endgroup
  \color{black}
}

\newcommand{\CaseMiniHeader}[1]{%
  \begingroup
  \par\noindent\colorbox{CaseFill}{\parbox{\dimexpr\linewidth-2\fboxsep\relax}{\textcolor{CaseBlue}{\bfseries #1}}}\par\smallskip
  \endgroup
  \normalcolor
}

\newcommand{\RouteToken}[1]{\texttt{\detokenize{#1}}}

\newcommand{\CaseResultCard}[3]{%
  \begingroup
  \par\smallskip
  \noindent\fcolorbox{CaseBorder}{white}{%
    \parbox{\dimexpr\linewidth-2\fboxsep-2\fboxrule\relax}{%
      \centering
      {\scriptsize\textcolor{CaseMuted}{#1}\par}
      {\Large\bfseries\textcolor{CaseBlue}{#2}\par}
      {\scriptsize\textcolor{CaseMuted}{#3}\par}
    }%
  }%
  \par\smallskip
  \endgroup
  \color{black}
}

\newcommand{\CaseCompactNativeResult}[3]{%
  \par\smallskip
  \begingroup
  \noindent\colorbox{CaseFill}{%
    \parbox{\dimexpr\linewidth-2\fboxsep\relax}{%
      \textcolor{CaseBlue}{\bfseries Native Output}%
    }%
  }\par\smallskip
  \noindent\fcolorbox{CaseBorder}{white}{%
    \parbox{\dimexpr\linewidth-2\fboxsep-2\fboxrule\relax}{%
      \centering
      {\scriptsize\textcolor{CaseMuted}{#1}\quad}%
      {\normalsize\bfseries\textcolor{CaseBlue}{#2}\quad}%
      {\scriptsize\textcolor{CaseMuted}{#3}}%
    }%
  }%
  \par\smallskip
  \endgroup
  \color{black}
}

\newcommand{\CaseResidues}[1]{%
  \par\noindent\textcolor{CaseBlue}{\textbf{#1}}\par
}

\lstdefinestyle{caseinput}{
  basicstyle=\ttfamily\scriptsize,
  frame=none,
  backgroundcolor=\color{CaseFill},
  rulecolor=\color{CaseBorder},
  showstringspaces=false,
  breaklines=true,
  breakatwhitespace=false,
  columns=fullflexible,
  keepspaces=true,
  xleftmargin=0.5em,
  xrightmargin=0.5em,
  aboveskip=0.4em,
  belowskip=0.4em
}

\lstdefinestyle{caseinputcompact}{
  style=caseinput,
  basicstyle=\ttfamily\tiny,
  aboveskip=0.2em,
  belowskip=0.2em
}

\newcommand{\CaseInputFile}[1]{\lstinputlisting[style=caseinput]{#1}}
\newcommand{\CaseInputFileCompact}[1]{\lstinputlisting[style=caseinputcompact]{#1}}

\newcommand{\CaseImagePair}[4]{%
  \begin{figure}[H]
    \centering
    \begin{minipage}[t]{0.47\textwidth}
      \centering
      \CaseMiniHeader{#1}
      \includegraphics[width=\linewidth]{#2}
    \end{minipage}\hfill
    \begin{minipage}[t]{0.47\textwidth}
      \centering
      \CaseMiniHeader{#3}
      \includegraphics[width=\linewidth]{#4}
    \end{minipage}
  \end{figure}
}

\newenvironment{CaseFailureNote}[1]{%
  \CaseNeedspace{4\baselineskip}%
  \def\FrameCommand##1{\fcolorbox{CaseFailureBorder}{CaseFailureFill}{##1}}%
  \MakeFramed{\advance\hsize-\width\FrameRestore}%
  \noindent\textcolor{CaseFailure}{\textbf{#1}}\par\smallskip
  \small\normalcolor
}{%
  \endMakeFramed
  \normalcolor
  \normalsize
}

\newcommand{\CaseSpeaker}[1]{%
  \CaseNeedspace{3\baselineskip}%
  \par\medskip\noindent\textcolor{CaseBlue}{\bfseries #1}\par\smallskip
}

\title{S1-Omni: A Unified Multimodal Reasoning Model for Scientific Understanding, Prediction, and Generation}

\author{
\bf ScienceOne AI\\
Wenge AI}
\begin{document}
\begin{CJK*}{UTF8}{gbsn}

\maketitle

\begin{abstract}
We present S1-Omni, a unified multimodal reasoning model for scientific understanding, prediction, and generation.
AI for Science (AI4S) has advanced significantly through domain-specific models, tool-augmented LLMs, and scientific language models. However, model capabilities remain highly fragmented, limiting the joint modeling of heterogeneous data, scientific laws, and expert knowledge.
S1-Omni addresses this gap by consolidating these capabilities into a single, coherent scientific reasoning model.
The architecture of S1-Omni is built upon three core components: unified representation of scientific data, natural-world knowledge alignment, and decoding for domain-specific tasks.
First, S1-Omni maps natural-language instructions and scientific objects, including CIF, SMILES, protein sequences, spectra, and scientific images, into a shared representation space.
Second, it incorporates scientific laws and expert knowledge into data construction and training, enabling the model to reason from scientific evidence.
Third, it performs task-specific decoding to support a broad range of applications, including property prediction, spectrum-to-molecular generation, protein site and structure prediction, and scientific image generation and editing.
S1-Omni is trained on S1-Omni-Corpus, which covers 200 scientific tasks and contains millions of reasoning samples, and is evaluated on over 60 scientific benchmarks.
It outperforms GPT-5.5 and Gemini-3.1-Pro on most benchmarks and matches or surpasses domain-specific models on several benchmarks.
Overall, S1-Omni provides a practical path toward unified scientific modeling.
\end{abstract}

\section{Introduction}
\label{sec:intro}

In recent years, AI for Science (AI4S) has advanced along three complementary directions. The first comprises domain-specific models, represented by AlphaFold~3\citep{abramson2024alphafold3} and ESM3~\citep{hayes2025simulating}, which develop high-performance capabilities for structure prediction, property prediction, or generation around specific scientific entities such as proteins, molecules, and materials. The second comprises tool-augmented general-purpose models, represented by GPT-Rosalind~\citep{openai2026rosalind} and Claude Science~\citep{anthropic2026claudescience}, which extend general language models with agents, scientific tools, and automated workflows for operating on and executing scientific tasks. The third comprises scientific language models, represented by NatureLM~\citep{xia2025naturelm}, LOGOS~\citep{li2026speaking}, and BioMatrix~\citep{pei2026biomatrix}, which organize scientific entities such as proteins, DNA, and molecules as jointly modeled symbolic sequences or representation spaces and explore unified modeling across scientific objects. Together, these advances have driven domain modeling, scientific workflow automation, and representation learning for scientific objects, providing important foundations for unified multimodal reasoning in science.

However, existing scientific intelligence remains fragmented across domain-specific models, tool-augmented LLMs, and scientific language models. Domain-specific models are typically tied to particular scientific entities and task types. Tool-augmented LLMs use external tools to complete scientific tasks. Scientific language models attempt to unify scientific knowledge within a shared embedding space. However, these approaches have yet to bring knowledge from different disciplines, scientific principles, and constraints together within a unified model. The key to unified scientific intelligence is therefore not merely to provide a natural-language interface or to convert different scientific objects into a single token representation. It requires organizing heterogeneous scientific data within the same model process, injecting scientific laws, expert heuristics, and domain-task constraints into task understanding and model reasoning through implicit representations, and producing outputs that can be verified in their corresponding native scientific forms. This work investigates how these distributed capabilities can be integrated into a unified multimodal reasoning model, allowing a shared backbone to develop cross-disciplinary, cross-modal, and cross-task scientific understanding and reasoning.

To this end, we introduce \model, a unified multimodal reasoning model for scientific understanding, prediction, and generation. \model is built around three core capabilities. First, \textbf{unified representation of scientific data.} The model organizes natural-language instructions and diverse scientific objects into a unified task representation, covering text, material CIFs, chemical SMILES, protein sequences, spectra, scientific images, and other scientific modalities. At the same time, it preserves object-type boundaries, representation structures, and necessary encoding paths rather than equating unified modeling with uniform tokenization. Second, \textbf{natural-world knowledge alignment.} Model training does not rely only on statistical associations between inputs and outputs, but also incorporates scientific laws and expert knowledge, together with experimental facts, into data construction, sample validation, and training, enabling the model to form intermediate judgments from scientific evidence in the current task and thereby improving scientific reasoning and interpretability. Third, \textbf{decoding for domain-specific tasks.} On top of shared task understanding and scientific reasoning, the model connects the appropriate output representation and decoders according to the task objective. It supports property prediction, spectrum-to-molecular generation, protein structure generation, and scientific image generation and editing, allowing a unified model to produce outputs in each domain's native result space.

The capabilities of \model are built on \sonecorpus, which covers 200 scientific tasks and contains millions of high-quality scientific reasoning samples. During dataset construction, we use scientific laws and expert knowledge to design task-specific reasoning chains, taking into account relevant reasoning dimensions, evidence sources, and output constraints. In this way, \model learns not only surface patterns in scientific data, but also how to reason from evidence in a specific task, providing an effective initial path toward a scalable unified scientific modeling paradigm.

Our evaluation covers multiple scientific modalities, including molecules, materials, proteins, spectra, and scientific images, and supports tasks such as property prediction, spectrum-to-molecular generation, protein functional-site prediction, protein structure prediction, and scientific image generation and editing. Systematic comparisons on over 60 benchmarks show that \model outperforms GPT-5.5 and Gemini-3.1-Pro on most evaluation metrics, and matches or surpasses task-specific models on several specialized benchmarks. Specifically, it outperforms both GPT-5.5 and Gemini-3.1-Pro on 16 of 18 ADMET evaluations; matches or exceeds representative domain-specific models on several important drug-property tasks involving CYP enzymes, hERG, and intestinal absorption; supports multiple residue-level functional-site prediction tasks within a unified framework and achieves leading performance on protein--protein interaction, epitope, and small-molecule binding-site prediction; and demonstrates strong performance in scientific image generation and editing, including MSD medical image segmentation, medical image translation, super-resolution, and scientific illustration generation. 

These results show that \model performs scientific understanding and reasoning across modalities and tasks in a single unified process, yielding outputs that are natively verifiable in their respective result spaces. To support reproducible research and further development, we release the \model weights, modeling and inference-serving code under the Apache-2.0 license, together with the high-quality S1-Omni-Corpus-10K subset.

\section{Related Work}
\label{sec:related}

\textbf{Domain-specific scientific models.}
Domain-specific scientific models are built around particular scientific objects, domain inductive biases, and evaluation protocols, and have achieved strong performance on their target tasks. AlphaFold~3~\citep{abramson2024alphafold3} predicts structures of biomolecular complexes containing proteins, nucleic acids, small molecules, ions, and modified residues. ESM3~\citep{hayes2025simulating} unifies generation across protein sequence, structure, and function, while ODesign~\citep{odesign2025} further targets biomolecular interaction design. Similar specialization has emerged in other scientific domains: Materials Property Axiom~\citep{deeprinciple2026mpa} focuses on heterogeneous experimental materials-property prediction, DiffSpectra~\citep{wang2025diffspectra} reconstructs molecular structures from spectra, and S1-Omni-Image~\citep{li2026s1omniimage} supports scientific image understanding, generation, and editing. These models effectively exploit domain knowledge and scientific constraints, leading to high scientific validity within specific object types and task settings. However, their capabilities are usually tied to particular objects, data formats, and evaluation protocols, making it difficult to extend them naturally to other scientific fields or tasks. This suggests that domain-specific models remain an important component of scientific intelligence, while also raising the question of how diverse scientific objects and domain capabilities can be connected within a unified model.

\textbf{Tool-augmented general models.}
General-purpose language models and open-domain multimodal models establish natural language as a unified interface for perception, generation, and tool use. From GPT-4~\citep{openai2023gpt4}, Gemini~\citep{team2023gemini}, Claude~3~\citep{anthropic2024claude3}, and ChatGLM~\citep{glm2024chatglm} to multimodal systems such as LLaVA~\citep{liu2023visual}, NExT-GPT~\citep{wu2023nextgpt}, GPT-4o~\citep{openai2024gpt4o}, Qwen2.5-Omni~\citep{xu2025qwen25omni}, SEED-X~\citep{ge2024seed}, and Janus-Pro~\citep{chen2025janus}, recent work has continuously expanded cross-modal understanding, generation, and task organization. In scientific settings, systems such as GPT-Rosalind~\citep{openai2026rosalind} and Claude Science~\citep{anthropic2026claudescience} further combine agents, scientific tools, and workflows to support literature analysis, data processing, and experimental collaboration. This line of work substantially improves the ability of general models to organize scientific tasks and invoke specialized capabilities. Nevertheless, scientific knowledge and execution still largely reside in external tools and workflow logic, rather than being internalized as scientific competence within a unified model.

\textbf{Scientific language models.}
Scientific language models attempt to map diverse scientific objects into unified representation spaces that can be jointly modeled, thereby improving interoperability across heterogeneous scientific data. NatureLM~\citep{xia2025naturelm} represents small molecules, proteins, RNA, and materials as sequences in a language of nature. LOGOS~\citep{li2026speaking} uses a scientific grammar to describe scientific objects, spatial constraints, and interactions. BioMatrix~\citep{pei2026biomatrix} jointly models biological sequences, structures, and natural language in a shared token space, while LucaOne~\citep{he2025generalized} studies unified sequence modeling across nucleic acids and proteins. Omni-Weather~\citep{zhou2025omni} unifies understanding and generation for weather radar data. These works show that shared tokens or scientific grammars can provide a common interface for heterogeneous scientific objects and improve cross-modal modeling. However, unifying object representations does not by itself amount to unifying scientific intelligence. How scientific laws, expert knowledge, and domain constraints should be incorporated into model training, and how they can further support consistent reasoning and result generation across scientific tasks, remain open problems.

\textbf{Scientific reasoning and knowledge alignment.}
Recent scientific reasoning models further explore how scientific laws and domain knowledge can be explicitly incorporated into model reasoning. Proteo-R1~\citep{wu2026proteo} introduces reasoning mechanisms into de novo protein design. BioReason-Pro~\citep{fallahpour2026bioreason} combines protein representations, multimodal biological context, and explicit reasoning for protein function prediction. DrugTrail~\citep{liudrugtrail} couples structured reasoning traces with druggability-oriented preference optimization. In chemistry, ether0~\citep{narayanan2026training}, MPPReasoner~\citep{zhuang2025reasoning}, and Chem-R~\citep{wang2025chem} explore reinforcement learning, molecular-property reasoning traces, and expert-like reasoning supervision, respectively. PiFlow~\citep{pu2025piflow} further uses scientific principles to guide exploration, validation, and iterative refinement. These studies demonstrate that incorporating scientific laws into model training can improve scientific reasoning. However, most existing methods remain focused on particular domains or individual tasks, and have not yet formed a foundation model that can uniformly connect heterogeneous scientific data, scientific laws, and domain tasks.

Different from the above work, \model explores a unified multimodal reasoning architecture for scientific understanding, prediction, and generation, centered on three core capabilities: unified representation of scientific data, natural-world knowledge alignment, and decoding for domain-specific tasks. Specifically, \model first encodes heterogeneous scientific data---including text, molecules, materials, proteins, spectra, and scientific images---within a unified modeling framework. It then performs natural-world knowledge alignment using scientific laws and expert knowledge, so that task decisions are constrained by scientific evidence. Finally, it performs decoding for domain-specific tasks, enabling unified scientific understanding, prediction, and generation, rather than relying on a composition of multiple domain-specific models, external tools, or scientific language models.

\section{Method}
\label{sec:method}

\model connects unified representation of scientific data, natural-world knowledge alignment, and decoding for domain-specific tasks within a single model process. Given a user instruction and scientific objects, a shared vision-language model first forms a task-conditioned hidden representation and generates optional scientific reasoning, a textual answer, and a task token. For tasks whose outputs can be expressed directly in natural language, the textual answer is the final result. For tasks with specialized output spaces---including properties, protein sites, molecular structures, 3D coordinates, and scientific images---the task token selects a result decoder that converts the shared representation into an output verifiable in its native scientific form.

Let the training set be
\begin{equation}
\mathcal{D}=\{d_i\}_{i=1}^{N},
\qquad
d_i=(x_i,y_i,t_i,m_i),
\end{equation}
where
\begin{equation}
x_i=(x_i^{\mathrm{text}},x_i^{\mathrm{task}}),
\qquad
y_i=(y_i^{\mathrm{text}},y_i^{\mathrm{task}}),
\qquad
y_i^{\mathrm{text}}=(s_i,r_i,a_i).
\end{equation}
Here, $x_i^{\mathrm{text}}$ contains the user instruction and any necessary textual context, and $x_i^{\mathrm{task}}$ contains one or more native scientific objects. The variables $s_i$, $r_i$, and $a_i$ denote an optional task token, optional scientific reasoning, and a textual answer, while $y_i^{\mathrm{task}}$ is the domain-native supervision target. The task type is $t_i\in\mathcal{T}\cup\{\varnothing\}$, and $m_i$ stores metadata such as provenance, data split, and traceable identifiers. Text tasks satisfy $t_i=s_i=\varnothing$ and $y_i^{\mathrm{task}}=\varnothing$. Domain tasks have $t_i\in\mathcal{T}$, with $s_i$ selecting the corresponding task decoder.

\subsection{Model Architecture}
\label{sec:model}

\begin{figure}[t!]
    \centering
    \includegraphics[width=\textwidth]{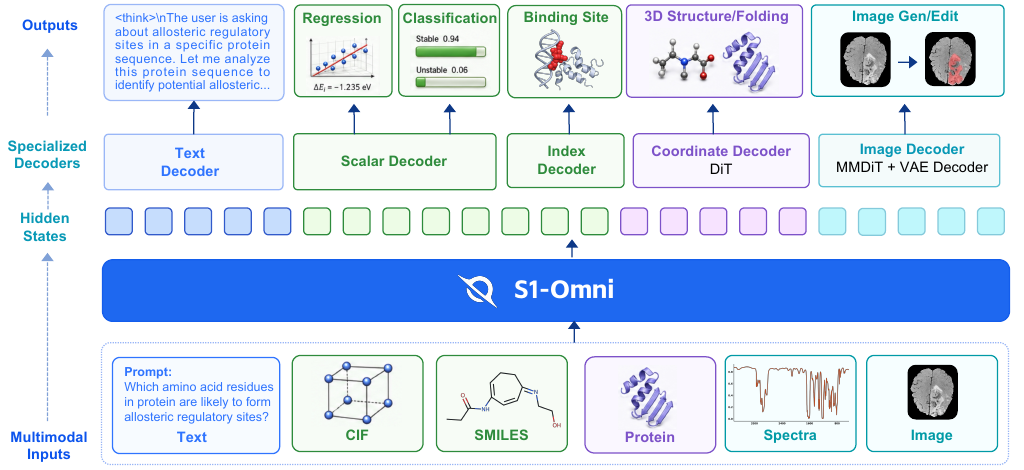}
    \caption{Unified architecture of \model. Text, CIF, SMILES, proteins, spectra, and scientific images are first processed by the shared vision-language model. Its hidden representations are then converted by text, scalar, position-index, coordinate, and image decoders into natural-language responses, property predictions, site predictions, three-dimensional structures, and scientific image generation or editing results.}
    \label{fig:s1_omni_architecture}
\end{figure}

As illustrated in Figure~\ref{fig:s1_omni_architecture}, \model uses \sonevl~\citep{li2026s1} as its shared backbone and task result decoders as domain output interfaces. The backbone performs cross-modal task understanding and scientific reasoning, while the result decoders preserve the numerical, discrete, structural, geometric, and visual constraints of their respective output spaces. The two components communicate through VLM prefill hidden representations. This design neither compresses every scientific object into a single raw representation nor requires all tasks to share one output head.

\subsubsection{Unified Representation of Scientific Data and Shared Task Representations}
\label{sec:unified_encoding}

\model organizes text instructions and heterogeneous scientific objects within a common task context. Serializable objects, including SMILES strings, molecular formulae, FASTA sequences, CIF files, Wyckoff representations, and structure files, enter through the text channel. Scientific images, medical images, and visualized spectra enter through the visual channel. Input construction preserves object boundaries and modality roles, while the instruction specifies whether each object serves as a prediction condition, explanatory evidence, generation target, or editing condition.

The VLM task context is
\begin{equation}
c_i=\operatorname{Fuse}
\left(x_i^{\mathrm{text}},x_i^{\mathrm{task}}\right).
\end{equation}
The operator $\operatorname{Fuse}$ comprises the serialization, visual encoding, and context assembly required by different scientific objects. High-resolution residue, spectral-peak, pixel, or geometric information remains available in $x_i^{\mathrm{task}}$ and can be read directly by a downstream task decoder.

The shared VLM encodes the task context during prefill and generates a task response from the resulting hidden representation:
\begin{equation}
\hat{h}_i=F_{\theta}^{\mathrm{prefill}}(c_i),
\qquad
(\hat{s}_i,\hat{r}_i,\hat{a}_i)
=F_{\theta}^{\mathrm{AR}}(\hat{h}_i).
\end{equation}
Natural-world knowledge alignment shapes this representation through evidence selection, scientific relations, and reasoning supervision in the training data. Beyond identifying the input objects and task objective, the model must organize task-relevant intermediate judgments in $\hat{r}_i$, encouraging $\hat{h}_i$ to jointly encode user intent, scientific evidence, and their relations. This representation supports both textual responses and domain-native result decoding.

When $\hat{s}_i\in\mathcal{T}$, the task-specific condition is extracted from the prefill representation as
\begin{equation}
\hat{e}_i
=\operatorname{Extract}_{\hat{s}_i}(\hat{h}_i).
\end{equation}
Depending on the decoder, $\operatorname{Extract}_t$ pools or selects token-level or global states, or projects the same hidden representation into a task-specific conditioning space. The task token selects only the task type and extraction rule; the domain prediction itself uses the shared condition $\hat{e}_i$.

\subsubsection{Decoding for Domain-Specific Tasks}
\label{sec:directed_decoding}

For a task requiring a domain-specific result, the selected decoder reads the shared task condition, when needed, the original scientific input:
\begin{equation}
\hat{y}_i^{\mathrm{task}}
=
G_{\phi_{\hat{s}_i}}
\left(\hat{e}_i,x_i^{\mathrm{task}}\right).
\end{equation}
The system output comprises a textual response and an optional domain result:
\begin{equation}
\hat{o}_i=
\begin{cases}
(\hat{r}_i,\hat{a}_i),
& \hat{s}_i=\varnothing,\\[4pt]
(\hat{r}_i,\hat{a}_i,\hat{y}_i^{\mathrm{task}}),
& \hat{s}_i\in\mathcal{T}.
\end{cases}
\label{eq:directed_decoding}
\end{equation}
The task token is an internal control signal and is not part of the user-visible output. When explicit reasoning is unnecessary, $\hat{r}_i=\varnothing$. Table~\ref{tab:task_decoders} summarizes how each task family extracts its condition from the shared representation and incorporates any auxiliary scientific input.

\begin{table}[t!]
\centering
\caption{Task tokens, prefill feature extraction, auxiliary conditions, decoders for domain-specific tasks, and domain-native results in \model.}
\label{tab:task_decoders}
\scriptsize
\setlength{\tabcolsep}{2.5pt}
\renewcommand{\arraystretch}{1.12}
\begin{tabularx}{\textwidth}{
  >{\raggedright\arraybackslash}p{0.15\textwidth}
  >{\raggedright\arraybackslash}p{0.13\textwidth}
  >{\raggedright\arraybackslash}p{0.08\textwidth}
  >{\raggedright\arraybackslash}p{0.16\textwidth}
  >{\raggedright\arraybackslash}p{0.20\textwidth}
  >{\raggedright\arraybackslash}X}
\toprule
Task family &
Task token &
$\operatorname{Extract}_t$ &
Auxiliary condition &
Decoder for domain-specific task &
Domain-native result \\
\midrule
Property regression &
\texttt{<linear\_pre>} &
pooled &
none &
shared linear readout $G_{\mathrm{linear}}$ &
continuous value \\
Property classification &
\texttt{<linear\_cla>} &
pooled &
none &
shared linear readout $G_{\mathrm{linear}}$ &
class score or probability \\
Protein-site prediction &
\texttt{<prot\_cla>} &
token-level &
ESM2 residue features &
residue-level decoder $G_{\mathrm{site}}$ &
residue probabilities \\
Spectrum-to-molecular generation &
\texttt{<spectra\_st>} &
projected &
SpecFormer spectral features &
molecular diffusion decoder $G_{\mathrm{mol}}$ &
molecular graph or 3D conformation \\
Protein structure prediction &
\texttt{<prot\_st>} &
global &
protein sequence representation &
geometric decoder $G_{\mathrm{fold}}$ &
three-dimensional coordinates \\
Scientific image generation and editing &
\texttt{<image\_gen>} / \texttt{<image\_edit>} &
token-level &
input image for editing &
MMDiT decoder $G_{\mathrm{img}}$ &
generated or edited image \\
\bottomrule
\end{tabularx}
\end{table}

In Table~\ref{tab:task_decoders}, pooled, token-level, projected, and global denote concrete implementations of $\operatorname{Extract}_t$. Property tasks extract a pooled prefill condition and use a shared linear network for regression or classification. Protein-site prediction retains token-level conditions and fuses them with ESM2 residue features. Spectrum-to-molecular generation projects the shared representation into a molecular generation space while reading SpecFormer spectral features. Protein structure prediction uses a global task condition to drive a geometric decoder. Scientific image generation and editing preserve token-level conditions and map them into an MMDiT conditioning space, with editing additionally reading the input image. The shared backbone therefore provides a unified state for task understanding and scientific reasoning, while each decoder performs prediction or generation in the corresponding native result space.

\subsection{Training Data}
\label{sec:training_data}

\subsubsection{Corpus Scope and Unified Records}
\label{sec:corpus_scope}

\begin{figure}[t]
    \centering
    \includegraphics[width=0.88\textwidth]{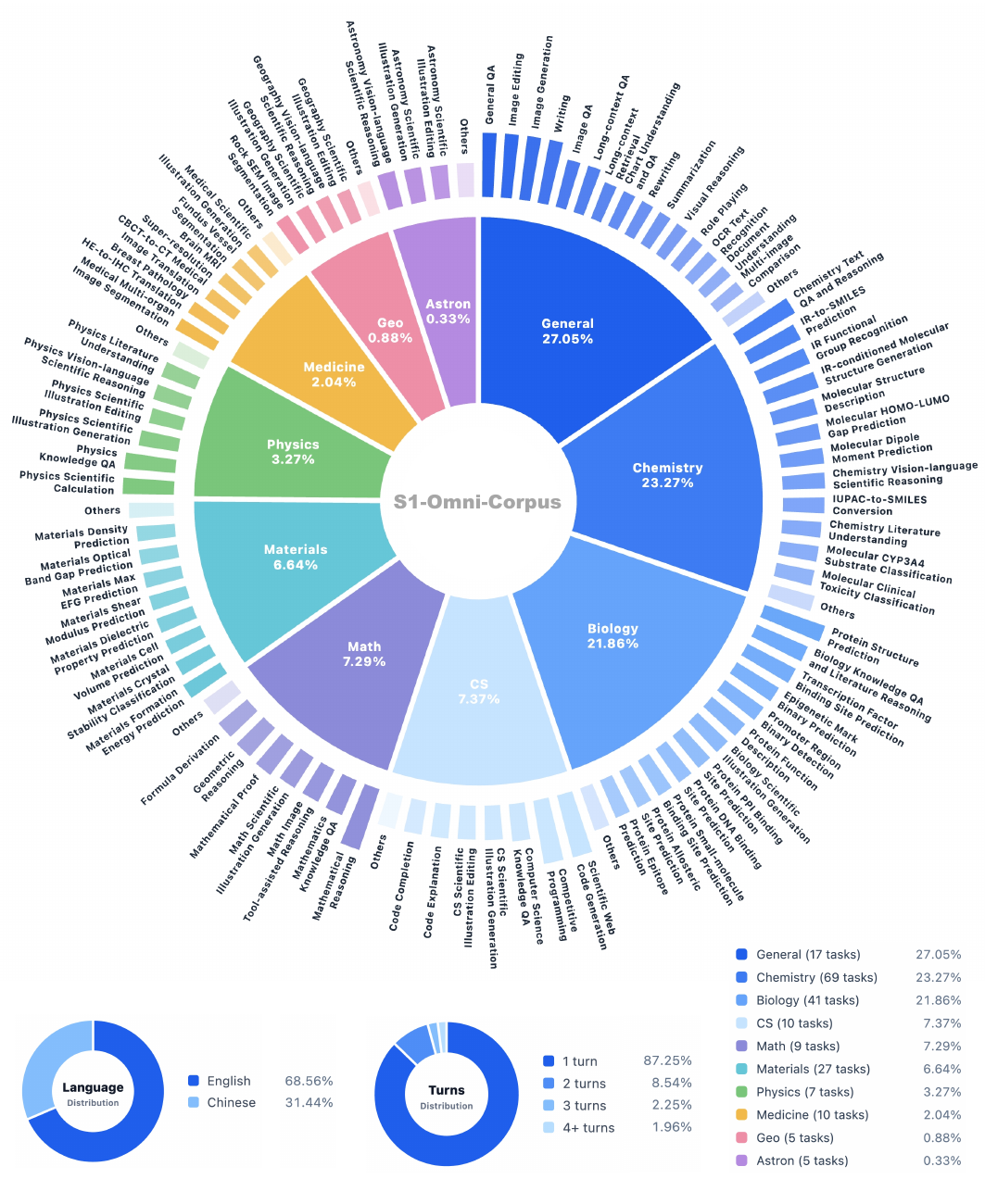}
    \caption{Discipline, task, language, and dialogue-turn distributions in \sonecorpus. These statistics characterize the coverage of the training data rather than define model capability directly by task count.}
    \label{fig:s1_omni_corpus}
\end{figure}

We construct \sonecorpus with partner research organizations and domain experts. The corpus contains more than 8 million filtered scientific reasoning and task supervision records spanning over 200 research tasks. It covers general text and multimodal tasks together with scientific domains including chemistry, biology, materials, physics, and medical imaging. Supported tasks include property prediction, spectrum-to-molecular generation, protein functional-site prediction, three-dimensional structure prediction, and scientific image generation and editing. Figure~\ref{fig:s1_omni_corpus} summarizes its discipline, task, language, and dialogue-turn distributions.

Every record follows $d_i=(x_i,y_i,t_i,m_i)$. The field $x_i^{\mathrm{text}}$ stores the user instruction and context, while $x_i^{\mathrm{task}}$ stores domain-native scientific objects. The textual target $y_i^{\mathrm{text}}$ provides supervision for optional reasoning, the answer, and the task token, whereas $y_i^{\mathrm{task}}$ contains a continuous value, class label, residue annotation, molecular structure, three-dimensional coordinates, or target image. Domain-native targets are never written into visible reasoning and are supplied to the corresponding result decoder through an independent supervision path. The metadata $m_i$ retains provenance, language, units, data split, augmentation history, and traceable identifiers. S1-Omni-Corpus-10K uses the same record format as the full corpus and supports data inspection, pipeline analysis, and community research.

\subsubsection{Data Construction for Natural-World Knowledge Alignment}
\label{sec:natural_world_alignment}

\begin{figure}[t]
    \centering
    \includegraphics[width=\textwidth]{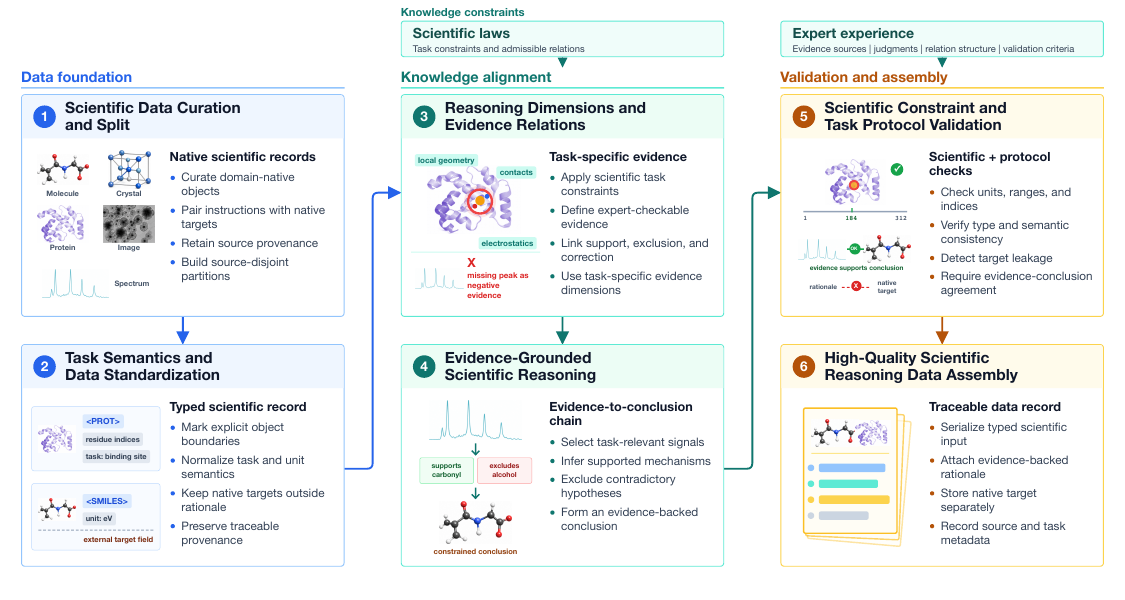}
    \caption{Data construction for natural-world knowledge alignment. Scientific laws and expert knowledge jointly constrain evidence dimensions, reasoning relations, task protocols, and sample validation.}
    \label{fig:s1_omni_data_pipeline}
\end{figure}

The data pipeline converts domain-native records into scientific task instances that are evidence-supported, protocol-consistent, and traceable. As shown in Figure~\ref{fig:s1_omni_data_pipeline}, it comprises three stages: data foundation, knowledge alignment, and validation and assembly.

\textbf{Data foundation.}
We first curate domain-native records for molecules, crystals, proteins, spectra, and scientific images. Deduplication, benchmark test-set overlap filtering, and train/validation/test partitioning are completed before data augmentation or reasoning distillation to prevent data leakage, ensuring that benchmark test instances or their near-duplicates do not enter the training corpus. Each raw record is then converted into a task instance $d_i$ with standardized task semantics, units, indices, and output requirements, while preserving native representations such as SMILES, CIF, protein sequences, three-dimensional structures, raw spectra, and images. The target $y_i^{\mathrm{task}}$ is stored separately from the visible context to prevent result information from entering the reasoning input.

\textbf{Natural-world knowledge alignment.}
Scientific laws specify the admissible evidence relations for a task, while domain experts determine the signals, decision process, and validation criteria that should be examined. Materials and chemistry tasks organize evidence around composition, structural fragments, local environments, and structure--property relations. Spectroscopy tasks jointly consider observed peaks, missing peaks, and cross-spectrum consistency. Protein tasks focus on sequence positions, local geometry, and residue contacts. Scientific image tasks examine entities, spatial relations, target regions, and editing boundaries. A candidate rationale $r_i$ selects verifiable information from $x_i$ and forms intermediate judgments through support, exclusion, and correction relations. If the available evidence cannot support a reliable explanation, we set $r_i=\varnothing$. Table~\ref{tab:evidence_alignment} summarizes representative evidence dimensions and the corresponding validation logic across scientific domains.

\textbf{Validation and assembly.}
Scientific consistency checks verify that the input evidence, reasoning process, textual answer, and domain target are mutually compatible. Task protocol checks cover object types, units, indices, media references, task tokens, output formats, and target leakage. The native target $y_i^{\mathrm{task}}$ is used only for supervision and consistency validation and is never introduced into $r_i$ as hidden evidence. Samples that pass validation retain the complete $x_i$, $y_i$, $t_i$, and $m_i$, and undergo task balancing, language expansion, and necessary data augmentation within the established data partitions.

\begin{table}[t]
\centering
\caption{Representative scientific evidence and validation logic used for natural-world knowledge alignment.}
\label{tab:evidence_alignment}
\footnotesize
\setlength{\tabcolsep}{5pt}
\renewcommand{\arraystretch}{1}
\begin{tabularx}{\textwidth}{
  >{\raggedright\arraybackslash}p{0.14\textwidth}
  >{\raggedright\arraybackslash}p{0.34\textwidth}
  >{\raggedright\arraybackslash}X}
\toprule
Domain &
Representative evidence &
Validation logic \\
\midrule
\textbf{Materials} &
Composition and stoichiometry; symmetry and dimensionality; local coordination and bonding &
Test whether these cues support the claimed property; reject stoichiometric, symmetry, or structural violations. \\
\addlinespace[2pt]
\textbf{Molecules} &
Functional groups and stereochemistry; charge distribution and electronic structure; intermolecular interactions &
Connect structural cues to the predicted property; reject valence, functional-group, or geometric inconsistencies. \\
\addlinespace[2pt]
\textbf{Drug discovery} &
Target interactions and pocket complementarity; structure--activity relations; ADMET mechanisms &
Verify that the proposed mechanism agrees with the endpoint definition, label semantics, and relevant molecular interactions. \\
\addlinespace[2pt]
\textbf{Proteins} &
Sequence conservation and motifs; residue environments; local geometry and three-dimensional contacts &
Check whether site, interface, function, or structure assignments agree with residue indices and local contacts. \\
\addlinespace[2pt]
\textbf{Spectra} &
Observed and missing peaks; functional-group assignments; cross-spectrum consistency &
Require candidates to explain the major signals; use missing peaks to reject conflicting structures. \\
\addlinespace[2pt]
\textbf{Images} &
Entities and target regions; spatial and topological relations; editing boundaries &
Preserve entities and non-target content, maintain spatial relations, and restrict modifications to the requested region. \\
\bottomrule
\end{tabularx}
\end{table}

\subsection{Training}
\label{sec:training}

\model uses two-stage training. The first stage trains the shared VLM under the unified task protocol to learn task understanding, scientific reasoning, textual answering, and task-token generation. The second stage freezes the shared VLM and trains each result decoder in its domain-native supervision space.

\subsubsection{Shared VLM Training}
\label{sec:vlm_training}

The shared VLM takes $x_i$ as input and uses $y_i^{\mathrm{text}}=(s_i,r_i,a_i)$ as response-side supervision. We serialize its nonempty fields in their response order and optimize the standard autoregressive cross-entropy objective:
\begin{equation}
\mathcal{L}_{\mathrm{VLM}}(\theta)
=-
\frac{1}{\sum_i |y_i^{\mathrm{text}}|}
\sum_i
\sum_{k=1}^{|y_i^{\mathrm{text}}|}
\log p_{\theta}\!\left(
y_{i,k}^{\mathrm{text}}
\mid
c_i,y_{i,<k}^{\mathrm{text}}
\right).
\end{equation}
The domain-native target $y_i^{\mathrm{task}}$ is not serialized into the VLM response and is reserved for task-decoder training in the second stage.

\subsubsection{Task Result Decoder Training}
\label{sec:module_training}

After the first stage, the shared VLM parameters are fixed at $\theta^\star$. For an instance with $t_i\in\mathcal{T}$, the task condition is extracted from the frozen model's prefill representation:
\begin{equation}
e_i=
\operatorname{Extract}_{t_i}
\left(F_{\theta^\star}^{\mathrm{prefill}}(c_i)\right).
\end{equation}
Each task decoder reads $e_i$ and any required native scientific input $x_i^{\mathrm{task}}$, and is supervised in its own result space:
\begin{align}
\mathcal{L}_{\mathrm{task}}
\left(\{\phi_t\}_{t\in\mathcal{T}}\right)
&=
\mathbb{E}_{\substack{d_i\sim\mathcal{D}\\ t_i\in\mathcal{T}}}
\left[
\operatorname{loss}_{t_i}\!\left(
G_{\phi_{t_i}}(e_i,x_i^{\mathrm{task}}),
y_i^{\mathrm{task}}
\right)
\right], \\
\{\phi_t^\star\}_{t\in\mathcal{T}}
&=
\arg\min_{\{\phi_t\}_{t\in\mathcal{T}}}
\mathcal{L}_{\mathrm{task}},
\qquad
\theta^\star\ \text{fixed}.
\end{align}
The loss $\operatorname{loss}_{t_i}$ is defined by the native result space. Continuous property tasks measure numerical deviation from the reference value. Classification and protein-site tasks measure prediction errors over class or residue labels. Spectrum-to-molecular generation and protein structure tasks use the structural or geometric supervision associated with their generation process. Scientific image generation and editing use the latent flow-matching objective of the generative model. Task decoders are trained independently and do not update through a shared cross-task weighted objective.

\FloatBarrier
\sloppy
\setlength{\LTcapwidth}{\textwidth}
\section{Evaluations}
\label{sec:evaluations}
\label{sec:experiments}

We evaluate \model as a unified multimodal reasoning model for scientific understanding, prediction, and generation across molecules, materials, proteins, spectra, and scientific images. The evaluation spans more than 60 benchmarks in five task families: property prediction, spectrum-to-molecular generation, residue-level protein functional-site prediction, all-atom protein structure prediction, and scientific image generation and editing. We ask whether unified representation of scientific data, natural-world knowledge alignment, and decoding for domain-specific tasks can produce transferable scientific representations within a single model and translate them into verifiable domain-native outputs, including scalar values, functional sites, molecular structures, atomic coordinates, and images.

\subsection{Property Prediction}
\label{sec:eval_4_1_property_prediction}

Property prediction evaluates whether a model can infer continuous properties and classification labels from molecular or crystal structures. The task family spans materials, quantum chemistry, molecular physicochemistry, and ADMET.

\subsubsection{Experimental Setup}
\label{sec:eval_4_1_1_experiment_setup}

The model predicts continuous properties or class probabilities from molecular SMILES strings or crystal CIF files. We evaluate 48 properties across materials, quantum chemistry, molecular physicochemistry, and ADMET to test whether one scientific model can maintain stable quantitative predictions across heterogeneous structural inputs and output types.

\textbf{Setting.} The evaluation combines 10 JARVIS-DFT material regression properties \citep{choudhary2020joint}, two Materials Project classification properties \citep{jain2013materials}, four COMPAS-1D quantum-chemical properties \citep{wahab2022compas}, seven QM9 properties \citep{ramakrishnan2014quantum}, seven MoleculeNet tasks \citep{wu2018moleculenet}, and 18 ADMET endpoints from the Therapeutics Data Commons \citep{huang2021therapeutics}. Following each benchmark, regression is evaluated by MAD/MAE, MAE, RMSE, or Spearman correlation, and classification by ROC-AUC or PRC-AUC. MAE and RMSE are lower-is-better; all other metrics are higher-is-better.

\textbf{Baselines.} Molecular and quantum-chemical comparisons use ChemFM-3B \citep{cai2025chemfm} and Uni-Mol2 1.1B \citep{evalJi2024UniMol2}; material comparisons use LLM-Prop-35M \citep{niyongabo2025llm} and MatBERT-109M \citep{trewartha2022quantifying}. General-purpose baselines are GPT-5.5 \citep{singh2025openai} and Gemini-3.1-Pro \citep{evalGeminiTeam2024}. Aggregate specialist counts use the stronger applicable specialist for each property. The distribution-shift analysis additionally compares GotenNet \citep{evalAykent2024GotenNet} and Regression Transformer \citep{evalBorn2023RegressionTransformer} under BOOM \citep{evalAntoniuk2025BOOM}.

\subsubsection{Main Results}
\label{sec:eval_4_1_2_main_results}

Tables~\ref{tab:eval_4_1_3}, \ref{tab:eval_4_1_2}, and~\ref{tab:eval_4_1_1} report results for molecular physicochemical and ADMET properties, quantum-chemical properties, and material properties, respectively. For each group, we compare \model with the applicable domain-specific models and with two general-purpose proprietary models.

\textbf{Molecular Physicochemical and ADMET Prediction.}
\label{sec:eval_4_molecular_physicochemical_and_admet_prediction}

MoleculeNet evaluates basic physicochemical properties together with molecular activity and toxicity, whereas the ADMET tasks cover absorption, distribution, metabolism, excretion, and toxicity. Collectively, these endpoints define a central family of early-stage filters in drug discovery.

\begin{table}[t!]
\centering
\caption{Property prediction on MoleculeNet and TDC ADMET. Each score uses the metric shown in the same row; arrows indicate the preferred direction. Best and second-best results in each row are bold and underlined, respectively.}\label{tab:eval_4_1_3}
\scriptsize
\setlength{\tabcolsep}{2pt}
\renewcommand{\arraystretch}{0.98}
\begin{adjustbox}{width=\textwidth}
\begin{tabular}{@{}>{\raggedright\arraybackslash}p{0.094\textwidth}>{\raggedright\arraybackslash}p{0.190\textwidth}>{\raggedright\arraybackslash}p{0.090\textwidth}>{\centering\arraybackslash}p{0.104\textwidth}>{\centering\arraybackslash}p{0.104\textwidth}>{\centering\arraybackslash}p{0.122\textwidth}>{\centering\arraybackslash}p{0.104\textwidth}@{}}
\toprule
Dataset & Property & Metric & ChemFM-3B & GPT-5.5 & Gemini-3.1-Pro & S1-Omni \\
\midrule
MoleculeNet & BBBP & ROC-AUC $\uparrow$ & \underline{0.7510} & 0.6773 & 0.6792 & \textbf{0.8578} \\
MoleculeNet & CLINTOX & ROC-AUC $\uparrow$ & \textbf{0.9180} & 0.5557 & 0.4304 & \underline{0.8063} \\
MoleculeNet & HIV & ROC-AUC $\uparrow$ & \textbf{0.8070} & 0.6688 & 0.6209 & \underline{0.6978} \\
MoleculeNet & SIDER & ROC-AUC $\uparrow$ & \textbf{0.7090} & 0.5587 & 0.5052 & \underline{0.5759} \\
MoleculeNet & ESOL & RMSE $\downarrow$ & \textbf{0.5160} & 68.2500 & 19.4000 & \underline{1.2217} \\
MoleculeNet & FREESOLV & RMSE $\downarrow$ & \underline{0.8300} & 29.2869 & 38.8370 & \textbf{0.7092} \\
MoleculeNet & LIPO & RMSE $\downarrow$ & \textbf{0.5450} & 17.8300 & 980.1600 & \underline{0.8845} \\
ADMET & CACO2\_\allowbreak{}WANG & MAE $\downarrow$ & \textbf{0.3220} & 7.0920 & 12.7883 & \underline{0.4005} \\
ADMET & LD50\_\allowbreak{}ZHU & MAE $\downarrow$ & \textbf{0.5410} & 18.3553 & 8.7699 & \underline{0.6625} \\
 ADMET & CYP2C9\_\allowbreak{}SUBSTRATE\_\allowbreak{}CARBONMANGELS & PRC-AUC $\uparrow$ & 0.4140 & \underline{0.5430} & 0.5172 & \textbf{0.6972} \\
ADMET & CYP2C9\_\allowbreak{}VEITH & PRC-AUC $\uparrow$ & \underline{0.7880} & 0.5223 & 0.5750 & \textbf{0.8292} \\
ADMET & CYP2D6\_\allowbreak{}SUBSTRATE\_\allowbreak{}CARBONMANGELS & PRC-AUC $\uparrow$ & \textbf{0.7390} & 0.5109 & 0.5279 & \underline{0.6651} \\
ADMET & CYP2D6\_\allowbreak{}VEITH & PRC-AUC $\uparrow$ & \underline{0.7040} & 0.5397 & 0.6500 & \textbf{0.8816} \\
ADMET & CYP3A4\_\allowbreak{}VEITH & PRC-AUC $\uparrow$ & \textbf{0.8780} & 0.5350 & 0.5309 & \underline{0.8410} \\
ADMET & AMES & ROC-AUC $\uparrow$ & \textbf{0.8540} & 0.6711 & 0.5987 & \underline{0.8340} \\
ADMET & BIOAVAILABILITY\_\allowbreak{}MA & ROC-AUC $\uparrow$ & \textbf{0.7150} & 0.5338 & 0.6742 & \underline{0.6960} \\
ADMET & CYP3A4\_\allowbreak{}SUBSTRATE\_\allowbreak{}CARBONMANGELS & ROC-AUC $\uparrow$ & \underline{0.6540} & 0.4621 & 0.5359 & \textbf{0.6654} \\
ADMET & DILI & ROC-AUC $\uparrow$ & \textbf{0.9200} & 0.5369 & 0.7090 & \underline{0.8675} \\
ADMET & HERG & ROC-AUC $\uparrow$ & \underline{0.8480} & 0.5616 & 0.5447 & \textbf{0.8734} \\
ADMET & HIA\_\allowbreak{}HOU & ROC-AUC $\uparrow$ & \underline{0.9840} & 0.5548 & 0.5961 & \textbf{0.9921} \\
ADMET & PGP\_\allowbreak{}BROCCATELLI & ROC-AUC $\uparrow$ & \textbf{0.9310} & 0.5665 & 0.6905 & \underline{0.9224} \\
ADMET & CLEARANCE\_\allowbreak{}HEPATOCYTE\_\allowbreak{}AZ & Spearman $\uparrow$ & \textbf{0.4950} & -0.0016 & -0.0367 & \underline{0.4794} \\
ADMET & CLEARANCE\_\allowbreak{}MICROSOME\_\allowbreak{}AZ & Spearman $\uparrow$ & \textbf{0.6110} & 0.1672 & 0.4418 & \underline{0.5926} \\
ADMET & HALF\_\allowbreak{}LIFE\_\allowbreak{}OBACH & Spearman $\uparrow$ & \underline{0.5510} & 0.4342 & \textbf{0.6465} & 0.5209 \\
ADMET & VDSS\_\allowbreak{}LOMBARDO & Spearman $\uparrow$ & \textbf{0.6620} & 0.3538 & \underline{0.5915} & 0.5861 \\
\bottomrule
\end{tabular}
\end{adjustbox}
\end{table}

On MoleculeNet, \model exceeds ChemFM-3B on BBBP, reaching 0.8578 ROC-AUC, and on FreeSolv, reaching 0.7092 RMSE; ChemFM-3B leads on the other five tasks (Table~\ref{tab:eval_4_1_3}). On TDC ADMET, \model surpasses the specialist result on \texttt{CYP2C9\_\allowbreak{}SUBSTRATE\_\allowbreak{}CARBONMANGELS}, \texttt{CYP2C9\_\allowbreak{}VEITH}, \texttt{CYP2D6\_\allowbreak{}VEITH}, \texttt{CYP3A4\_\allowbreak{}SUBSTRATE\_\allowbreak{}CARBONMANGELS}, \texttt{HERG}, and \texttt{HIA\_\allowbreak{}HOU}. It also outperforms the stronger closed model on all seven MoleculeNet tasks and 16 of 18 ADMET endpoints; the exceptions are \texttt{HALF\_\allowbreak{}LIFE\_\allowbreak{}OBACH} and \texttt{VDSS\_\allowbreak{}LOMBARDO}.

The strongest results occur on classification and ranking endpoints. BBBP, CYP, hERG, and HIA require the same molecular structure to be interpreted through distinct notions of permeability, metabolic recognition, transport, or structural alerts, and the role of a functional group can change with the endpoint. Conditioning the shared molecular representation on the task instruction helps separate these property semantics. The organization of metabolic soft spots, transport, and toxicity mechanisms in the structured reasoning data is also consistent with this result pattern. By contrast, ESOL, LIPO, clearance, and half-life expose numerical calibration and continuous ranking as the principal remaining sources of error.

\textbf{Quantum Chemical Property Prediction.}
\label{sec:eval_4_quantum_chemical_property_prediction}

COMPAS-1D and QM9 test whether a model can recover electron affinity, ionization potential, dipole moment, frontier-orbital energies, energy gaps, polarizability, and zero-point vibrational energy from molecular structure. These quantities underpin analyses of chemical reactivity and molecular design.

\begin{table}[t!]
\centering
\caption{Property prediction on COMPAS-1D and QM9. Each score uses the metric shown in the same row; lower is better for all rows. Best and second-best results in each row are bold and underlined, respectively.}\label{tab:eval_4_1_2}
\scriptsize
\setlength{\tabcolsep}{2pt}
\renewcommand{\arraystretch}{1.05}
\begin{adjustbox}{width=\textwidth}
\begin{tabular}{@{}>{\raggedright\arraybackslash}p{0.085\textwidth}>{\raggedright\arraybackslash}p{0.130\textwidth}>{\raggedright\arraybackslash}p{0.074\textwidth}>{\centering\arraybackslash}p{0.112\textwidth}>{\centering\arraybackslash}p{0.128\textwidth}>{\centering\arraybackslash}p{0.100\textwidth}>{\centering\arraybackslash}p{0.119\textwidth}>{\centering\arraybackslash}p{0.100\textwidth}@{}}
\toprule
Dataset & Property & Metric & ChemFM-3B & Uni-Mol2 1.1B & GPT-5.5 & Gemini-3.1-Pro & S1-Omni \\
\midrule
COMPAS-1D & AEA\_\allowbreak{}EV & MAE $\downarrow$ & \textbf{0.0093} & -- & 3.1778 & 2.4314 & \underline{0.0599} \\
COMPAS-1D & AIP\_\allowbreak{}EV & MAE $\downarrow$ & \textbf{0.0074} & -- & 4.2155 & 1.7910 & \underline{0.0508} \\
COMPAS-1D & D3\_\allowbreak{}DISP\_\allowbreak{}CORR\_\allowbreak{}EV & MAE $\downarrow$ & \textbf{0.0067} & -- & 9.4398 & 9.1830 & \underline{0.0699} \\
COMPAS-1D & DIPMOM\_\allowbreak{}DEBYE & MAE $\downarrow$ & \textbf{0.0170} & -- & 1.0284 & 0.1701 & \underline{0.0613} \\
QM9 & ALPHA & MAE $\downarrow$ & -- & \textbf{0.3050} & 68.0723 & 44.8064 & \underline{6.7366} \\
QM9 & CV & MAE $\downarrow$ & -- & \textbf{0.1440} & 35.1990 & 7.7852 & \underline{3.0004} \\
QM9 & GAP & MAE $\downarrow$ & -- & \textbf{0.0035} & 4.6780 & 0.7438 & \underline{0.0224} \\
QM9 & HOMO & MAE $\downarrow$ & -- & \textbf{0.0035} & 1.9033 & 4.8781 & \underline{0.0224} \\
QM9 & LUMO & MAE $\downarrow$ & -- & \textbf{0.0035} & 1.5952 & 2.1027 & \underline{0.0242} \\
QM9 & MU & MAE $\downarrow$ & -- & \textbf{0.0890} & 1.8037 & 1.0133 & \underline{0.8445} \\
QM9 & ZPVE & MAE $\downarrow$ & -- & \textbf{0.0005} & 29.0415 & 124.3091 & \underline{0.0244} \\
\bottomrule
\end{tabular}
\end{adjustbox}
\end{table}

\model outperforms both closed models on all 11 quantum-chemical properties, but the applicable specialists retain substantially lower error (Table~\ref{tab:eval_4_1_2}). Its four COMPAS-1D MAEs range from 0.0508 to 0.0699, compared with 0.0067 to 0.0170 for ChemFM-3B. On QM9, the largest specialist margins occur for \texttt{ALPHA}, \texttt{CV}, and \texttt{ZPVE}. These quantities depend on different electronic and geometric factors and are sensitive to unit scale and small structural changes. The shared predictor stabilizes the output form, but three-dimensional pretraining gives specialist models a finer account of conformation and electronic environment.

\textbf{Materials Property Prediction.}
\label{sec:eval_4_materials_property_prediction}

The JARVIS-DFT regression tasks predict electronic, mechanical, and optoelectronic properties from CIF structures for materials screening. The two Materials Project tasks classify whether a material has a direct band gap and whether it is stable, corresponding to screening for optoelectronic utility and thermodynamic synthesizability.

\begin{table}[t!]
\centering
\caption{Property prediction on JARVIS-DFT and the Materials Project. Each score uses the metric shown in the same row; arrows indicate the preferred direction. Best and second-best results in each row are bold and underlined, respectively.}\label{tab:eval_4_1_1}
\scriptsize
\setlength{\tabcolsep}{2pt}
\begin{adjustbox}{width=\textwidth}
\begin{tabular}{@{}>{\raggedright\arraybackslash}p{0.090\textwidth}>{\raggedright\arraybackslash}p{0.100\textwidth}>{\raggedright\arraybackslash}p{0.100\textwidth}>{\centering\arraybackslash}p{0.116\textwidth}>{\centering\arraybackslash}p{0.116\textwidth}>{\centering\arraybackslash}p{0.100\textwidth}>{\centering\arraybackslash}p{0.119\textwidth}>{\centering\arraybackslash}p{0.100\textwidth}@{}}
\toprule
Dataset & Property & Metric & LLM-Prop-35M & MatBERT-109M & GPT-5.5 & Gemini-3.1-Pro & S1-Omni \\
\midrule
JARVIS-DFT & BANDGAP\_\allowbreak{}OPT & \mbox{MAD/MAE~$\uparrow$} & \underline{3.3310} & \textbf{5.4830} & 1.4281 & 2.9381 & 1.5718 \\
JARVIS-DFT & EPS\_\allowbreak{}DFPT & \mbox{MAD/MAE~$\uparrow$} & \textbf{1.5780} & 1.5090 & 0.1953 & 0.5605 & \underline{1.5559} \\
JARVIS-DFT & EXF\_\allowbreak{}EN & \mbox{MAD/MAE~$\uparrow$} & \underline{1.0440} & \textbf{1.3740} & 0.214 & 0.2319 & 0.7264 \\
JARVIS-DFT & FEPA\_\allowbreak{}JARVIS & \mbox{MAD/MAE~$\uparrow$} & \textbf{12.9960} & \underline{10.2110} & 1.1604 & 3.3473 & 1.5125 \\
JARVIS-DFT & GV & \mbox{MAD/MAE~$\uparrow$} & \textbf{2.6230} & \underline{2.4190} & 0.4622 & 1.3881 & 1.8545 \\
JARVIS-DFT & KV & \mbox{MAD/MAE~$\uparrow$} & \underline{4.1280} & \textbf{4.2520} & 1.9599 & 3.2534 & 2.9970 \\
JARVIS-DFT & MAX\_\allowbreak{}EFG & \mbox{MAD/MAE~$\uparrow$} & \underline{1.9360} & \textbf{2.1430} & 0.5167 & 0.7287 & 1.5360 \\
JARVIS-DFT & SLME & \mbox{MAD/MAE~$\uparrow$} & \underline{2.1750} & \textbf{2.2080} & 0.8678 & 1.1965 & 1.6484 \\
JARVIS-DFT & SPILLAGE & \mbox{MAD/MAE~$\uparrow$} & \textbf{1.4440} & \underline{0.9160} & 0.3646 & 0.8256 & 0.7037 \\
JARVIS-DFT & TOT\_\allowbreak{}EN & \mbox{MAD/MAE~$\uparrow$} & \textbf{23.5090} & \underline{15.1410} & 0.4565 & 0.6738 & 2.9861 \\
Materials Project & IS\_\allowbreak{}GAP\_\allowbreak{}DIRECT & \mbox{ROC-AUC~$\uparrow$} & \underline{0.7000} & \textbf{0.7100} & 0.505 & 0.5886 & 0.5772 \\
Materials Project & IS\_\allowbreak{}STABLE & \mbox{ROC-AUC~$\uparrow$} & \underline{0.7760} & \textbf{0.7900} & 0.5163 & 0.6431 & 0.6231 \\
\bottomrule
\end{tabular}
\end{adjustbox}
\end{table}

At least one specialist model outperforms \model on each of the 12 material properties (Table~\ref{tab:eval_4_1_1}). \model nevertheless exceeds the stronger closed model on \texttt{EPS\_\allowbreak{}DFPT}, \texttt{EXF\_\allowbreak{}EN}, \texttt{GV}, \texttt{MAX\_\allowbreak{}EFG}, \texttt{SLME}, and \texttt{TOT\_\allowbreak{}EN}. Its MAD/MAE of 1.5559 on \texttt{EPS\_\allowbreak{}DFPT} approaches LLM-Prop-35M at 1.5780, whereas the two Materials Project classifications reach 0.5772 and 0.6231, below MatBERT-109M at 0.7100 and 0.7900. Band gaps, dielectric response, elastic moduli, formation energies, and stability depend on different combinations of lattice periodicity, local coordination, and electronic structure. The results show that the shared encoder forms a readable task representation from CIF inputs, but high-precision materials prediction still benefits substantially from specialist inductive biases over atomic neighborhoods, long-range periodic relations, and electronic structure.

Across all three groups, \model exceeds the reported specialist result on 8 of 48 properties and the stronger closed model on 40 of 48. The unified representation of scientific data supports both CIF and SMILES inputs and both regression and classification, while the remaining specialist advantage quantifies the value of object-specific structural priors.

\subsubsection{Ablation Analysis}
\label{sec:eval_4_1_3_ablation_analysis}

The ablations examine the structure of reasoning supervision, hidden-state pooling, explicit numerical ranges in the reasoning trace, predictor sharing, training stage and data scale, and label normalization.

Reasoning supervision is evaluated on six JARVIS-DFT properties. Pooling, predictor sharing, and training-stage experiments use a 12-task suite comprising six JARVIS-DFT and six MoleculeNet properties. Explicit range prediction uses the two chemical properties ESOL and LIPO, and label normalization uses all 48 properties.

\textbf{Effect of Reasoning Supervision Structure.} \label{sec:eval_4_effect_of_reasoning_supervision_structure} We fix the Stage-2 shared predictor, question-only pooling, training scale, and optimization, and vary only the Stage-1 reasoning supervision. Both reasoning datasets are distilled from Qwen3.7 and differ only in the system prompt, allowing us to isolate reasoning quality. Structured, property-constrained reasoning requires a systematic analysis of lattice structure, composition, local coordination, and symmetry, followed by selection of the three to five factors most relevant to the target property and an account of their direction, strength, and interactions. Unconstrained free-form reasoning provides only a broad request for analysis and leaves the evidence, order, and depth to the teacher model. The no-reasoning condition retains an empty reasoning span and the task-routing token.

\begin{table}[H]
\centering
\caption{Effect of Stage-1 reasoning supervision on six material regression properties. All variants use the same shared predictor and question-only pooling; lower MAE and higher MAD/MAE are better. Best and second-best results are bold and underlined, respectively.}\label{tab:eval_4_1_4}
\scriptsize
\setlength{\tabcolsep}{2pt}
\begin{adjustbox}{width=\textwidth}
\begin{tabular}{@{}>{\raggedright\arraybackslash}p{0.379\textwidth}>{\centering\arraybackslash}p{0.246\textwidth}>{\centering\arraybackslash}p{0.279\textwidth}@{}}
\toprule
Stage-1 data & Material MAE $\downarrow$ & Material MAD/MAE $\uparrow$ \\
\midrule
Structured, property-constrained reasoning & \textbf{17.255} & \textbf{1.260} \\
No reasoning text & \underline{20.457} & \underline{0.941} \\
Unconstrained free-form reasoning & 34.818 & 0.888 \\
\bottomrule
\end{tabular}
\end{adjustbox}
\end{table}

Structured reasoning gives the lowest material MAE and the highest MAD/MAE (Table~\ref{tab:eval_4_1_4}). Free-form reasoning reaches an MAE of 34.818, compared with 17.255 for structured reasoning and 20.457 without reasoning text. Adding arbitrary reasoning text therefore does not improve property prediction by itself.

The benefit of reasoning supervision depends on whether the scientific evidence is organized consistently around the target property. Structured supervision provides a stable learning signal; unconstrained traces vary in evidence coverage and analytical path and can be less effective than omitting reasoning altogether.

\textbf{Effect of Hidden-State Pooling Scope.}
\label{sec:eval_4_effect_of_hidden_state_pooling_scope}
We next test which part of the sequence should supply the task-conditioned representation to the scalar predictor. On the 12-task suite, we fix the backbone, predictor, training data, and optimization and vary only the hidden-state span covered by mean pooling. Question-only pooling aggregates the CIF or SMILES input and its property instruction, whereas full pooling additionally includes the subsequently generated reasoning and answer states. The two variants use the same parameterization and training procedure, so their difference isolates the extraction scope.

\begin{table}[H]
\centering
\caption{Effect of hidden-state pooling scope on six chemical and six material properties. Higher chemical F1 and material MAD/MAE are better; lower chemical RMSE is better. Best and second-best results are bold and underlined, respectively.}\label{tab:eval_4_1_5}
\scriptsize
\setlength{\tabcolsep}{2pt}
\begin{adjustbox}{width=\textwidth}
\begin{tabular}{@{}>{\raggedright\arraybackslash}p{0.239\textwidth}>{\centering\arraybackslash}p{0.219\textwidth}>{\centering\arraybackslash}p{0.211\textwidth}>{\centering\arraybackslash}p{0.223\textwidth}@{}}
\toprule
Pooling scope & Chemical classification F1 $\uparrow$ & Chemical regression RMSE $\downarrow$ & Material regression \mbox{MAD/MAE $\uparrow$} \\
\midrule
Question only & \textbf{0.822} & \textbf{0.908} & \textbf{1.454} \\
Full & \underline{0.773} & \underline{1.663} & \underline{0.899} \\
\bottomrule
\end{tabular}
\end{adjustbox}
\end{table}

Question-only pooling improves every metric (Table~\ref{tab:eval_4_1_5}): chemical F1 rises from 0.773 to 0.822, chemical RMSE falls from 1.663 to 0.908, and material MAD/MAE rises from 0.899 to 1.454. Answer-side states mix scientific evidence with wording, reasoning length, and termination patterns. The result indicates that property information can already be extracted before answer generation; reasoning supervision shapes the prefill representation rather than directly producing the numerical prediction.

\textbf{Effect of Explicit Range Output in Reasoning.}
\label{sec:eval_4_effect_of_explicit_range_output}
We test whether predicting a plausible numerical interval before the final value provides a useful intermediate target. The two settings retain the same reasoning process and differ only in whether the model must first state an approximate range. The motivation is that an explicit interval might narrow the search space and provide an intermediate constraint for the final estimate.

\begin{table}[H]
\centering
\caption{Effect of a preliminary range prediction on mixed ESOL and LIPO regression. Lower RMSE is better. Best and second-best results are bold and underlined, respectively.}\label{tab:eval_4_1_6}
\scriptsize
\setlength{\tabcolsep}{2pt}
\begin{adjustbox}{width=\textwidth}
\begin{tabular}{@{}>{\raggedright\arraybackslash}p{0.44\textwidth}>{\centering\arraybackslash}p{0.23\textwidth}>{\centering\arraybackslash}p{0.23\textwidth}@{}}
\toprule
Reasoning supervision & Training records & RMSE $\downarrow$ \\
\midrule
Explicit preliminary value range & 4,000 & \textbf{0.2812} \\
No preliminary value range & 4,000 & \underline{0.2814} \\
\bottomrule
\end{tabular}
\end{adjustbox}
\end{table}

The hypothesis is not supported. The model does not reliably predict the target range: in some cases, the point estimate is close to the label even when the preliminary interval is substantially displaced. Final performance is also essentially unchanged, with RMSE values of 0.2812 with range prediction and 0.2814 without it (Table~\ref{tab:eval_4_1_6}). Explicit range prediction is therefore not an effective intermediate reasoning signal. A more promising direction is supervision that helps decompose the problem and identify the evidence directly relevant to the target.

\textbf{Effect of Predictor Sharing.}
\label{sec:eval_4_effect_of_predictor_sharing}
We test whether sharing one predictor across all properties introduces interference between heterogeneous label spaces. On the 12-task suite, we fix the 140k training records, Stage 2, question-only pooling, SFT optimization, and evaluation metrics, and vary only the predictor organization. The shared variant uses one predictor module for all properties, whereas the property-specific variant assigns an independent predictor to each property. Both settings cover six chemical and six material properties.

\begin{table}[t]
\centering
\caption{Effect of shared versus property-specific predictors on six chemical and six material properties under the same 140k Stage-2 setting. Higher chemical F1 and material MAD/MAE are better; lower chemical RMSE is better. Best and second-best results are bold and underlined, respectively.}\label{tab:eval_4_1_7}
\scriptsize
\setlength{\tabcolsep}{2pt}
\begin{adjustbox}{width=\textwidth}
\begin{tabular}{@{}>{\raggedright\arraybackslash}p{0.23\textwidth}>{\centering\arraybackslash}p{0.22\textwidth}>{\centering\arraybackslash}p{0.22\textwidth}>{\centering\arraybackslash}p{0.22\textwidth}@{}}
\toprule
Predictor & Chemical classification F1 $\uparrow$ & Chemical regression RMSE $\downarrow$ & Material regression \mbox{MAD/MAE $\uparrow$} \\
\midrule
Shared & \textbf{0.8741} & \underline{0.8789} & \underline{1.5553} \\
Property-specific & \underline{0.8705} & \textbf{0.8250} & \textbf{2.0283} \\
\bottomrule
\end{tabular}
\end{adjustbox}
\end{table}

The shared predictor is marginally better for chemical classification, with F1 values of 0.8741 and 0.8705, whereas property-specific heads improve both regression metrics (Table~\ref{tab:eval_4_1_7}). Chemical RMSE falls from 0.8789 to 0.8250 and material MAD/MAE rises from 1.5553 to 2.0283. Shared parameters preserve the simplest cross-property interface; independent heads reduce interference among continuous targets with different units, ranges, and structural dependencies.

\textbf{Effect of Training Stage and Data Scale.}
\label{sec:eval_4_effect_of_training_stage_and_data_scale}
To distinguish learning a terminal predictor from adapting the shared representation, and to measure how data scale affects each regime, we construct four SFT settings on the 12-task suite. Stage 2 freezes the backbone and trains only the property predictor; Stage 3 updates both. Each endpoint is evaluated with 27k and 140k training records under identical question-only pooling, predictor architectures, and metrics.

\begin{table}[H]
\centering
\caption{Joint effects of property-training stage and data scale on six chemical and six material properties. Higher chemical F1 and material MAD/MAE are better; lower chemical RMSE is better. Best and second-best results are bold and underlined, respectively.}\label{tab:eval_4_1_8}
\scriptsize
\setlength{\tabcolsep}{2pt}
\begin{adjustbox}{width=\textwidth}
\begin{tabular}{@{}>{\raggedright\arraybackslash}p{0.25\textwidth}>{\centering\arraybackslash}p{0.21\textwidth}>{\centering\arraybackslash}p{0.21\textwidth}>{\centering\arraybackslash}p{0.21\textwidth}@{}}
\toprule
Training endpoint and data scale & Chemical classification F1 $\uparrow$ & Chemical regression RMSE $\downarrow$ & Material regression MAD/MAE $\uparrow$ \\
\midrule
Stage 2, 27k & 0.822 & 0.908 & 1.454 \\
Stage 2, 140k & \textbf{0.874} & 0.879 & \underline{1.555} \\
Stage 3, 27k & 0.862 & \underline{0.855} & 1.525 \\
Stage 3, 140k & \underline{0.872} & \textbf{0.790} & \textbf{1.865} \\
\bottomrule
\end{tabular}
\end{adjustbox}
\end{table}

Data scale and end-to-end adaptation are complementary (Table~\ref{tab:eval_4_1_8}). Increasing Stage-2 data from 27k to 140k improves chemical F1 from 0.822 to 0.874, lowers chemical RMSE from 0.908 to 0.879, and raises material MAD/MAE from 1.454 to 1.555. The same expansion in Stage 3 improves chemical F1 from 0.862 to 0.872, lowers RMSE from 0.855 to 0.790, and raises material MAD/MAE from 1.525 to 1.865. At fixed data scale, Stage 3 improves all three metrics with 27k records; with 140k records, it substantially improves both regression metrics while leaving chemical F1 nearly unchanged. More data improve coverage of the mapping from a fixed representation to labels, whereas end-to-end property supervision further reshapes the shared representation for lightweight continuous prediction. The 140k Stage-3 setting therefore gives the best chemical and material regression results.

\textbf{Effect of Per-Property Label Normalization.}
\label{sec:eval_4_effect_of_per_property_label_normalization}
Across all 48 properties, we compare raw labels with independent normalization of each regression property followed by restoration to the original units at inference time; classification labels remain binary in both settings. Outcomes are grouped by whether they exceed, fall within 5\% of, or remain more than 5\% below the reported specialist.

\begin{table}[H]
\centering
\caption{Effect of regression-label normalization across 48 properties with a shared predictor. Counts compare each property with its reported specialist result. Best and second-best results are bold and underlined, respectively.}\label{tab:eval_4_1_9}
\scriptsize
\setlength{\tabcolsep}{2pt}
\begin{adjustbox}{width=\textwidth}
\begin{tabular}{@{}>{\raggedright\arraybackslash}p{0.181\textwidth}>{\centering\arraybackslash}p{0.181\textwidth}>{\centering\arraybackslash}p{0.181\textwidth}>{\centering\arraybackslash}p{0.167\textwidth}>{\centering\arraybackslash}p{0.170\textwidth}@{}}
\toprule
Normalization strategy & Better than reported specialist (count) $\uparrow$ & Within 5\% of reported specialist (count) $\uparrow$ & More than 5\% worse (count) $\downarrow$ & Better or within 5\% (count) $\uparrow$ \\
\midrule
No normalization & \textbf{8} & \underline{7} & \underline{33} & \underline{15} \\
Per-property normalization & \underline{7} & \textbf{11} & \textbf{30} & \textbf{18} \\
\bottomrule
\end{tabular}
\end{adjustbox}
\end{table}

Raw labels yield eight strict specialist wins, compared with seven after normalization (Table~\ref{tab:eval_4_1_9}). Normalization increases the number within 5\% from 7 to 11 and reduces the number more than 5\% worse from 33 to 30, raising the combined competitive count from 15 to 18. Per-property normalization therefore improves balance across heterogeneous label scales, whereas raw labels attain a higher peak on more individual properties.

Despite its broader coverage, per-property normalization requires a separate normalization and inverse transformation for every property. In open-ended deployment, a user request may not be resolved reliably to a single property and its associated transform. To avoid this task-identification and parameter-matching dependency, the main model uses raw labels and returns predictions directly in their native units.

\textbf{Effect of Distribution Shift.}
\label{sec:eval_4_effect_of_distribution_shift}

\textbf{Setting.} We follow the BOOM full label-shift protocol \citep{evalAntoniuk2025BOOM} on eight QM9 properties: $\alpha$, $C_v$, gap, HOMO, LUMO, $\mu$, $R^2$, and ZPVE. For each property, BOOM estimates the target density $P(y)$ by kernel density estimation, assigns the lowest-density regions of label space to the OOD test set, randomly draws an ID test set from the remaining samples, and uses the rest for training. After task-level filtering, we reconstruct mutually exclusive training, ID, and OOD splits by the same rule. Their actual sizes therefore depend on the filtered data rather than on a fixed fraction of the complete QM9 collection.

We report ID and OOD RMSE, the OOD/ID RMSE ratio, and binned $R^2$ within the lower and upper OOD tails. RMSE measures absolute prediction error, OOD/ID measures relative error amplification from ID to OOD, and binned $R^2$ measures predictive correlation in the two extremes of the OOD label distribution. Lower RMSE and OOD/ID are better; higher binned $R^2$ is better.

\textbf{Baselines.} We compare with GotenNet and Regression Transformer (RT) as reported by BOOM. GotenNet is an equivariant graph neural network over three-dimensional molecular structure. RT takes SMILES as input and decodes numerical values autoregressively. \model uses a unified encoder for CIF and SMILES and a shared linear regression head across molecular properties. All models are compared under the BOOM full label-shift protocol.

\begin{table}[H]
\centering
\caption{ID and OOD RMSE on eight QM9 molecular properties under the BOOM full label-shift protocol. Each entry reports ID/OOD RMSE; lower is better. Best and second-best results are bold and underlined, respectively.}
\label{tab:eval_4_1_10}

\scriptsize
\renewcommand{\arraystretch}{1.15}
\setlength{\tabcolsep}{2.5pt}

\begin{tabularx}{\textwidth}{@{}>{\raggedright\arraybackslash}p{0.10\textwidth}*{8}{>{\centering\arraybackslash}X}@{}}
\toprule
Model & $\alpha$ & $C_v$ & Gap & HOMO & LUMO & $\mu$ & $R^2$ & ZPVE \\
\midrule
GotenNet &
\textbf{0.550/1.830} &
\textbf{0.130/0.340} &
\underline{0.090/0.280} &
\underline{0.052/0.170} &
\underline{0.020/0.033} &
\textbf{0.320/1.500} &
\textbf{5.070/17.070} &
\textbf{0.001/0.001} \\

Regression Transformer &
2.950/74345 &
1.280/2.810 &
0.210/0.550 &
0.130/0.319 &
0.190/0.860 &
0.870/2.350 &
21.350/25458 &
\underline{0.008}/0.019 \\

S1-Omni &
\underline{2.290/3.770} &
\underline{1.060/1.390} &
\textbf{0.014/0.036} &
\textbf{0.011/0.020} &
\textbf{0.017/0.026} &
\underline{0.800/1.920} &
\underline{19.290/154.640} &
\underline{0.008/0.009} \\
\bottomrule
\end{tabularx}

\end{table}

\textbf{Absolute error.} Table~\ref{tab:eval_4_1_10} does not yield a single model ranking; instead, performance is strongly property dependent. GotenNet has lower ID and OOD RMSE on $\alpha$, $C_v$, $\mu$, $R^2$, and ZPVE, whereas \model is lowest on gap, HOMO, and LUMO. The split is consistent with their inductive biases: equivariant three-dimensional representations favor properties governed by molecular scale, spatial conformation, and three-dimensional charge distribution, while the unified structural representation is effective for the orbital-energy targets in this benchmark. Label shift therefore does not remove the importance of representation choice; extrapolation still depends on whether the representation contains the structural information required by the property.

RT exhibits a qualitatively different anomaly on $\alpha$ and $R^2$: RMSE grows from 2.95 and 21.35 in ID to 74,345 and 25,458 in OOD, respectively. This is not ordinary error growth but an orders-of-magnitude numerical explosion, indicating catastrophic failure in these OOD label regions. Because RT generates values autoregressively, the result suggests that discrete numerical decoding can produce extreme outliers when targets leave the main support of the training labels. Neither GotenNet nor the continuous regression head of \model exhibits numerical instability at this scale.

\begin{figure}[t]
  \centering
  \includegraphics[width=0.85\textwidth]{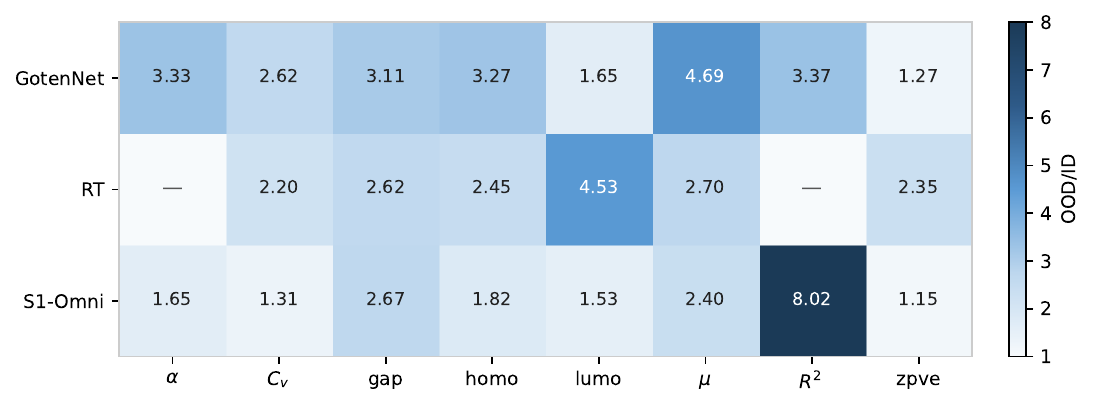}
  \caption{OOD-to-ID RMSE ratios for GotenNet, Regression Transformer (RT), and \model on eight QM9 properties under BOOM label shift. Values closer to 1 indicate less relative error growth; dashes denote catastrophic RT failures excluded from the ratio scale.}
  \label{fig:eval_4_1_ood_heatmap}
\end{figure}

\textbf{Relative OOD degradation.} Figure~\ref{fig:eval_4_1_ood_heatmap} provides a different view from absolute RMSE. \model has the lowest OOD/ID ratio on $\alpha$, $C_v$, HOMO, LUMO, $\mu$, and ZPVE and is close to RT on gap. It therefore incurs less additional error from ID to OOD on most properties. This does not imply uniformly better absolute predictions: GotenNet has lower ID and OOD RMSE on $\alpha$, $C_v$, $\mu$, and ZPVE. OOD/ID measures relative degradation rather than error magnitude, and a large ID baseline can itself reduce the ratio. The ratio must therefore be interpreted jointly with absolute RMSE.

For RT, $\alpha$ and $R^2$ correspond to error amplifications of approximately $2.52\times10^4$ and $1.19\times10^3$. Ratios of this magnitude no longer describe gradual degradation under shift; they reflect direct numerical instability on specific tasks. The result identifies a second form of OOD risk beyond a representation that fails to cover tail structures: autoregressive numerical decoding can produce uncontrolled extreme outputs, and cross-task averages can conceal these catastrophic failures.

The $R^2$ property illustrates a further distinction. \model has an OOD/ID ratio of 8.02, substantially above GotenNet at 3.37, yet its binned $R^2$ remains positive. The principal failure is therefore not a complete loss of predictive ordering but a large increase in error scale within the OOD region. Conversely, several electronic-structure properties have smaller OOD/ID ratios and low absolute RMSE but do not preserve the relative order of tail samples. Relative error amplification captures only one aspect of distribution shift.

\textbf{Tail correlation.} Binned $R^2$ directly tests whether a model preserves variation within the extreme OOD regions (Table~\ref{tab:eval_4_1_11}). GotenNet reaches 0.77, 0.99, 1.00, and 1.00 on $\alpha$, $C_v$, $R^2$, and ZPVE, respectively, showing that its three-dimensional representation both reduces error and preserves ordering across the lower and upper tails. \model reaches 0.66, 0.74, 0.46, and 0.70 on these properties. Its unified representation retains transferable trend information for several physical and thermodynamic quantities, but less consistently than explicit three-dimensional modeling.

\begin{table}[H]
\centering
\caption{Binned $R^2$ on the lower and upper OOD tails of eight QM9 molecular properties under BOOM label shift; higher is better. Best and second-best results are bold and underlined, respectively.}\label{tab:eval_4_1_11}
\scriptsize
\setlength{\tabcolsep}{2pt}
\begin{adjustbox}{width=\textwidth}
\begin{tabular}{@{}>{\raggedright\arraybackslash}p{0.20\textwidth}*{8}{>{\centering\arraybackslash}p{0.08875\textwidth}}@{}}
\toprule
Model & $\alpha$ & $C_v$ & Gap & HOMO & LUMO & $\mu$ & $R^2$ & ZPVE \\
\midrule
GotenNet & \textbf{0.77} & \textbf{0.99} & \textbf{0.03} & \textbf{0.15} & \textbf{0.33} & \textbf{0.11} & \textbf{1.00} & \textbf{1.00} \\
Regression Transformer & 0.06 & 0.08 & \textbf{0.03} & \underline{0.02} & \underline{0.02} & \underline{0.09} & 0.10 & 0.04 \\
S1-Omni & \underline{0.66} & \underline{0.74} & \underline{-3.04} & -1.53 & -1.91 & -0.57 & \underline{0.46} & \underline{0.70} \\
\bottomrule
\end{tabular}
\end{adjustbox}
\end{table}

The more consequential pattern appears for gap, HOMO, LUMO, and $\mu$. \model has the lowest RMSE on the first three electronic-structure properties, yet binned $R^2$ is negative for all four. Low average error alongside negative tail correlation means that the model is accurate in densely populated label regions but does not reliably learn the direction of variation in low-density tails; extreme predictions contract toward the center of the training distribution. Neither ID accuracy nor aggregate OOD RMSE is therefore sufficient evidence of extrapolation. Preserving tail structure is the stricter test.

\begin{figure}[t]
  \centering
  \includegraphics[width=\textwidth]{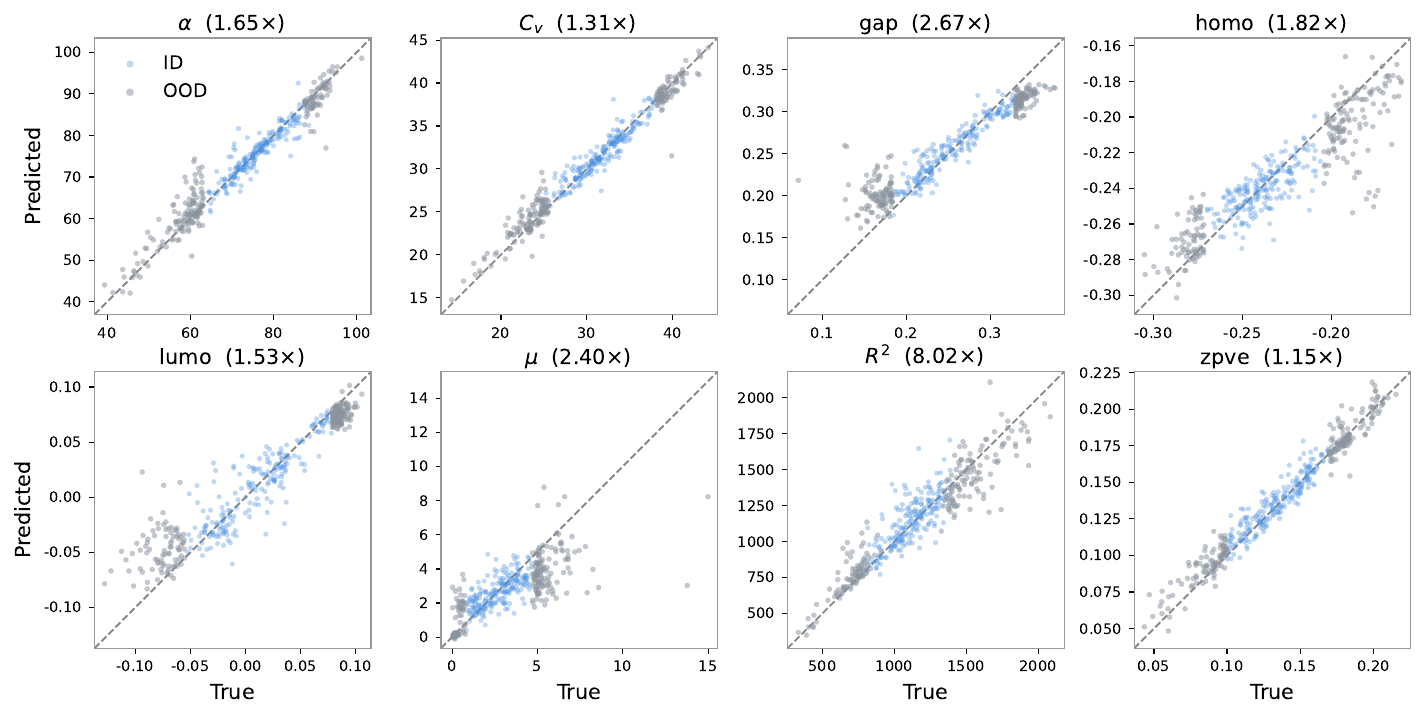}
  \caption{Predicted versus reference values for \model on ID and OOD splits of eight QM9 properties. Panel titles report the OOD/ID RMSE ratio; dashed lines show ideal agreement.}
  \label{fig:eval_4_1_parity}
\end{figure}

\textbf{Per-property behavior.} Figure~\ref{fig:eval_4_1_parity} separates these outcomes into two failure modes. The first is regression to the mean: OOD samples occupy both ends of the label distribution, but the predicted range is compressed, so low values are overestimated and high values underestimated. This pattern is clearest for gap, HOMO, LUMO, and $\mu$ and is consistent with their low or negative binned $R^2$. The failure is not a uniform increase in error but an inability to recover local slope and ordering in the tails.

The second mode is a mismatch in error scale. For $C_v$ and ZPVE, OOD/ID is 1.31 and 1.15, and both ID and OOD samples remain close to the diagonal, indicating that their structure--property relations extend comparatively stably into low-density regions. $\alpha$ shows some bilateral contraction but retains positive tail correlation. $R^2$ has the largest OOD/ID ratio, yet the scatter preserves a clear monotonic trend: the model captures the direction of change but underestimates the amplitude of high OOD values. Unlike the ordering failure for electronic-structure properties, this behavior is closer to a calibration or scale-extrapolation error.

Across all three metrics, the principal strengths of \model are smaller relative degradation on most properties and lower absolute error on the electronic-structure targets. Its limitation is not a single error shared by all OOD examples: different properties lose tail ordering or exhibit amplified error scale. The unified representation has learned property information that transfers across distributions, but it does not consistently encode the geometric and electronic factors controlling extreme values. Improving OOD generalization will require stronger supervision in low-density label regions and more direct three-dimensional constraints for geometry-sensitive properties, rather than further reductions in ID error alone.

\subsection{Spectrum-to-Molecular Generation}
\label{sec:eval_4_2_spectrum_to_molecule_reconstruction}

\subsubsection{Experimental Setup}
\label{sec:eval_4_2_1_experiment_setup}

The Spectrum-to-molecular generation infers a molecular structure from IR, Raman, and UV-Vis spectra. The output includes both a two-dimensional molecular graph and a 3D conformation. IR and Raman encode vibrational information, while UV-Vis reflects electronic transitions. Together, these modalities constrain functional groups, molecular connectivity, and geometry.

\textbf{Setting.} We use QM9S \citep{zou2023deep}, which provides 3D structures together with simulated IR, Raman, and UV-Vis spectra. Each UV-Vis spectrum has 601 values over 1.5 to 13.5 eV. The IR and Raman spectra each have 3,501 values over 500 to 4,000 cm$^{-1}$. Spectrum-to-molecular generation is evaluated with Acc@1, maximum common edge subgraph distance (MCES), Morgan and MACCS fingerprint similarity, FraggleSim, and functional-group similarity (FGSim). Lower MCES is better; all other generation metrics are higher-is-better. We also report AtomStable, MolStable, valid and complete (V\&C), valid and unique (V\&U), valid, unique, and novel (V\&U\&N), Frechet ChemNet Distance (FCD), similarity to nearest neighbor (SNN), fragment similarity (Frag), and scaffold similarity (Scaf). S1-Omni encodes normalized spectrum images as VLM representations. The \texttt{<spectra\_st>} route token selects the molecular diffusion decoder, and a projector maps the VLM representations into its conditioning space before the decoder generates the molecular graph and conformation.

\textbf{Baselines.} DiffSpectra \citep{wang2025diffspectra} is a 2D and 3D molecular diffusion model directly conditioned on multiple spectra. JODO \citep{huang2023learning} and BioMatrix \citep{pei2026biomatrix} provide reference values for unconditional molecular generation. We also evaluate GPT-5.5 from the GPT-5 family \citep{singh2025openai} and Gemini-3.1-Pro from the Gemini family \citep{evalGeminiTeam2024}. Each model receives both images converted from the original JCAMP-DX spectra and the corresponding raw sequences, and is prompted to produce a molecular structure in standard 3D SDF format.

\subsubsection{Molecular Structure Accuracy}
\label{sec:eval_4_2_1_spectrum_to_3d_structure}

Table~\ref{tab:eval_4_2_1} reports spectrum-to-molecular generation results. We first conduct a 100-sample pilot evaluation of GPT-5.5 and Gemini-3.1-Pro. For GPT-5.5, 71 of the 100 predictions form valid prediction-reference pairs after strict SDF parsing and molecular sanitization. The reported Acc@1 of 0.0563 is therefore computed conditionally over these valid pairs, corresponding to 4 exact matches out of 71. When invalid predictions are also counted as failures, its end-to-end Acc@1 over the full pilot set is 0.0400 (4/100). This distinction shows that GPT-5.5 is reasonably capable of producing parseable structures, but its exact structure accuracy remains low.
Direct generation is limited in this setting: GPT-5.5 has low exact-match structure accuracy, while Gemini-3.1-Pro has substantially lower output validity and does not reliably return valid, parseable SDF structures. Because neither output validity nor end-to-end generation quality supports scaling the pilot, we do not extend these models to the full test set. DiffSpectra and S1-Omni are evaluated on 10,000 samples, so their comparison measures performance on a matched scale.

\begin{table}[t!]
\centering
\caption{Spectrum-to-molecular generation from combined IR, Raman, and UV-Vis inputs on QM9S. Rows marked with \texttt{*} evaluate direct SDF generation by general-purpose models on 100 samples; the other rows evaluate spectrum-to-molecular generation models on 10,000 samples. Bold and underlined values denote the better and second-best results, respectively, between the matched spectrum-to-molecular generation models.}
\label{tab:eval_4_2_1}

\scriptsize
\setlength{\tabcolsep}{2pt}

\begin{adjustbox}{width=\textwidth}
\begin{tabular}{@{}>{\raggedright\arraybackslash}p{0.115\textwidth}>{\centering\arraybackslash}p{0.111\textwidth}>{\centering\arraybackslash}p{0.092\textwidth}>{\centering\arraybackslash}p{0.108\textwidth}>{\centering\arraybackslash}p{0.104\textwidth}>{\centering\arraybackslash}p{0.092\textwidth}>{\centering\arraybackslash}p{0.108\textwidth}>{\centering\arraybackslash}p{0.115\textwidth}@{}}
\toprule
Model &
Acc@1 $\uparrow$ &
MCES $\downarrow$ &
$\mathrm{TaniSim}_{\mathrm{MG}} \uparrow$ &
$\mathrm{CosSim}_{\mathrm{MG}} \uparrow$ &
$\mathrm{TaniSim}_{\mathrm{MA}} \uparrow$ &
FraggleSim $\uparrow$ &
FGSim $\uparrow$ \\
\midrule
Gemini-3.1-Pro* & - & - & - & - & - & - & - \\
GPT-5.5* & 0.0563 & 5.7817 & 0.2330 & 0.3492 & 0.3603 & 0.4152 & 0.5059 \\
DiffSpectra & \underline{0.4056} & \underline{1.3273} & \underline{0.7879} & \underline{0.8456} & \underline{0.9260} & \textbf{0.9503} & \textbf{0.9657} \\
S1-Omni & \textbf{0.4569} & \textbf{1.1765} & \textbf{0.8227} & \textbf{0.8698} & \textbf{0.9322} & \underline{0.9436} & \underline{0.9574} \\
\bottomrule
\end{tabular}
\end{adjustbox}

\end{table}

Relative to DiffSpectra, S1-Omni raises Acc@1 by 0.0513 and lowers MCES by 0.1508. Morgan Tanimoto, Morgan cosine, and MACCS Tanimoto increase by 0.0348, 0.0242, and 0.0062. Acc@1 requires an exact match of the complete two-dimensional graph, while MCES and fingerprint similarity measure graph and local-topology agreement for the remaining predictions. The consistent gains indicate that S1-Omni exactly reconstructs more targets and produces candidates with closer global connectivity and fingerprint representations.

FraggleSim and FGSim decrease by 0.0067 and 0.0083, leaving DiffSpectra ahead in local fragment and functional-group similarity. Exact graph recovery, fingerprint overlap, and local fragment coverage assess different forms of structural agreement. S1-Omni improves exact and global reconstruction, while DiffSpectra better preserves specific local chemical motifs. Gemini-3.1-Pro does not produce a valid 3D SDF file; dashes in its row denote direct-generation failure rather than measured zero scores.

\subsubsection{Molecule Stability}
\label{sec:eval_4_2_2_molecule_stability}

Tables~\ref{tab:eval_4_2_2} and~\ref{tab:eval_4_2_3} evaluate S1-Omni in terms of 2D chemical quality and 3D geometric fidelity. The training-set statistics are provided only as references and are excluded from model ranking.

\begin{table*}[t!]
\centering
\caption{2D chemical stability, validity, uniqueness, novelty, and distribution quality of molecules generated on QM9S. All stability, validity, uniqueness, and novelty values are proportions. The best and second-best model results in each metric column are bold and underlined, respectively; the training-set reference is excluded from ranking.}
\label{tab:eval_4_2_2}
\scriptsize
\setlength{\tabcolsep}{3pt}
\begin{tabular*}{\textwidth}{@{\extracolsep{\fill}}l*{9}{c}@{}}
\toprule
2D-Metric
& AtomStable $\uparrow$
& MolStable $\uparrow$
& V\&C $\uparrow$
& V\&U $\uparrow$
& V\&U\&N $\uparrow$
& FCD $\downarrow$
& SNN $\uparrow$
& Frag $\uparrow$
& Scaf $\uparrow$ \\
\midrule
Train
& 0.9990
& 0.9880
& 0.9890
& 0.9890
& 0.0000
& 0.0630
& 0.4900
& 0.9920
& 0.9460 \\

JODO
& \underline{0.9990}
& \textbf{0.9880}
& \underline{0.9900}
& 0.9600
& 0.7800
& 0.1380
& 0.5220
& 0.9860
& 0.9340 \\

BioMatrix-1.7B
& \textbf{1.0000}
& \textbf{0.9880}
& \textbf{1.0000}
& 0.9480
& 0.7490
& \textbf{0.0640}
& 0.4950
& \underline{0.9910}
& \underline{0.9440} \\

BioMatrix-4B
& \textbf{1.0000}
& \textbf{0.9880}
& \textbf{1.0000}
& 0.9490
& 0.7420
& \underline{0.0660}
& 0.4940
& \underline{0.9910}
& 0.9420 \\

DiffSpectra
& 0.9983
& 0.9804
& 0.9819
& \underline{0.9674}
& \underline{0.8998}
& 0.0792
& \textbf{0.5314}
& \textbf{0.9921}
& \textbf{0.9448} \\

S1-Omni
& 0.9989
& \underline{0.9850}
& 0.9870
& \textbf{0.9782}
& \textbf{0.9105}
& 0.0920
& \underline{0.5283}
& \textbf{0.9921}
& 0.9426 \\
\bottomrule
\end{tabular*}
\end{table*}

\begin{table}[t!]
\centering
\caption{3D stability and distribution quality of molecules generated on QM9S. The best and second-best model results in each metric column are bold and underlined; the training-set reference is excluded from ranking.}
\label{tab:eval_4_2_3}
\scriptsize
\setlength{\tabcolsep}{3pt}
\begin{tabular*}{\textwidth}{@{\extracolsep{\fill}}l*{5}{c}@{}}
\toprule
3D-Metric
& AtomStable $\uparrow$
& Bond length $\downarrow$
& Bond angle $\downarrow$
& Dihedral angle $\downarrow$
& FCD $\downarrow$ \\
\midrule
Train
& 0.9940
& 5.44E-4
& 4.65E-4
& 1.78E-4
& 0.8770 \\

JODO
& \underline{0.9920}
& \textbf{1.48E-1}
& 1.21E-2
& 6.29E-4
& \textbf{0.8850} \\

BioMatrix-1.7B
& 0.8970
& 1.05E+0
& 3.16E-2
& 4.76E-4
& 1.0400 \\

BioMatrix-4B
& 0.8970
& 1.05E+0
& 3.09E-2
& 6.09E-4
& 1.0120 \\

DiffSpectra
& \textbf{0.9922}
& \underline{1.64E-1}
& \underline{7.30E-3}
& \textbf{2.45E-4}
& \underline{0.9218} \\

S1-Omni
& 0.9909
& 1.87E-1
& \textbf{6.13E-3}
& \underline{3.54E-4}
& 0.9541 \\

\bottomrule
\end{tabular*}
\end{table}

As shown in Table~\ref{tab:eval_4_2_2}, S1-Omni achieves the best V\&U and V\&U\&N scores among all evaluated models, reaching 0.9782 and 0.9105. Compared with the spectrum-to-molecule baseline DiffSpectra, S1-Omni improves MolStable from 0.9804 to 0.9850, V\&C from 0.9819 to 0.9870, V\&U from 0.9674 to 0.9782, and V\&U\&N from 0.8998 to 0.9105. It also improves AtomStable from 0.9983 to 0.9989. These consistent improvements demonstrate that S1-Omni generates a larger proportion of chemically stable molecules that are simultaneously valid, unique and novelty. In particular, its leading V\&U\&N result highlights its ability to avoid duplicated outputs and excessive overlap with the training set while preserving chemical validity.

S1-Omni also maintains competitive agreement with the reference distribution. It achieves the best Frag score of 0.9921, tied with DiffSpectra, and the second-best SNN score of 0.5283. Although its FCD and Scaf scores are slightly weaker than those of DiffSpectra (0.0920 vs 0.0792 and 0.9426 vs 0.9448), the differences in SNN, Frag, and Scaf remain small. Therefore, the gains in validity, uniqueness, and novelty do not result in a substantial departure from the reference chemical space. Compared with the unconditional baselines, S1-Omni improves V\&U\&N by 0.1305, 0.1615, and 0.1685 over JODO, BioMatrix-1.7B, and BioMatrix-4B.

The 3D results in Table~\ref{tab:eval_4_2_3} further demonstrate S1-Omni's ability to recover meaningful molecular geometry from spectral inputs. S1-Omni obtains the lowest bond-angle error among all evaluated models, achieving $6.13\times10^{-3}$, compared with $7.30\times10^{-3}$ for DiffSpectra and $1.21\times10^{-2}$ for JODO. This result indicates particularly accurate modeling of local angular geometry. S1-Omni also achieves an AtomStable score of 0.9909, remaining close to DiffSpectra 0.9922 and JODO 0.9920, while substantially outperforming both BioMatrix variants 0.8970. Its bond-length, dihedral-angle, and FCD results remain competitive, although DiffSpectra performs better on these metrics. These results support the effectiveness of S1-Omni for spectrum-conditioned molecular generation.

\subsubsection{Effect of Hidden-State Extraction}
\label{sec:eval_4_2_3_effect_of_hidden_state_extraction}

To assess whether molecular structural information is encoded in the VLM representation and can be effectively decoded, we train a 20-label multilabel probe on QM9S to predict molecular functional groups. The development set is randomly divided into training and validation subsets at a 95:5, and an independent set of 2,048 samples is used for evaluation. The probe is optimized using \texttt{BCEWithLogitsLoss}. Predictions are obtained using a fixed threshold of 0.5. Exact Match requires all 20 functional-group labels of a molecule to be predicted correctly. Micro-averaged metrics aggregate decisions across all molecule-label pairs, whereas Macro-averaged metrics assign equal weight to each functional-group class.

We compare four strategies for extracting representations from the IR image-token feature sequence. \texttt{IR-image\_avg} averages all image-token states, \texttt{IR-image\_end} selects the final image-token state, and \texttt{IR-image\_max\_min\_avg} concatenates the results of maximum, minimum, and average pooling. In contrast to these fixed aggregation strategies, \texttt{IR-image\_atten} uses learnable attention to adaptively aggregate the image-token sequence. This comparison examines whether functional-group information is more effectively preserved by fixed global summarization or by adaptive aggregation that selectively integrates token-level features and their context.

\begin{table}[t!]
\centering
\caption{20-label molecular-substructure probe using four IR-spectrum extraction strategies. Results use an independent 2,048-sample test set. Best and second-best results are bold and underlined, respectively.}\label{tab:eval_4_2_4}
\scriptsize
\setlength{\tabcolsep}{2pt}
\begin{adjustbox}{width=\textwidth}
\begin{tabular}{@{}>{\raggedright\arraybackslash}p{0.23\textwidth}*{5}{>{\centering\arraybackslash}p{0.134\textwidth}}@{}}
\toprule
Extraction strategy & Exact Match $\uparrow$ & Micro-P $\uparrow$ & Micro-F1 $\uparrow$ & Macro-P $\uparrow$ & Macro-F1 $\uparrow$ \\
\midrule
\texttt{IR-image\_avg} & 0.4833 & 0.8945 & 0.8725 & \underline{0.8188} & 0.7658 \\
\texttt{IR-image\_end} & \underline{0.5003} & \underline{0.9014} & \underline{0.8792} & 0.8151 & \underline{0.7761} \\
\texttt{IR-image\_max\_min\_avg} & 0.4007 & 0.8988 & 0.8404 & 0.7779 & 0.6343 \\
\texttt{IR-image\_atten} & \textbf{0.7292} & \textbf{0.9441} & \textbf{0.9392} & \textbf{0.9035} & \textbf{0.8976} \\
\bottomrule
\end{tabular}
\end{adjustbox}
\end{table}

As shown in Table~\ref{tab:eval_4_2_4}, \texttt{IR-image\_atten} achieves the best performance across all five reported metrics. Relative to \texttt{IR-image\_end}, the strongest fixed aggregation strategy overall, \texttt{IR-image\_atten} improves Exact Match by 22.89 percentage points, Micro-F1 by 6.00 points, and Macro-F1 by 12.15 points. In particular, the substantial improvement in Exact Match indicates that attention-based aggregation improves not only individual functional-group decisions but also the consistency of jointly predicting all 20 labels for a molecule.

Among the fixed aggregation strategies, \texttt{IR-image\_end} performs best overall and slightly outperforms \texttt{IR-image\_avg}, whereas \texttt{IR-image\_max\_min\_avg} maintains relatively high Micro precision but shows substantial declines in Exact Match, Micro-F1, and Macro-F1. This result suggests that concatenating additional global statistics does not necessarily preserve more functional-group information. In particular, extreme-value pooling may overemphasize isolated token responses without adequately capturing relationships across the image-token sequence. Moreover, the improvement of \texttt{IR-image\_atten} is larger in Macro-F1 than in Micro-F1, suggesting that its benefit is not driven solely by frequently occurring classes and extends to functional groups that are less well represented by fixed aggregation strategies.

These results indicate that functional-group evidence is distributed across the IR image-token sequence and can be decoded more effectively through adaptive attention-based aggregation than through fixed global summarization. By selectively integrating token-level features and their contextual relationships. This provides evidence that the VLM representation encodes decodable molecular structural information, while also showing that the choice of representation extraction strategy substantially affects how effectively that information can be recovered.

To compare the evidence provided by each spectral modality, we fix the sequence-attention extraction strategy and train the same 20-label QM9S probe separately on IR, Raman, and UV-Vis inputs. Table~\ref{tab:eval_4_2_5} reports Exact Match together with micro and macro precision and F1.

\begin{table}[t!]
\centering
\caption{Modality-specific results of the 20-label functional-group probe with sequence-attention feature extraction. Each row uses the listed spectrum modality as input. Best and second-best results are bold and underlined.}\label{tab:eval_4_2_5}
\scriptsize
\setlength{\tabcolsep}{2pt}
\begin{adjustbox}{width=\textwidth}
\begin{tabular}{@{}>{\raggedright\arraybackslash}p{0.14\textwidth}*{5}{>{\centering\arraybackslash}p{0.14\textwidth}}@{}}
\toprule
Modality  & Exact Match $\uparrow$ & Micro-P $\uparrow$ & Micro-F1 $\uparrow$ & Macro-P $\uparrow$ & Macro-F1 $\uparrow$ \\
\midrule
Raman & \underline{64.24\%} & \underline{92.09\%} & \underline{92.02\%} & \underline{87.99\%} & \underline{87.05\%} \\
UV & 21.80\% & 76.76\% & 74.94\% & 69.67\% & 66.52\% \\
IR & \textbf{72.92\%} & \textbf{94.41\%} & \textbf{93.92\%} & \textbf{90.35\%} & \textbf{89.76\%} \\
\bottomrule
\end{tabular}
\end{adjustbox}
\end{table}

IR attenuation achieves the highest scores across all five evaluation metrics, followed by Raman attenuation, with corresponding Micro-F1 values of 0.9392 and 0.9202 (Table~\ref{tab:eval_4_2_5}). By comparison, UV-Vis attenuation yields a Micro-F1 of 0.7494 and an Exact Match of only 0.2180, lagging noticeably behind the two vibrational spectroscopy modalities. The performance gap arises because the vibrational features captured by IR and Raman align more directly with the functional-group labels used in this task, while UV-Vis spectra predominantly encode electronic transition information. The consistent performance ranking of modalities across all metrics demonstrates that the proposed unified representation retains the unique scientific characteristics of each spectral type, rather than compressing them into undifferentiated generic spectral embeddings.

\subsection{Protein Site Prediction}
\label{sec:eval_4_3_protein_site_prediction}

Protein functional-site prediction identifies residues that mediate specific biological functions, including protein--protein, DNA, RNA, metal-ion, small-molecule, antibody, and antigen interactions. Accurate site prediction supports mechanistic interpretation, drug and antibody design, enzyme engineering, and disease research. We follow the metrics and comparison protocols of each public benchmark; because data splits and label definitions differ, scores are comparable only within the same subtask. Precision is the fraction of predicted sites that are correct, recall is the fraction of true sites recovered, F1 balances precision and recall, and MCC measures binary classification under class imbalance. AUROC and AUPR assess residue ranking across thresholds. All metrics are higher-is-better. Within a comparable table column, bold and underlined values denote the best and second-best results, respectively.

\subsubsection{Protein-Protein Interaction Site Prediction}
\label{sec:eval_4_3_1_protein_protein_interaction_site_prediction}

Protein--protein interaction (PPI) site prediction identifies residues on a target protein that directly contact another protein, yielding a residue-level description of a protein-complex interface.

\textbf{Setting.} We evaluate on the MPBind test set \citep{wang2025mpbind}. AUPR is the primary metric, AUROC provides a complementary threshold-independent ranking measure, and F1 and MCC assess residue classification at a fixed threshold. \model jointly encodes the target sequence and the PPI task description. Cross-attention allows every residue to retrieve context dynamically according to the task semantics before producing a site probability. The mean evaluated sequence length is 169 residues.

\textbf{Baselines.} Task-specific baselines are MPBind \citep{wang2025mpbind} and PeSTo \citep{krapp2023pesto}. GPT-5.5 and Gemini-3.1-Pro directly generate residue positions and are included only in the F1 and MCC comparisons.

\begin{table}[t!]
\centering
\caption{Residue-level PPI-site prediction on the MPBind test set. AUPR is the primary metric; higher values are better for all metrics. Best and second-best results are bold and underlined, respectively.}\label{tab:eval_4_3_1}
\scriptsize
\setlength{\tabcolsep}{2pt}
\begin{adjustbox}{width=\textwidth}
\begin{tabular}{@{}>{\raggedright\arraybackslash}p{0.331\textwidth}>{\centering\arraybackslash}p{0.128\textwidth}>{\centering\arraybackslash}p{0.128\textwidth}>{\centering\arraybackslash}p{0.152\textwidth}>{\centering\arraybackslash}p{0.141\textwidth}@{}}
\toprule
Method & F1 $\uparrow$ & MCC $\uparrow$ & AUROC $\uparrow$ & AUPR $\uparrow$ \\
\midrule
MPBind & - & - & \underline{0.830} & \underline{0.540} \\
PeSTo & - & - & 0.760 & 0.380 \\
GPT-5.5 & 0.284 & \underline{0.088} & - & - \\
Gemini-3.1-Pro & \underline{0.353} & 0.036 & - & - \\
S1-Omni & \textbf{0.632} & \textbf{0.513} & \textbf{0.859} & \textbf{0.666} \\
\bottomrule
\end{tabular}
\end{adjustbox}
\end{table}

The main gain appears in AUPR, which is especially sensitive to sparse positive residues. \model reaches 0.666 AUPR, 0.126 above MPBind, while AUROC rises from 0.830 to 0.859 (Table~\ref{tab:eval_4_3_1}). The concordant improvements show that the gain extends across both global positive--negative ranking and the precision--recall regime. F1 and MCC also exceed both general-purpose models, indicating that residue-level output remains effective after thresholding.

\subsubsection{Epitope Prediction}
\label{sec:eval_4_3_2_epitope_prediction}

Epitope prediction locates antigen-surface residues recognized and bound by an antibody, reversing the direction of paratope prediction on the antibody side.

\textbf{Setting.} We use the RoBep test set \citep{xu2026robep}. AUPR is the primary ranking metric, while F1 and MCC evaluate residue classification at a fixed threshold. \model forms a task-conditioned representation from the antigen sequence and epitope semantics, and a residue-level head predicts binding probabilities on the antigen. The mean evaluated sequence length is 253 residues.

\textbf{Baselines.} Specialized methods include RoBep \citep{xu2026robep}, GraphBepi \citep{zeng2023identifying}, SEMA 2.0 \citep{ivanisenko2024sema}, DiscoTope 3.0 \citep{hoie2024discotope}, CALIBER \citep{israeli2024single}, and SEPPA 3.0 \citep{zhou2019seppa}. We additionally compare with GPT-5.5 and Gemini-3.1-Pro.

\begin{table}[t!]
\centering
\caption{Residue-level epitope prediction on the RoBep test set. AUPR is the primary metric; higher values are better for all metrics. Best and second-best results are bold and underlined, respectively.}\label{tab:eval_4_3_2}
\scriptsize
\setlength{\tabcolsep}{2pt}
\begin{adjustbox}{width=\textwidth}
\begin{tabular}{@{}>{\raggedright\arraybackslash}p{0.405\textwidth}>{\centering\arraybackslash}p{0.157\textwidth}>{\centering\arraybackslash}p{0.157\textwidth}>{\centering\arraybackslash}p{0.172\textwidth}@{}}
\toprule
Method & F1 $\uparrow$ & MCC $\uparrow$ & AUPR $\uparrow$ \\
\midrule
RoBep & \underline{0.379} & \underline{0.340} & \underline{0.268} \\
GraphBepi & 0.276 & 0.189 & 0.215 \\
SEMA 2.0 & 0.284 & 0.196 & 0.236 \\
DiscoTope 3.0 & 0.300 & 0.234 & 0.237 \\
CALIBER & 0.241 & 0.164 & 0.158 \\
SEPPA 3.0 & 0.194 & 0.194 & 0.194 \\
GPT-5.5 & 0.209 & 0.076 & - \\
Gemini-3.1-Pro & 0.188 & 0.030 & - \\
S1-Omni & \textbf{0.546} & \textbf{0.493} & \textbf{0.472} \\
\bottomrule
\end{tabular}
\end{adjustbox}
\end{table}

\model obtains 0.472 AUPR, 0.204 above RoBep, with corresponding gains of 0.167 in F1 and 0.153 in MCC (Table~\ref{tab:eval_4_3_2}). Improvements in both ranking and threshold-based metrics show that the result is not tied to a particular operating point. In contrast to the PPI comparison, this benchmark supports both types of metric for the specialist models and thus more directly demonstrates that the same residue-level interface can be redirected toward antigen-side labels. Epitope recognition depends not only on local sequence patterns but also on immunological knowledge and global functional context. The biological knowledge encoded by \model helps it interpret protein functions and interaction semantics beyond what is available to conventional sequence-only baselines.

\subsubsection{Small-Molecule-Binding Site Prediction}
\label{sec:eval_4_3_5_small_molecule_binding_site_prediction}

Small-molecule-binding site prediction identifies protein residues that contact drugs, metabolites, or other small-molecule ligands.

\textbf{Setting.} We evaluate on COACH420 as curated by CLAPE-SMB \citep{wang2024protein}. AUROC is the primary ranking metric; precision, recall, F1, and MCC characterize residue classification at a fixed threshold. \model places the protein sequence and small-molecule ligand context in the same task condition and decodes a binding probability for each residue. The mean evaluated sequence length is 284 residues.

\textbf{Baselines.} The specialized baselines are P2Rank \citep{krivak2018p2rank} and GraphBind \citep{xia2021graphbind}, together with GPT-5.5 and Gemini-3.1-Pro as general-purpose references.

\begin{table}[t!]
\centering
\caption{Residue-level small-molecule-binding site prediction on COACH420. AUROC is the primary metric; higher values are better for all metrics. Best and second-best results are bold and underlined, respectively.}\label{tab:eval_4_3_5}
\scriptsize
\setlength{\tabcolsep}{2pt}
\begin{adjustbox}{width=\textwidth}
\begin{tabular}{@{}>{\raggedright\arraybackslash}p{0.263\textwidth}>{\centering\arraybackslash}p{0.151\textwidth}>{\centering\arraybackslash}p{0.129\textwidth}>{\centering\arraybackslash}p{0.102\textwidth}>{\centering\arraybackslash}p{0.102\textwidth}>{\centering\arraybackslash}p{0.121\textwidth}@{}}
\toprule
Method & Precision $\uparrow$ & Recall $\uparrow$ & F1 $\uparrow$ & MCC $\uparrow$ & AUROC $\uparrow$ \\
\midrule
P2Rank & 0.079 & \textbf{0.888} & - & 0.224 & - \\
GraphBind & \underline{0.223} & 0.477 & - & \underline{0.303} & \underline{0.870} \\
GPT-5.5 & 0.096 & 0.330 & \underline{0.148} & 0.101 & - \\
Gemini-3.1-Pro & 0.077 & 0.515 & 0.134 & 0.097 & - \\
S1-Omni & \textbf{0.275} & \underline{0.600} & \textbf{0.377} & \textbf{0.366} & \textbf{0.880} \\
\bottomrule
\end{tabular}
\end{adjustbox}
\end{table}

\model reaches 0.880 AUROC, 0.010 above GraphBind, and improves MCC from 0.303 to 0.366 (Table~\ref{tab:eval_4_3_5}). Relative to GraphBind, precision and recall rise jointly from 0.223 and 0.477 to 0.275 and 0.600, so the improvement is present in both ranking and threshold statistics. P2Rank attains the highest recall, 0.888, but with a precision of 0.079, reflecting a high-recall, low-precision operating regime.

\subsubsection{Metal-Binding Site Prediction}
\label{sec:eval_4_3_3_metal_binding_site_prediction}

Metal-binding site prediction identifies residues that coordinate a specified metal ion. Because coordination environments differ across ions, we report MCC for each ion and average MCC across ions.

\textbf{Setting.} We use LABind DS1 \citep{zhang2025labind}, which covers Ca, Mg, Mn, Zn, Fe(III), Fe(II), Cu, Na, and K. Each ion-specific column reports MCC for the corresponding residue-level binary task. \model jointly conditions on the protein sequence, ion identity, and task description and produces an ion-specific classification through the shared residue interface. The mean evaluated sequence length is 391 residues.

\textbf{Baselines.} Specialized methods are LigBind \citep{xia2023ligbind} and LABind \citep{zhang2025labind}; the general-purpose references are GPT-5.5 and Gemini-3.1-Pro.

\begin{table}[t!]
\centering
\caption{Metal-binding site prediction on LABind DS1. MCC is reported by ion and higher is better; average MCC summarizes cross-ion performance. Best and second-best results are bold and underlined, respectively.}\label{tab:eval_4_3_3}
\scriptsize
\setlength{\tabcolsep}{2pt}
\begin{adjustbox}{width=\textwidth}
\begin{tabular}{@{}>{\raggedright\arraybackslash}p{0.129\textwidth}>{\centering\arraybackslash}p{0.063\textwidth}>{\centering\arraybackslash}p{0.063\textwidth}>{\centering\arraybackslash}p{0.063\textwidth}>{\centering\arraybackslash}p{0.063\textwidth}>{\centering\arraybackslash}p{0.081\textwidth}>{\centering\arraybackslash}p{0.078\textwidth}>{\centering\arraybackslash}p{0.063\textwidth}>{\centering\arraybackslash}p{0.063\textwidth}>{\centering\arraybackslash}p{0.059\textwidth}>{\centering\arraybackslash}p{0.081\textwidth}@{}}
\toprule
Method & Ca $\uparrow$ & Mg $\uparrow$ & Mn $\uparrow$ & Zn $\uparrow$ & Fe(III) $\uparrow$ & Fe(II) $\uparrow$ & Cu $\uparrow$ & Na $\uparrow$ & K $\uparrow$ & Average MCC $\uparrow$ \\
\midrule
LigBind & \underline{0.465} & \underline{0.354} & \underline{0.608} & \underline{0.750} & \underline{0.679} & 0.678 & 0.552 & 0.166 & \underline{0.232} & \underline{0.498} \\
LABind & \textbf{0.664} & \textbf{0.461} & \textbf{0.710} & \textbf{0.843} & \textbf{0.768} & \textbf{0.848} & \textbf{0.771} & \textbf{0.387} & \textbf{0.444} & \textbf{0.655} \\
GPT-5.5 & 0.320 & 0.156 & 0.045 & 0.488 & 0.129 & 0.231 & 0.386 & 0.033 & -0.015 & 0.197 \\
Gemini-3.1-Pro & 0.313 & 0.183 & 0.255 & 0.508 & -0.010 & 0.507 & 0.514 & 0.001 & -0.008 & 0.251 \\
S1-Omni & 0.351 & 0.273 & 0.591 & 0.652 & 0.627 & \underline{0.712} & \underline{0.606} & \underline{0.167} & 0.224 & 0.467 \\
\bottomrule
\end{tabular}
\end{adjustbox}
\end{table}

\model reaches an average MCC of 0.467, 0.188 below LABind at 0.655, but above GPT-5.5 and Gemini-3.1-Pro (Table~\ref{tab:eval_4_3_3}). It exceeds LigBind on Fe(II), Cu, and Na and is close on Mn and K, whereas LABind leads for all nine ions. These ion-dependent differences show that the shared residue interface captures part of the conditioned coordination pattern. Specialist models, however, explicitly incorporate metal-specific chemical priors and structural features, whereas \model relies primarily on sequence and task semantics and therefore remains less effective at resolving fine-grained coordination rules.

\subsubsection{Paratope Prediction}
\label{sec:eval_4_3_4_paratope_prediction}

Paratope prediction locates residues in antibody variable regions that directly contact an antigen, defining the antibody side of the interface.

\textbf{Setting.} We evaluate on the MIPE test set \citep{wang2024improving}. F1 is the primary metric; precision and recall expose the fixed-threshold tradeoff, MCC complements them under class imbalance, and AUROC and AUPR measure ranking across thresholds. \model jointly encodes the antibody sequence and paratope semantics, allowing the shared prediction head to emit antibody-side binding residues according to the requested task direction. The mean evaluated sequence length is 77 residues.

\textbf{Baselines.} Specialized methods are MIPE \citep{wang2024improving}, ParaSurf \citep{papadopoulos2025parasurf}, and VASCIF \citep{liu2026interpretable}; we additionally compare with GPT-5.5 and Gemini-3.1-Pro.

\begin{table}[t!]
\centering
\caption{Residue-level paratope prediction on the MIPE test set. F1 is the primary metric; higher values are better for all metrics. Best and second-best results are bold and underlined, respectively.}\label{tab:eval_4_3_4}
\scriptsize
\setlength{\tabcolsep}{2pt}
\begin{adjustbox}{width=\textwidth}
\begin{tabular}{@{}>{\raggedright\arraybackslash}p{0.230\textwidth}>{\centering\arraybackslash}p{0.132\textwidth}>{\centering\arraybackslash}p{0.113\textwidth}>{\centering\arraybackslash}p{0.089\textwidth}>{\centering\arraybackslash}p{0.089\textwidth}>{\centering\arraybackslash}p{0.105\textwidth}>{\centering\arraybackslash}p{0.098\textwidth}@{}}
\toprule
Method & Precision $\uparrow$ & Recall $\uparrow$ & F1 $\uparrow$ & MCC $\uparrow$ & AUROC $\uparrow$ & AUPR $\uparrow$ \\
\midrule
MIPE & - & - & \underline{0.627} & \underline{0.554} & 0.927 & 0.741 \\
ParaSurf & \textbf{0.576} & \textbf{0.860} & \textbf{0.690} & \textbf{0.659} & \underline{0.967} & \textbf{0.781} \\
VASCIF & - & - & - & - & \textbf{0.981} & \underline{0.742} \\
GPT-5.5 & 0.302 & 0.625 & 0.407 & 0.138 & - & - \\
Gemini-3.1-Pro & 0.339 & \underline{0.745} & 0.466 & 0.240 & - & - \\
S1-Omni & \underline{0.456} & 0.649 & 0.536 & 0.361 & 0.757 & 0.479 \\
\bottomrule
\end{tabular}
\end{adjustbox}
\end{table}

\model attains 0.456 precision and 0.649 recall, yielding 0.536 F1, whereas ParaSurf reaches 0.690 F1 with higher precision and recall (Table~\ref{tab:eval_4_3_4}). Its AUROC and AUPR of 0.757 and 0.479 also trail MIPE and ParaSurf. Antibody--antigen recognition depends strongly on three-dimensional structure, conformational change, and spatial complementarity. Because \model primarily uses sequence and language semantics without explicit antibody structure, it does not reach the specialist results. It nevertheless exceeds GPT-5.5 and Gemini-3.1-Pro in F1 by 0.129 and 0.070, respectively, making it the strongest general-purpose model in this comparison.

\subsubsection{DNA-Binding Site Prediction}
\label{sec:eval_4_3_6_dna_binding_site_prediction}

DNA-binding site prediction identifies protein residues that contact DNA, characterizing nucleic-acid interfaces involved in transcriptional regulation, replication, and repair.

\textbf{Setting.} We use DNATest-129 from USPDB \citep{mi2025uspdb}. AUROC is the primary ranking metric; precision, recall, F1, and MCC describe fixed-threshold residue classification. \model uses the protein sequence and DNA-binding task semantics to produce per-residue probabilities, redirecting the shared interface toward nucleic-acid-binding labels. The mean evaluated sequence length is 291 residues.

\textbf{Baselines.} Specialized methods include MegSite \citep{hu2025megsite}, USPDB \citep{mi2025uspdb}, HGCTBind \citep{gong2026hgctbind}, and IPDLPre \citep{feng2025protein}, alongside GPT-5.5 and Gemini-3.1-Pro.

\begin{table}[t!]
\centering
\caption{Residue-level DNA-binding site prediction on DNATest-129. AUROC is the primary metric; higher values are better for all metrics. Best and second-best results are bold and underlined, respectively.}\label{tab:eval_4_3_6}
\scriptsize
\setlength{\tabcolsep}{2pt}
\begin{adjustbox}{width=\textwidth}
\begin{tabular}{@{}>{\raggedright\arraybackslash}p{0.263\textwidth}>{\centering\arraybackslash}p{0.151\textwidth}>{\centering\arraybackslash}p{0.129\textwidth}>{\centering\arraybackslash}p{0.102\textwidth}>{\centering\arraybackslash}p{0.102\textwidth}>{\centering\arraybackslash}p{0.121\textwidth}@{}}
\toprule
Method & Precision $\uparrow$ & Recall $\uparrow$ & F1 $\uparrow$ & MCC $\uparrow$ & AUROC $\uparrow$ \\
\midrule
MegSite & 0.484 & \textbf{0.739} & \textbf{0.585} & \textbf{0.567} & \textbf{0.952} \\
USPDB & \textbf{0.532} & 0.612 & 0.551 & 0.527 & \underline{0.948} \\
HGCTBind & 0.514 & \underline{0.654} & \underline{0.574} & \underline{0.549} & 0.945 \\
IPDLPre & \underline{0.531} & 0.514 & 0.522 & 0.522 & 0.914 \\
GPT-5.5 & 0.138 & 0.427 & 0.194 & 0.132 & - \\
Gemini-3.1-Pro & 0.090 & 0.495 & 0.160 & 0.099 & - \\
S1-Omni & 0.384 & 0.593 & 0.466 & 0.436 & 0.890 \\
\bottomrule
\end{tabular}
\end{adjustbox}
\end{table}

\model reaches 0.890 AUROC, below all four specialists and 0.062 below the best result from MegSite (Table~\ref{tab:eval_4_3_6}). At the fixed threshold, its precision, recall, and F1 are 0.384, 0.593, and 0.466, respectively, giving a comparatively balanced precision--recall profile. Recall exceeds IPDLPre, but F1 and MCC remain below every specialist. Ranking and threshold metrics thus give the same overall conclusion: specialized models are stronger on DNA-binding sites. \model nonetheless substantially exceeds both general-purpose baselines on precision, recall, F1, and MCC, demonstrating the benefit of directed residue-level decoding. The remaining gap is consistent with the joint dependence of DNA interfaces on sequence conservation, structural environment, and electrostatics, none of which is explicitly imposed as a task-specific prior.

\subsubsection{RNA-Binding Site Prediction}
\label{sec:eval_4_3_7_rna_binding_site_prediction}

RNA-binding site prediction identifies protein residues that contact RNA, including interfaces involved in RNA processing, transport, and translational regulation.

\textbf{Setting.} We evaluate on RNATest-117 from MegSite \citep{hu2025megsite}. AUROC is the primary ranking metric; precision, recall, F1, and MCC characterize classification at a fixed threshold. \model fuses the protein sequence with RNA-binding task semantics to form task-conditioned residue representations, from which the shared prediction head outputs RNA-binding probabilities. The mean evaluated sequence length is 319 residues.

\textbf{Baselines.} The specialized methods are SPLiNet \citep{ye2026splinet} and MegSite \citep{hu2025megsite}; GPT-5.5 and Gemini-3.1-Pro serve as general-purpose references.

\begin{table}[t!]
\centering
\caption{Residue-level RNA-binding site prediction on RNATest-117. AUROC is the primary metric; higher values are better for all metrics. Best and second-best results are bold and underlined, respectively.}\label{tab:eval_4_3_7}
\scriptsize
\setlength{\tabcolsep}{2pt}
\begin{adjustbox}{width=\textwidth}
\begin{tabular}{@{}>{\raggedright\arraybackslash}p{0.263\textwidth}>{\centering\arraybackslash}p{0.151\textwidth}>{\centering\arraybackslash}p{0.129\textwidth}>{\centering\arraybackslash}p{0.102\textwidth}>{\centering\arraybackslash}p{0.102\textwidth}>{\centering\arraybackslash}p{0.121\textwidth}@{}}
\toprule
Method & Precision $\uparrow$ & Recall $\uparrow$ & F1 $\uparrow$ & MCC $\uparrow$ & AUROC $\uparrow$ \\
\midrule
SPLiNet & \textbf{0.398} & \textbf{0.859} & \textbf{0.544} & \textbf{0.516} & \underline{0.842} \\
MegSite & \underline{0.356} & \underline{0.571} & \underline{0.439} & \underline{0.411} & \textbf{0.899} \\
GPT-5.5 & 0.110 & 0.247 & 0.152 & 0.090 & - \\
Gemini-3.1-Pro & 0.070 & 0.445 & 0.120 & 0.048 & - \\
S1-Omni & 0.277 & 0.419 & 0.333 & 0.294 & 0.807 \\
\bottomrule
\end{tabular}
\end{adjustbox}
\end{table}

\model obtains 0.807 AUROC, 0.092 below MegSite, and trails both specialists in precision, recall, F1, and MCC (Table~\ref{tab:eval_4_3_7}). Compared with the general-purpose models, \model has higher precision, F1, and MCC. Gemini-3.1-Pro has slightly higher recall, 0.445 versus 0.419, but only 0.070 precision and 0.120 F1, indicating that its recall is driven by many false positives. Taken together, the metal-, paratope-, DNA-, and RNA-binding results show that directed residue decoding substantially improves general-purpose modeling, but a clear specialist gap remains on tasks governed by three-dimensional structure and biophysical priors. Sequence and semantic representations alone do not fully encode the required spatial constraints and interactions.

\subsubsection{Effect of Residue-Level Decoding and Protein-Task Fusion}
\label{sec:eval_4_3_8_residue_decoding_fusion}

\textbf{Setting.} We isolate the effects of prediction granularity and protein--task fusion on MIPE paratope prediction using four progressively stronger designs. The first fine-tunes the VLM and directly generates residue indices. The second reads the final-token hidden state and applies a linear classifier. The third preserves per-residue VLM hidden states and trains a residue-level prediction head with weighted BCE. The full \model instead uses ESM2 residue embeddings, applies cross-attention to the VLM-encoded task context, and uses the same weighted BCE objective. All variants are evaluated by F1.

\begin{table}[t!]
\centering
\caption{Architectural progression for paratope prediction on the MIPE test set. The columns specify the prediction target, protein representation, task-conditioning mechanism, and treatment of class imbalance. Best and second-best F1 values are bold and underlined, respectively.}\label{tab:eval_4_3_8}
\scriptsize
\setlength{\tabcolsep}{2pt}
\begin{adjustbox}{width=\textwidth}
\begin{tabular}{@{}>{\raggedright\arraybackslash}p{0.14\textwidth}>{\raggedright\arraybackslash}p{0.18\textwidth}>{\raggedright\arraybackslash}p{0.18\textwidth}>{\raggedright\arraybackslash}p{0.18\textwidth}>{\raggedright\arraybackslash}p{0.14\textwidth}>{\centering\arraybackslash}p{0.08\textwidth}@{}}
\toprule
Method & Prediction target & Protein representation & Task conditioning & Imbalance handling & F1 $\uparrow$ \\
\midrule
VLM + SFT & Generated residue-index text & Autoregressive VLM output & Prompt context & None & $\approx 0$ \\
VLM + linear head & Sequence-level classification & Final-token VLM hidden state & Prompt context & None & $\approx 0$ \\
Pure VLM + Residue Prediction & Residue-level classification & Per-residue VLM hidden states & Context encoded within the VLM & Weighted BCE & \underline{$\approx 0.30$} \\
S1-Omni & Residue-level classification & ESM2 residue embeddings & Cross-attention to VLM task context & Weighted BCE & \textbf{0.536} \\
\bottomrule
\end{tabular}
\end{adjustbox}
\end{table}

Generated residue-index text and final-token classification both yield near-zero F1 (Table~\ref{tab:eval_4_3_8}). Free-form text and a single global state fail to preserve the positional correspondence required for residue labels, confirming that protein-site prediction is a fine-grained localization problem rather than conventional knowledge-based question answering. Per-residue VLM classification raises F1 to approximately 0.30: maintaining a distinct representation at each sequence position converts the task semantics into a trainable dense prediction problem, although the resulting site information remains incomplete. Adding ESM2 residue representations and task-context cross-attention increases F1 to 0.536, a further gain of approximately 0.236. ESM2 preserves residue-level relations in the protein sequence, while cross-attention lets those representations retrieve context relevant to the current task.

The progression supports three conclusions. First, text generation alone cannot deliver precise residue-level localization. Second, protein-site prediction is inherently a dense residue-level task, and per-residue prediction is substantially more effective than sequence-level classification. Third, separating protein representation learning from task interpretation and coupling them through cross-attention is central to a unified multi-task protein-site predictor.

\subsection{Protein Structure Prediction}
\label{sec:eval_4_4_protein_structure_prediction}

\subsubsection{Experimental Setup}
\label{sec:eval_4_4_1_experiment_setup}

\textbf{Setting.} We follow the SimpleFold evaluation protocol on CAMEO22 \citep{haas2018continuous,wang2025simplefold}. TM-score \citep{zhang2004scoring} and GDT-TS measure global fold and topological agreement. lDDT \citep{mariani2013lddt} and lDDT-Ca evaluate all-atom and backbone local geometry, respectively, while RMSD measures coordinate deviation after structural alignment. We report both the mean and median of every metric to distinguish aggregate behavior from structural quality on a typical target. Higher values are better except for RMSD. \model uses S1-VL to jointly encode the protein sequence and task instruction, with reasoning supervision producing a reasoning-conditioned hidden representation. A lightweight projector maps this representation into the structural conditioning space, after which a fixed-capacity SimpleFold-700M decoder generates all-atom coordinates. This design connects scientific reasoning representations to decoding for domain-specific tasks while keeping the structural backbone and evaluation protocol fixed.

Protein structure prediction reconstructs three-dimensional atomic coordinates from an amino acid sequence and must jointly resolve local residue geometry, long-range contacts, and global fold topology. We study a controlled question: with the geometric decoder fixed, can a task representation shaped by scientific reasoning improve structure prediction? We select SimpleFold \citep{wang2025simplefold} because its streamlined architecture, explicit conditioning interface, and family of larger model variants make it suitable for isolating the effect of reasoning integration. All \model experiments and controlled variants use SimpleFold-700M, so differences primarily reflect the interaction between the reasoning condition and structural decoding.

\textbf{Baselines.} The central controlled baseline is SimpleFold-700M without external conditioning under the same evaluation protocol, which directly measures the effect of reasoning conditioning. MSA-based methods include RoseTTAFold \citep{baek2021accurate}, AlphaFlow \citep{jing2024alphafold}, AlphaFold2 \citep{jumper2021highly}, and RoseTTAFold2 \citep{baek2023efficient}. Single-sequence or protein-language-model methods include ESM3 \citep{hayes2025simulating}, ESMDiff \citep{lu2025structure}, EigenFold \citep{jing2023eigenfold}, OmegaFold \citep{wu2022high}, ESMFlow \citep{jing2024alphafold}, and ESMFold \citep{lin2023evolutionary}. Together, these methods establish the performance range of major structure-prediction paradigms on CAMEO22. Gemini-3.1-Pro and GPT-5.5 are directly prompted general-purpose models and report only TM-score.

The central comparison in Table~\ref{tab:eval_4_4_1} is between unconditioned SimpleFold-700M and \model with a reasoning-conditioned hidden representation. Because the geometric backbone is fixed, their difference directly measures the effect of the reasoning condition on global topology, local geometry, and coordinate alignment. The other methods provide broader reference points for structure-prediction performance.

\subsubsection{Main Results}
\label{sec:eval_4_4_2_main_results}

\begin{table}[t!]
\centering
\caption{Protein structure prediction on CAMEO22. Each metric is reported as the mean / median across test targets, with the two statistics ranked separately. Best and second-best results are bold and underlined, respectively.}\label{tab:eval_4_4_1}
\scriptsize
\setlength{\tabcolsep}{2pt}
\begin{adjustbox}{width=\textwidth}
\begin{tabular}{@{}>{\raggedright\arraybackslash}p{0.109\textwidth}>{\raggedright\arraybackslash}p{0.135\textwidth}>{\centering\arraybackslash}p{0.128\textwidth}>{\centering\arraybackslash}p{0.123\textwidth}>{\centering\arraybackslash}p{0.118\textwidth}>{\centering\arraybackslash}p{0.125\textwidth}>{\centering\arraybackslash}p{0.118\textwidth}@{}}
\toprule
Model family & Method & TM-score (mean / median) $\uparrow$ & GDT-TS (mean / median) $\uparrow$ & lDDT (mean / median) $\uparrow$ & lDDT-Ca (mean / median) $\uparrow$ & RMSD (mean / median) $\downarrow$ \\
\midrule
\multirow{4}{=}{MSA-based} & RoseTTAFold & 0.7800 / 0.8600 & 0.7150 / 0.7750 & 0.5750 / 0.6050 & 0.7980 / 0.8270 & 5.7210 / 2.8640 \\
& AlphaFlow & 0.8400 / 0.9270 & 0.8080 / 0.8530 & 0.7410 / 0.7980 & 0.8550 / 0.8930 & 3.8460 / 2.1220 \\
& AlphaFold2 & \underline{0.8630 / 0.9420} & \underline{0.8440 / 0.9030} & \textbf{0.8160 / 0.8560} & \textbf{0.8930} / \underline{0.9230} & \underline{3.5780 / 1.8570} \\
& RoseTTAFold2 & \textbf{0.8640 / 0.9470} & \textbf{0.8450 / 0.9040} & 0.7270 / 0.7670 & \textbf{0.8930 / 0.9260} & \textbf{3.5710 / 1.7070} \\
\multirow{6}{=}{PLM-based} & ESM3 & 0.7460 / 0.8400 & 0.6940 / 0.7580 & - & - & - \\
& ESMDiff & 0.7540 / 0.8470 & 0.7010 / 0.7600 & - & - & - \\
& EigenFold & 0.7500 / 0.8400 & 0.7100 / 0.7900 & - & - & - \\
& OmegaFold & 0.8050 / 0.8990 & 0.7670 / 0.8440 & 0.7460 / 0.8150 & 0.8290 / 0.8920 & 5.2940 / 2.6220 \\
& ESMFlow & 0.8180 / 0.8930 & 0.7740 / 0.8320 & 0.6960 / 0.7450 & 0.8270 / 0.8670 & 4.5280 / 2.6930 \\
& ESMFold & 0.8530 / 0.9330 & 0.8260 / 0.8750 & \underline{0.7920 / 0.8340} & \underline{0.8710} / 0.9060 & 3.9730 / 2.0190 \\
\multirow{2}{=}{General-purpose LLM} & Gemini-3.1-Pro & 0.3175 / 0.2490 & - & - & - & - \\
& GPT-5.5 & 0.1118 / 0.0820 & - & - & - & - \\
Internal baseline & SimpleFold-700M, no condition & 0.8288 / 0.9140 & 0.7850 / 0.8470 & 0.7724 / 0.8070 & 0.8480 / 0.8830 & 4.5571 / 2.6300 \\
Ours & S1-Omni & 0.8291 / 0.9090 & 0.7848 / 0.8430 & 0.7709 / 0.8040 & 0.8461 / 0.8800 & 4.5913 / 2.5040 \\
\bottomrule
\end{tabular}
\end{adjustbox}
\end{table}

\textbf{SimpleFold provides a scalable backbone for reasoning integration.} Results from AlphaFold2, RoseTTAFold2, and ESMFold show that greater capacity, MSAs, pair representations, and finer-grained geometric modeling continue to raise the ceiling of structural accuracy. We use SimpleFold-700M to isolate reasoning conditioning in a fixed, structurally transparent backbone. The low TM-scores of directly prompted general-purpose models further show that three-dimensional coordinates require a specialized decoder that enforces domain-specific geometric constraints. The present 700M experiment establishes the compatibility of reasoning representations with structure generation and provides a direct path to test the same interaction with larger SimpleFold variants when computational resources permit.

\textbf{Reasoning conditioning produces metric-specific structural gains.} Relative to unconditioned SimpleFold-700M, \model raises mean TM-score from 0.828760 to 0.829126, an increase of 0.000366, and lowers median RMSD from 2.630 to 2.504, an improvement of 0.126. Changes in mean GDT-TS, lDDT, and lDDT-Ca are each below 0.002, while mean RMSD rises from 4.557120 to 4.591322. The gains are therefore concentrated in global topology and coordinate alignment on the median target; local geometry and mean RMSD remain near the baseline. Overall, \model preserves the principal folding ability of SimpleFold-700M while improving mean TM-score and median RMSD, indicating that the reasoning representation supplies complementary structural information to the decoding process.

\subsubsection{Effect of Reasoning and Structural Conditioning}
\label{sec:eval_4_4_3_effect_of_reasoning_and_structural_conditioning}

To verify that these gains arise from information usable by the structural decoder, we compare reasoning hidden states, explicit secondary-structure conditions, and the scope of parameter updates. All variants use SimpleFold-700M on CAMEO22 and are evaluated against the same model without external conditioning. The comparison varies ground-truth versus predicted eight-state secondary structure, the VLM checkpoint and hidden-state extraction position, and projector-only versus projector-plus-diffusion training.

The \texttt{Oracle\ DSSP\ one-hot} condition is computed from the target structure and supplies the eight-state DSSP assignment as a one-hot input. The \texttt{predicted\ SS8\ hard\ labels} condition instead uses discrete secondary-structure predictions. Continuous conditions compare the S1-VL \texttt{prefill-last} representation, extracted after the protein sequence and instruction have been prefilled but before answer generation, with the S1-VL \texttt{im-end-last} representation extracted at the end-of-message token. Both continuous variants freeze SimpleFold-700M and train only the projector. The final variant jointly updates the projector and diffusion module to test whether a broader trainable parameter set is beneficial.

\begin{table}[t!]
\centering
\caption{Ablation of the conditioning source, hidden-state extraction, and trainable-module scope on CAMEO22. All metrics are averaged across test targets. Best and second-best results are bold and underlined, respectively.}\label{tab:eval_4_4_2}
\scriptsize
\setlength{\tabcolsep}{2pt}
\begin{adjustbox}{width=\textwidth}
\begin{tabular}{@{}>{\raggedright\arraybackslash}p{0.165\textwidth}>{\raggedright\arraybackslash}p{0.168\textwidth}>{\centering\arraybackslash}p{0.113\textwidth}>{\centering\arraybackslash}p{0.105\textwidth}>{\centering\arraybackslash}p{0.097\textwidth}>{\centering\arraybackslash}p{0.109\textwidth}>{\centering\arraybackslash}p{0.097\textwidth}@{}}
\toprule
Comparison axis & Variant & TM-score $\uparrow$ & GDT-TS $\uparrow$ & lDDT $\uparrow$ & lDDT-Ca $\uparrow$ & RMSD $\downarrow$ \\
\midrule
Reference setting & SimpleFold-700M without external condition & 0.8288 & 0.7850 & \underline{0.7724} & \underline{0.8480} & \underline{4.5571} \\
Ground-truth secondary structure & Oracle DSSP eight-state one-hot condition & \textbf{0.8318} & \textbf{0.7898} & 0.6782 & \textbf{0.8512} & \textbf{4.4885} \\
Predicted secondary structure & Predicted SS8 hard-label condition & 0.8252 & 0.7811 & 0.7700 & 0.8457 & 4.5998 \\
Hidden-state extraction & S1-VL, last prefill hidden state & \underline{0.8298} & 0.7852 & \textbf{0.7725} & 0.8469 & 4.5799 \\
Hidden-state extraction & S1-VL, end-of-message hidden state & 0.8294 & \underline{0.7855} & 0.7716 & 0.8471 & 4.5671 \\
Trainable-module scope & Jointly train projector and diffusion module & 0.8279 & 0.7829 & 0.7658 & 0.8442 & 4.6428 \\
\bottomrule
\end{tabular}
\end{adjustbox}
\end{table}

\textbf{Effect of ground-truth secondary-structure conditioning.} Oracle DSSP raises TM-score from 0.828760 to 0.831836 and GDT-TS from 0.785044 to 0.789831, while reducing RMSD from 4.557120 to 4.488475 (Table~\ref{tab:eval_4_4_2}). At the same time, lDDT falls to 0.678164. The condition improves global topology and coordinate alignment but has the opposite effect on all-atom local distance accuracy, showing that eight-state secondary structure affects global folding and local atomic geometry differently in this setting.

\textbf{Effect of hidden-state extraction.} Relative to unconditioned SimpleFold-700M, S1-VL prefill-last raises TM-score by 0.001016, GDT-TS by 0.000196, and lDDT by 0.000087. S1-VL im-end-last raises TM-score by 0.000601 and GDT-TS by 0.000464. Both extraction positions improve TM-score and GDT-TS, while exhibiting different tradeoffs in local geometry and RMSD. The consistent direction of the global metrics indicates that reasoning-conditioned hidden states convey task information that SimpleFold-700M can exploit for folding.

\textbf{Effect of predicted secondary-structure conditioning.} Predicted SS8 hard labels yield lower TM-score, GDT-TS, lDDT, and lDDT-Ca than the unconditioned reference, together with higher RMSD. The divergence between Oracle DSSP and predicted SS8 shows that discrete secondary-structure conditioning depends on the fidelity of the labels to the target structure. Here, predicted labels provide weaker geometric constraints than ground-truth DSSP. An effective reasoning condition must therefore preserve scientific relations consistent with the structural evidence before a geometric decoder can use it.

\textbf{Effect of trainable-module scope.} Jointly training the projector and diffusion module produces lower values on all five metrics than either projector-only hidden-state variant. Projector-only training concentrates adaptation on the mapping from the shared hidden state to the structural conditioning space while preserving the pretrained geometric parameters of SimpleFold-700M. The controlled results favor this lightweight adaptation strategy.

Across the main comparison and conditioning experiments, reasoning conditioning improves mean TM-score with a fixed SimpleFold-700M backbone, and both hidden-state extraction positions consistently improve TM-score and GDT-TS. Oracle DSSP further confirms that the conditioning pathway can use structure-related information to improve global folding. The central conclusion is that task states formed through scientific reasoning can enter a geometric decoder through a lightweight interface and positively affect structure prediction. With a larger computational budget, the same fusion mechanism can be evaluated on larger SimpleFold variants or stronger structural backbones to determine whether the gain scales with decoder capacity.

\subsection{Scientific Image Generation and Editing}
\label{sec:eval_4_5_scientific_image_generation_and_editing}

\subsubsection{Experimental Setup}
\label{sec:eval_4_5_1_experiment_setup}

Scientific image generation synthesizes research illustrations, mechanism diagrams, technical workflows, and text-dense visual explanations from language instructions. Scientific image editing produces a modified image from an input image and an instruction. We evaluate both the organization of scientific semantics in generated images and pixel-space operations including segmentation, modality translation, super-resolution, and multi-step local editing.

\textbf{Setting.} Image generation is evaluated on three complementary benchmarks. GenExam \citep{GenExam2025} covers multidisciplinary exam-style scientific illustrations; GPT-5 judges semantic criteria, spelling, readability, and logical consistency to produce a weighted Relaxed Score. TechImage-Bench \citep{Ni2025TechImageBench} covers biological, engineering, and general technical diagrams and uses o4-mini to evaluate hierarchical binary rubrics. CVTG-2K \citep{evalTai2025TextInsulation} evaluates complex multi-region text rendering with OCR-based Word Accuracy. Image editing covers three operation classes. MSD \citep{antonelli2022medical} and cigRockSEM \citep{zhang2025benchmark} evaluate medical-image and rock-SEM segmentation; both predictions and ground truth are converted to binary masks by RGB thresholding before Dice or F1 is computed. SynthRAD2025 \citep{thummerer2025synthrad2025} and BCI \citep{liu2022bci} evaluate CBCT-to-CT and H\&E-to-IHC translation, respectively, while IXI \citep{ixiDataset} evaluates MRI super-resolution. All three reconstruction tasks use PSNR: SynthRAD2025 follows the official protocol in the HU domain within the anatomical region, whereas BCI and IXI use the standard 8-bit image domain. Volumetric data are processed as two-dimensional slices. 

\textbf{Baselines.} General-purpose image models are Qwen-Image \citep{Wu2025QwenImage}, GLM-Image \citep{glmimage2026}, GPT-Image-2 \citep{chatgptimages2026}, and Nano Banana 2 (\texttt{gemini-3.1-flash-image-preview}) \citep{raisinghani2026nano}. Each editing dataset also has an independent task-specific reference: Swin UNETR for MSD \citep{tang2022self}, U-Net for cigRockSEM \citep{zhang2025benchmark}, VBoussot for SynthRAD2025 \citep{boussot2025registration}, Pyramid Pix2pix for BCI \citep{liu2022bci}, and CodeBrain for IXI \citep{wu2026virtual}. These are different models rather than a single specialist evaluated across all five columns.

\subsubsection{Scientific Image Generation}
\label{sec:eval_4_5_2_scientific_image_generation}

\begin{table}[t!]
\centering
\caption{Scientific image generation benchmark results. All metrics are reported on a 0--100 scale; higher is better. Best and second-best results are bold and underlined, respectively.}\label{tab:eval_4_5_1}
\scriptsize
\setlength{\tabcolsep}{2pt}
\renewcommand{\arraystretch}{0.92}
\begin{adjustbox}{width=\textwidth}
\begin{tabular}{@{}>{\raggedright\arraybackslash}p{0.244\textwidth}>{\centering\arraybackslash}p{0.203\textwidth}>{\centering\arraybackslash}p{0.243\textwidth}>{\centering\arraybackslash}p{0.203\textwidth}@{}}
\toprule
Image generation model & GenExam Relaxed Score $\uparrow$ & TechImage-Bench rubric accuracy $\uparrow$ & CVTG-2K Word Accuracy $\uparrow$ \\
\midrule
Qwen-Image & 23.8 & 51.11 & 72.38 \\
GLM-Image & 46.1 & 54.47 & \textbf{91.16} \\
GPT-Image-2 & \textbf{93.8} & \textbf{83.87} & 77.94 \\
Nano Banana 2 & \underline{92.6} & \underline{78.38} & 77.88 \\
S1-Omni & 51.4 & 60.68 & \underline{80.39} \\
\bottomrule
\end{tabular}
\end{adjustbox}
\end{table}

S1-Omni exceeds both open baselines on GenExam and TechImage-Bench (Table~\ref{tab:eval_4_5_1}). Relative to Qwen-Image, the gains are 27.6 and 9.57 points; relative to GLM-Image, they are 5.3 and 6.21 points. GPT-Image-2 and Nano Banana 2 remain substantially stronger on both rubric-based scientific illustration benchmarks.

CVTG-2K produces a different ranking. S1-Omni scores 80.39, above Qwen-Image, GPT-Image-2, and Nano Banana 2, while GLM-Image leads at 91.16. GenExam and TechImage-Bench jointly assess scientific entities, relations, layout, and labels; CVTG-2K isolates character rendering. The results therefore separate content organization from typography: S1-Omni is the stronger open model on structured scientific content, whereas character-level fidelity remains an independent bottleneck.

S1-Omni and Qwen-Image share the same image-generation backbone. Their gap on GenExam and TechImage-Bench therefore shows that the task-conditioned hidden representation provides the decoder with constraints on scientific entities, relations, and layout. The clearer partitioning and directional relations in Appendix Figure~\ref{fig:eval_4_5_1} qualitatively complement the rubric results, whereas the remaining label errors are consistent with the character-level rendering gap measured by CVTG-2K.

\subsubsection{Scientific Image Editing}
\label{sec:eval_4_5_3_scientific_image_editing}

\begin{table}[t!]
\centering
\caption{Scientific image editing benchmark results. MSD and cigRockSEM evaluate segmentation with Dice and F1, respectively; SynthRAD2025 evaluates CBCT-to-CT translation, BCI evaluates H\&E-to-IHC translation, and IXI evaluates 2$\times$/4$\times$ MRI super-resolution with PSNR in dB. Higher is better for all metrics; best and second-best results are bold and underlined, respectively.}\label{tab:eval_4_5_2}
\scriptsize
\setlength{\tabcolsep}{2pt}
\renewcommand{\arraystretch}{0.92}
\begin{adjustbox}{width=\textwidth}
\begin{tabular}{@{}>{\raggedright\arraybackslash}p{0.158\textwidth}>{\centering\arraybackslash}p{0.131\textwidth}>{\centering\arraybackslash}p{0.145\textwidth}>{\centering\arraybackslash}p{0.158\textwidth}>{\centering\arraybackslash}p{0.131\textwidth}>{\centering\arraybackslash}p{0.145\textwidth}@{}}
\toprule
Model & \makecell[c]{MSD\\Dice $\uparrow$} & \makecell[c]{cigRockSEM\\F1 $\uparrow$} & \makecell[c]{SynthRAD2025\\PSNR (dB) $\uparrow$} & \makecell[c]{BCI\\PSNR (dB) $\uparrow$} & \makecell[c]{IXI MRI SR\\PSNR (dB) $\uparrow$} \\
\midrule
Per-dataset specialist & \underline{0.7868} & \underline{0.9733} & 34.78 & \textbf{21.16} & 24.67 \\
Qwen-Image & 0.0396 & 0.9210 & \underline{42.03} & 0.00 & 14.25 \\
GLM-Image & 0.0442 & 0.6394 & 35.00 & 12.74 & 11.66 \\
GPT-Image-2 & 0.3787 & 0.9001 & 40.49 & 14.30 & 19.96 \\
Nano Banana 2 & 0.3263 & 0.8962 & 40.08 & 14.33 & \underline{25.27} \\
S1-Omni & \textbf{0.8496} & \textbf{0.9823} & \textbf{47.87} & \underline{20.38} & \textbf{26.72} \\
\bottomrule
\end{tabular}
\end{adjustbox}
\end{table}

Under the two-dimensional editing protocol, S1-Omni achieves the best performance on four of the five scientific image editing benchmarks: MSD, cigRockSEM, SynthRAD2025, and IXI (Table~\ref{tab:eval_4_5_2}). On MSD, S1-Omni obtains a Dice score of 0.8496, outperforming the task-specific Swin UNETR specialist by 0.0628. On cigRockSEM, it reaches an F1 score of 0.9823, exceeding the dedicated U-Net baseline by 0.0090. For SynthRAD2025 and IXI, S1-Omni achieves PSNR values of 47.87 dB and 26.72 dB, surpassing the corresponding specialist models, VBoussot and CodeBrain, by 13.09 dB and 2.05 dB, respectively. On BCI, S1-Omni ranks second with 20.38 dB, while the pathology-specific Pyramid Pix2pix model retains a 0.78 dB advantage.

The segmentation results highlight the difficulty general-purpose image editing models face in performing spatially precise scientific editing. Segmentation-as-editing requires the model to translate a language-defined target into an exact spatial region, render the requested mask or color overlay consistently, and preserve the surrounding background. Boundary localization, mask rendering, color encoding, and background preservation therefore jointly determine performance. On MSD, all general-purpose models remain substantially below S1-Omni, with Dice scores ranging from 0.0396 to 0.3787. On cigRockSEM, S1-Omni also consistently outperforms both the specialist U-Net baseline and all general-purpose models. These results show that a shared image-editing decoder can perform accurate region-level operations across both medical and materials-science images, while Dice and F1 jointly characterize localization accuracy and mask-rendering quality.

Image translation and super-resolution reveal more fine-grained differences between the unified model and task-specific specialists. On SynthRAD2025, S1-Omni achieves 47.87 dB PSNR, outperforming the specialist VBoussot model by 13.09 dB and exceeding the strongest general-purpose baseline by 5.84 dB. This result indicates that S1-Omni can accurately reconstruct the target computed tomography pixel distribution from cone-beam computed tomography inputs under the slice-based evaluation protocol. On IXI, S1-Omni reaches 26.72 dB PSNR, outperforming CodeBrain by 2.05 dB. The smaller margin suggests that MRI super-resolution remains relatively close to the task-specific specialist regime, with Nano Banana 2 also obtaining a competitive result of 25.27 dB.

On BCI, Pyramid Pix2pix achieves 21.16 dB, leading S1-Omni by 0.78 dB and retaining a task-specific advantage in hematoxylin-and-eosin-to-immunohistochemistry stain translation. Nevertheless, S1-Omni substantially outperforms all evaluated general-purpose image models on this benchmark. The result suggests that highly specialized pathological appearance mappings can still benefit from architectures and training objectives tailored to a single stain-translation setting. Although PSNR measures pixel-level fidelity on SynthRAD2025, BCI, and IXI, high PSNR alone does not guarantee anatomical, pathological, or clinical validity, which should be further evaluated using corresponding domain-specific criteria.

Overall, S1-Omni supports a broad range of scientific image editing tasks, including segmentation, cross-modality image translation, and super-resolution. Across medical and materials-science benchmarks, it demonstrates strong adaptability to heterogeneous transformation objectives, ranging from spatially precise region delineation to modality conversion and fine-detail reconstruction. These results show that a unified scientific multimodal model can effectively handle diverse image editing settings, while task-specific metrics provide complementary views of its performance under different application requirements.

\subsection{Aggregate Comparison with General-Purpose Models}
\label{sec:eval_4_6_general_model_comparison}

\begin{figure}[t]
  \centering
  \includegraphics[width=\textwidth]{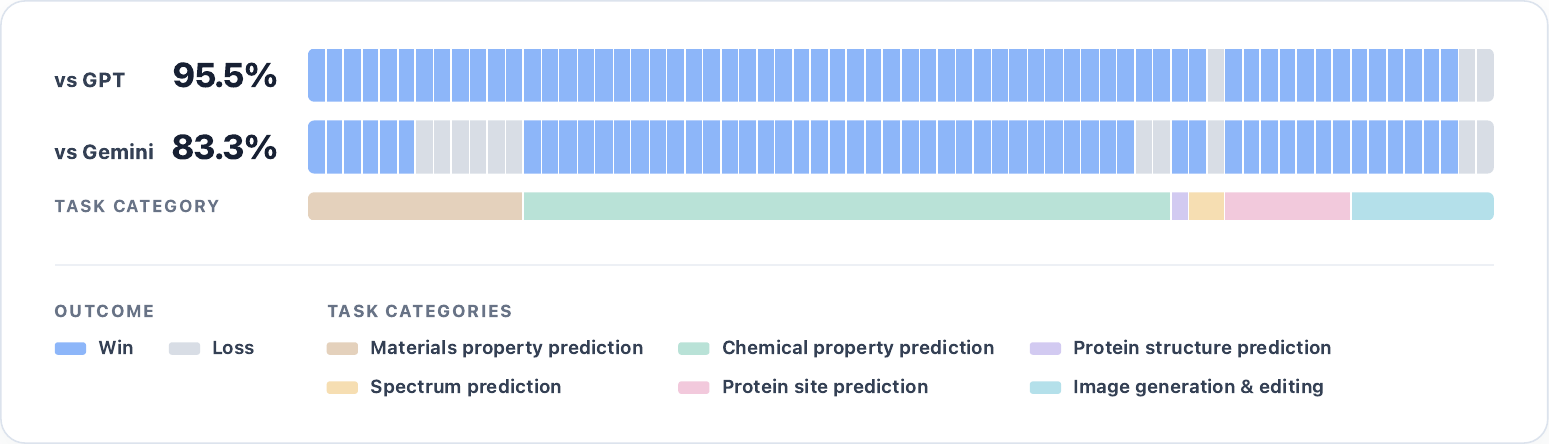}
  \caption{Aggregate task-level comparison with general-purpose model families under the shared text-and-image input protocol. The outcome strips summarize wins and losses for \model, and the lower strip shows the composition of the six scientific task categories.}
  \label{fig:discussion_general_model_comparison}
\end{figure}

To consolidate the general-purpose model comparisons reported throughout Section~\ref{sec:evaluations}, we conduct a task-level analysis under a common input protocol. The full evaluation contains 66 tasks across six categories: materials property prediction, chemical property prediction, spectrum prediction, protein structure prediction, protein site prediction, and scientific image generation and editing. For comparability, every system receives only text and images, including textual serializations of scientific objects where required. Tasks evaluated through textual or structured predictions use GPT-5.5~\citep{singh2025openai} and Gemini-3.1-Pro~\citep{evalGeminiTeam2024}, whereas image-output tasks use GPT-Image-2~\citep{chatgptimages2026} and Nano Banana 2~\citep{raisinghani2026nano}.

Across the eligible comparisons, \model attains win rates of 95.5\% against the GPT model family and 83.3\% against the Gemini model family (Figure~\ref{fig:discussion_general_model_comparison}). The advantage extends across heterogeneous prediction and generation settings rather than being determined by a single benchmark family. These aggregate rates count task outcomes rather than performance margins and should therefore be interpreted together with the per-benchmark results in Section~\ref{sec:evaluations}; nevertheless, their consistency across scalar, residue-level, structural, and pixel-space outputs provides evidence that \model is the strongest unified system in this comparison.

This comparison also clarifies where scientific specialization matters. General-purpose models provide a flexible language-and-vision interface, but many scientific tasks are evaluated in constrained output spaces whose validity depends on domain structure: calibrated scalar predictions, residue sets, molecular graphs, three-dimensional coordinates, segmentation masks, and reconstructed images. \model couples a shared reasoning backbone with decoders for domain-specific tasks that operate directly in these spaces, allowing scientific evidence to be translated into outputs that respect the corresponding metric, geometry, or spatial constraint. The results therefore support specialization at the output interface as an effective complement to unified multimodal reasoning: a common backbone enables transfer across domains, while task-aligned output spaces supply the inductive biases needed for reliable scientific prediction and generation.
\FloatBarrier

\fussy
\FloatBarrier

\section{Discussion}
\label{sec:discussion}

\subsection{Reasoning Supervision Improves Scientific Representations}
\label{sec:discussion_reasoning_supervision}

Across property prediction, protein structure prediction, and scientific image generation, reasoning supervision improves the representations supplied to downstream modules, but the construction of the reasoning data is decisive. On six material regression properties, structured, property-constrained supervision reduces MAE from 34.818 with unconstrained free-form reasoning to 17.255 and raises MAD/MAE from 0.888 to 1.260 (Table~\ref{tab:eval_4_1_4}). It also outperforms the no-reasoning condition, whose MAE and MAD/MAE are 20.457 and 0.941, respectively. Because the Stage-2 predictor, data scale, optimization, and question-only pooling are fixed, these differences isolate the organization of the Stage-1 reasoning supervision rather than predictor capacity or answer-side extraction.

Appendix Sections~\ref{sec:case_reasoning_unconstrained} and~\ref{sec:case_reasoning_structured} illustrate the distinction on the same Lu$_2$ZnPd crystal and Spillage request. The unconstrained trace identifies potentially relevant ingredients, including heavy-element content, spin--orbit coupling, orbital character, and lattice symmetry, but it is free to choose its own coverage and analytical order. The structured trace first records a common set of structural and compositional observations, then selects a smaller property-relevant subset and states the competing directions explicitly: cell simplicity and orbital localization are treated as downward factors, whereas heavy-element-induced spin--orbit coupling acts upward. The important difference is therefore not trace length or the presence of scientific terminology, but whether the supervision exposes a consistent mapping from observed evidence to property-specific factors, their direction, and their interaction.

The case comparison is illustrative; the controlled ablation provides the aggregate evidence. Together, they show why fluent but unconstrained rationales can be a poor training target: variation in evidence coverage and causal ordering introduces nuisance variation into the training signal received by the shared backbone, causing unconstrained supervision to perform worse than omitting reasoning text altogether. Property-constrained construction reduces this variation by imposing a stable analytical schema while still requiring the teacher to select task-relevant factors. This result supports structured reasoning as a representation-learning objective, but does not imply that a template guarantees factual correctness; scientific claims within the traces still require data validation and domain review.

The protein experiments provide a controlled geometric test. SimpleFold-700M is fixed and only its conditioning representation is changed; reasoning-conditioned states produce small improvements in global folding metrics. This result is modest, but it establishes that a language-model task state can carry information usable by a geometric decoder without altering the decoder backbone.

Appendix Figure~\ref{fig:discussion_think_before_generate} makes the same mechanism visible in image generation. Without think-before-generate planning, a long instruction must be mapped directly to an image, and the model often fails to coordinate module grouping, hierarchy, arrows, labels, and global layout. Explicit reasoning first decomposes the request into executable visual units, such as input regions, the backbone, VTP, VVP, PIR, predictions, and the legend, and then specifies their positions, colors, connections, and information flow. The resulting scientific semantics, task intent, and layout plan are injected into MMDiT through reasoning-diffusion alignment. The decoder is therefore conditioned on an organized task state rather than the raw prompt alone, improving module completeness, topology, hierarchy, and readability.

Together, these results suggest that the principal value of reasoning supervision is representational. It teaches the shared backbone to form task states that remain useful to numerical predictors, molecular and protein structure generators, and image decoders, even when the downstream module does not consume the generated reasoning text.

\subsection{Prefill Representations as a Stable Downstream Interface}
\label{sec:discussion_prefill_extraction}

A central design choice in decoding for domain-specific tasks is how to transfer a shared VLM representation to heterogeneous scientific modules. One option extracts the final answer or route-token state and compresses generation into a single vector. A second pools the chain-of-thought and answer states. A third uses question-side prefill representations formed from the scientific input and task instruction before autoregressive decoding.

The experiments show a consistent pattern in feature extraction from the shared backbone: task information required by domain decoders is largely present before answer generation, and pooling the full reasoning trajectory gives no stable advantage. In property prediction, question-only pooling improves both classification and regression relative to pooling reasoning and answer tokens (Table~\ref{tab:eval_4_1_5}). In protein structure prediction, prefill-last and end-of-message states both provide useful conditions, but post-generation states are not consistently better (Table~\ref{tab:eval_4_4_2}). In the spectrum-to-molecular generation task, evidence distributed along the input sequence is best retrieved by attention over input-side latent tokens, without waiting for a chain of thought or answer (Table~\ref{tab:eval_4_2_4}).

Long reasoning states mix scientific evidence with linguistic organization, repeated explanation, autoregressive history, and output formatting; final answer states are additionally shaped by termination prediction. Pooling them can dilute a task signal already formed in the prefill. Question-side states, by contrast, are determined directly by the scientific object and requested task. A practical division of labor follows: reasoning supervision shapes the shared scientific representation during training, whereas domain decoders preferentially extract features from prefill states at inference, using pooling or attention only when evidence is distributed across input positions.

The present architecture still passes conditions predominantly in one direction, from the shared VLM to each domain decoder. Deeper bidirectional fusion between shared and domain-specific representations remains an important direction for unified scientific reasoning.

\subsection{Backbone Contributions Under Strong Expert Decoders}
\label{sec:discussion_backbone_vs_experts}

A natural question is whether the gains in unified scientific modeling primarily come from stronger task experts or from the shared backbone representation. Our analysis experiments are designed to isolate this question by reusing two strong, task-specific decoders---a spectrum-to-molecular generation expert and a protein-folding expert---and changing only the conditioning representation supplied by the shared backbone. Under this controlled reuse of expert modules, improvements can be attributed to the backbone states rather than to decoder redesign.

The effect is clearest for spectrum-to-molecular generation. With a spectrum-conditioned expert decoder in place, \model achieves a substantially higher exact structure match rate than the specialist baseline (Table~\ref{tab:eval_4_2_1}), indicating that the shared backbone encodes spectral evidence in a form that is easier for the expert to convert into a valid molecular structure.

For protein folding, the gains are smaller but still directionally positive. When SimpleFold-700M is held fixed, reasoning-conditioned backbone representations yield modest improvements in selected global and alignment metrics (Table~\ref{tab:eval_4_4_1}), suggesting that the shared backbone can inject useful structural signals even when the downstream geometric prior is already strong.

Overall, these controlled results support the view that a meaningful fraction of \model's benefit comes from backbone representations that enhance downstream experts, rather than from task decoders alone.

\subsection{General Conversational Competence Under Scientific Specialization}
\label{sec:discussion_multiturn_dialogue}

We construct a general multi-turn dialogue evaluation to examine whether science-oriented training weakens general conversational ability and whether the model can switch tasks reliably within a long context. The conversation begins with non-specialized requests, including a self-introduction and the generation of a mathematics lesson plan, before introducing a protein PPI-site prediction task in the same context (Appendix Section~\ref{sec:case_multiturn}). We use the identical protein sequence and task instruction as an independent single-turn control. This setting tests whether the model can track a multi-turn context, isolate history unrelated to the current task, and preserve the scientific task protocol and output format when switching from general-purpose generation to scientific prediction.

The multi-turn and single-turn settings produce exactly the same final PPI-site indices, indicating that the preceding general conversation does not alter the scientific prediction in this example. After generating a self-introduction and educational content, the model recognizes the protein sequence, recovers the task representation required for site prediction, and returns the same structured answer as the single-turn control. This result suggests that science-oriented training does not eliminate the model's general multi-turn capability: \model can retain context understanding, task switching, and scientific output consistency across heterogeneous turns. Longer and more complex interactions require broader evaluation, but this case provides initial evidence that scientific specialization can coexist with general conversational competence.

\section{Limitations}
\label{sec:limitations}

We view \model as an initial step toward scalable unified multimodal reasoning for scientific understanding, prediction, and generation. The present work establishes a continuous modeling process linking heterogeneous scientific data, shared task representations, and domain-native results, and tests its feasibility on representative tasks. Several limitations become more important as the model moves toward larger and more open scientific settings.

\textbf{Large-scale scientific pretraining.}
\model has not undergone large-scale scientific pretraining. It starts from an existing vision-language model and is trained on curated scientific tasks, a setting that isolates unified task protocols and module integration but does not reveal scaling laws with data, parameters, or compute. Future work should extend the current task system to continual scientific pretraining at larger scale and measure how scientific representations, cross-task transfer, and domain capabilities change.

\textbf{Modular generation across scientific output spaces.}
Molecular structures, protein coordinates, and scientific images are not generated natively by the shared backbone; they rely on diffusion or specialist decoders. This design preserves geometric accuracy and generation quality in each output space, but remains an intermediate point between unified task representation and unified native generation. Systematic comparisons of token, diffusion, and hybrid approaches are needed to determine how much structural representation and generation can move into the shared backbone without losing the inductive biases required by continuous spaces.

\textbf{Task coverage.}
The current evaluation uses property prediction, structure reconstruction, protein tasks, and scientific images as initial test beds for connecting different input and output spaces. It does not yet provide sufficient disciplinary depth, complex multistage problems, out-of-distribution generalization across task families, or cross-task transfer. Broader tasks and more systematic transfer studies are required to test whether existing capabilities can be composed to solve genuinely new scientific problems.

\textbf{General capabilities and long-horizon tasks.}
Scientific task training reduces performance on some general tasks relative to the base model, and the present evaluation concentrates on bounded single-turn problems. Multi-turn planning, cross-modal interaction, long-horizon state tracking, tool use, and correction across steps remain underexplored. Future training should treat retention of general capabilities and scientific performance as joint objectives, while evaluating planning, execution, and feedback-driven revision in long-horizon research workflows.

\section{Conclusion}
\label{sec:conclusion}

In this work, we present \model, a unified multimodal reasoning model for scientific understanding, prediction, and generation trained on the \sonecorpus. Through unified representation of scientific data, natural-world knowledge alignment, and decoding for domain-specific tasks, \model links heterogeneous scientific inputs, evidence-grounded reasoning, and native scientific outputs within a single model, offering a scalable path towards unified multimodal reasoning for science. More broadly, we view the central challenge of scientific AI systems as threefold: identifying the scientific evidence from which models should learn, determining the reusable knowledge and capabilities they should acquire, and distilling these into scalable, efficient systems that generalize across domains and tasks while remaining scientifically reliable and verifiable. Future work will explore continual scientific pretraining, structure-aware generation, out-of-distribution generalization, and tool-augmented research workflows, with the goal of building scalable, verifiable, and collaborative scientific AI.

\section*{Author Contributions}

Core contributors and contributors are listed alphabetically by given name within each group.

\textbf{Core Contributors.}
Jiahao Zhao, Junyi Liu, Lifeng Xu, Nan Xu, Qingli Wang, Qingxiao Li, Tianle Chen, Xiaoyu Wu, Yawen Zheng, Zikai Wang

\textbf{Contributors.}
Guanming Liu, Hequn Zhou, Jingyi Wang, Jingyuan Shu, Keqi Wang, Li He, Songyang Diao, Wenhui Xu, Xinyu Ren, Yaqin Fan, Yujin Zhou, Zhanao Yao

\FloatBarrier
\normalcolor

\bibliographystyle{colm2024_conference}

\bibliography{reference}

\appendix
\clearpage
\FloatBarrier
\section{Appendix}
\label{sec:case}

This section first presents thirteen complete reasoning examples using the same task-family order as Section~\ref{sec:evaluations}. It then provides targeted visual analyses of scientific image generation and editing, examines representative failure modes, and evaluates whether scientific prediction remains stable after a long, general-purpose multi-turn conversation. The original language of every interaction is retained.

\subsection{Reasoning Examples}

Each example starts on a new page and shows the full user input, the complete model response, and the emitted task token. Native numerical, structural, or image output is included when it is available in the source record. The two controlled reasoning-supervision examples report no native scalar output, so none is inferred here. Table~\ref{tab:case_catalog} provides a compact overview.

\begin{table}[h]
\centering
\caption{Case catalogue ordered by the task families used in the Evaluation section. Each route token selects the corresponding native result interface.}
\label{tab:case_catalog}
\scriptsize
\begin{tabularx}{\textwidth}{@{}>{\raggedright\arraybackslash}p{0.14\textwidth}>{\raggedright\arraybackslash}p{0.29\textwidth}>{\raggedright\arraybackslash}p{0.16\textwidth}>{\raggedright\arraybackslash}p{0.13\textwidth}>{\raggedright\arraybackslash}X@{}}
\toprule
Task family & Case & Input & Route token & Native output \\
\midrule
Property & DFPT dielectric constant & Crystal CIF & \RouteToken{<linear_pre>} & Scalar: 30.15 \\
Property & Exfoliation energy & Crystal CIF & \RouteToken{<linear_pre>} & Scalar: 49.97 meV/atom \\
Property & Topological spillage & Crystal CIF & \RouteToken{<linear_pre>} & Scalar: 0.215 \\
Property & Spillage: unconstrained free-form reasoning & Crystal CIF & \RouteToken{<linear_pre>} & Not provided in source \\
Property & Spillage: structured, property-constrained reasoning & Crystal CIF & \RouteToken{<linear_pre>} & Not provided in source \\
Property & ESOL aqueous solubility & Molecular SMILES & \RouteToken{<linear_pre>} & Scalar: $-3.2656$ log S \\
Property & CYP2C9 substrate & Molecular SMILES & \RouteToken{<linear_cla>} & Non-substrate \\
Property & CYP2D6 inhibition & Molecular SMILES & \RouteToken{<linear_cla>} & Non-inhibitor \\
Spectrum & IR spectrum-to-molecular generation & Infrared spectrum & \RouteToken{<spectra_st>} & Cyclopropane nitrile structure \\
Protein site & ArcA DNA-binding site & Amino-acid sequence & \RouteToken{<prot_cla>} & Residue indices \\
Protein structure & GPR168 structure & Amino-acid sequence & \RouteToken{<prot_st>} & All-atom coordinates \\
Image generation & Chemical synapse diagram & Text instruction & \RouteToken{<image_gen>} & Scientific image \\
Image editing & Tumor and edema segmentation & FLAIR MRI + instruction & \RouteToken{<image_edit>} & Spatial overlay \\
\bottomrule
\end{tabularx}
\end{table}

\begingroup
\let\subsection\subsubsection
\clearpage
\color{black}
\subsection{Crystal Property Prediction: DFPT Dielectric Constant}
\label{sec:case_material_dielectric}

\begin{CaseBlock}{User Input}
下面给出了一个晶体结构的 CIF 文件内容：

\CaseInputFileCompact{assets/case_study/pr2alno3.cif}

请给出其对应的 总介电常数 $\varepsilon$（DFPT）（无量纲）精确值。
\end{CaseBlock}

\begin{CaseBlockCompact}{Model Response}
\setlength{\parskip}{0.05em}
\setlength{\smallskipamount}{1pt}
\CaseProtocolMarker{<think>}

首先对提供的 CIF 文件进行结构定量表征。该材料化学式为 Pr2AlNO3，结晶于三斜晶系，空间群为 P 1（第 1 号），这是对称性最低的晶系，意味着原子位置几乎没有对称性约束。晶格参数显示 a 与 b 相等均为 3.752 Å，c 轴较长为 6.875 Å，角度 α 和 β 均为 105.837°，γ 为 90.000°，单胞体积约为 89.3 Å³，每个晶胞包含 1 个化学式单元。从组成特征来看，材料含有重元素稀土镨（Pr），其原子序数较大且具有 f 电子，同时包含轻金属铝（Al）及非金属氮（N）和氧（O）。平均原子质量较重，电负性差异显著，尤其是 Pr 与 O/N 之间。局域结构方面，由于空间群为 P1，所有原子均处于不等价位点，配位环境复杂，Pr 原子可能形成高配位数的多面体，而 Al 可能与 N/O 形成较硬的四面体或八面体连接。结构拓扑上，较小的晶胞体积与较多的原子数（7 个原子/晶胞）表明结构堆积较为致密，不存在明显的层状弱连接特征，整体呈现三维框架结构。

接着筛选影响总介电常数的关键结构特征。第一是稀土元素 Pr 的存在，Pr 具有较大的离子半径和丰富的 f 轨道电子，这会显著增加电子极化率并可能通过软模机制增强离子极化率，对介电常数有强烈的正向提升作用。第二是低对称性的三斜结构，P1 空间群允许更多的声学支和光学支振动模式耦合，降低了限制离子位移的对称性壁垒，有利于提高离子对介电常数的贡献。第三是较高的结构致密性，89.3 Å³ 的体积容纳 7 个原子意味着较高的数密度，通常高密度与较高的电子介电常数相关联，但过强的键合也可能限制离子位移，这是一个竞争因素。

然后进行多因素竞争分析。虽然 Al-N/O 键通常具有较强的共价性且刚度较大，倾向于限制晶格畸变从而降低离子介电贡献，但 Pr 元素的引入极大地补偿了这一点。Pr-O/N 键的离子性较强且 Pr 离子本身易极化，主导了整体的极化行为。低对称性进一步释放了晶格弛豫的潜力，使得离子贡献不会被完全抑制。因此，稀土极化效应与低对称性效应共同主导，克服了轻元素骨架可能带来的刚性限制，预计总介电常数将高于普通简单氧化物。

最后进行定性逻辑推导。一般简单二元氧化物的静态介电常数通常处于基础水平。考虑到该材料含有两个高极化率的 Pr 原子，基础值应向上修正。再结合三斜晶系低对称性带来的额外离子贡献，数值有望进一步攀升。然而，由于晶胞体积较小且含有刚性的 Al 组分，介电常数不太可能达到铁电体或超高介电常数材料的水平。综合评估，上述结构与机理因素共同支撑最终预测任务。

\CaseProtocolMarker{</think>}

材料分子式为 Pr2AlNO3，目标预测性质为总介电常数 $\varepsilon$（DFPT），单位为无量纲。该晶体属于三斜晶系，空间群为 P 1，晶胞体积约为 89.3 Å³，包含稀土元素镨以及铝、氮、氧组分，背景条件涉及低频极化与电子云畸变的共同贡献。

由于稀土镨元素的高极化率显著增强了电子与离子贡献，且三斜低对称性结构允许更多的晶格弛豫模式，导致介电响应处于中等偏高水平；结合较高的原子堆积密度进一步限制了极端低值出现的可能性，表明其为具有良好介电性能的潜在介质材料。接下来，我将完成 \texttt{eps\_dfpt} 的最终预测任务。

\CaseProtocolMarker{<linear_pre>}
\end{CaseBlockCompact}

\CaseCompactNativeResult{总介电常数 $\varepsilon$}{30.15}{DFPT，无量纲}

\clearpage
\color{black}
\subsection{Crystal Property Prediction: Exfoliation Energy}
\label{sec:case_material_exfoliation}

\begin{CaseBlock}{User Input}
Given the CIF file of a crystal structure:

\CaseInputFileCompact{assets/case_study/ta2cs2.cif}

Please estimate the exfoliation energy (Exf. En., meV/atom) of this material.
\end{CaseBlock}

\begin{CaseBlockCompact}{Model Response}
\setlength{\parskip}{0.05em}
\setlength{\smallskipamount}{1pt}
\CaseProtocolMarker{<think>}

First, the crystal structure of Ta2CS2 is examined based on the provided CIF data, revealing a triclinic symmetry (P 1) with highly symmetric lattice parameters where a=b=c=8.833 Å and all angles are acute at 21.505°. This specific geometric configuration, despite the low symmetry label, suggests a rhombohedral stacking sequence often associated with layered transition metal compounds. The unit cell contains five atoms, comprising two tantalum atoms, one carbon atom, and two sulfur atoms, indicating a dense packing with a volume of approximately 81.15 Å³.

Next, the chemical composition is analyzed to understand the bonding nature. Tantalum is a heavy transition metal capable of forming strong metallic and covalent bonds, while sulfur is a chalcogen known for facilitating van der Waals gaps in dichalcogenides. The inclusion of carbon introduces strong covalent character, likely strengthening the intralayer connectivity compared to pure tantalum sulfide. The atomic positions aligned along the body diagonal further support a layered motif where sulfur atoms likely terminate the slabs, creating potential cleavage planes.

Then, the key factors influencing exfoliation energy are weighed. The primary driver for low exfoliation energy is the presence of sulfur terminations which typically result in weak interlayer van der Waals forces. However, the carbon atoms may act as interlayer bridges or intralayer strengtheners, potentially increasing the energy required to separate layers compared to standard MX2 compounds. The competition between the weak S-S interlayer interactions and the strong Ta-C intralayer bonds determines the final value.

Finally, qualitative logic is applied based on these structural insights. Typical layered transition metal dichalcogenides exhibit characteristic exfoliation energies. Given the reinforcing effect of carbon in this specific Ta2CS2 stoichiometry, the value is expected to be slightly elevated but still within the regime of easily exfoliable materials. Therefore, the energy is predicted to be consistent with a stable yet separable layered structure.

\CaseProtocolMarker{</think>}

The material Ta2CS2 is evaluated for its Exfoliation Energy (Exf. En.), expressed in meV/atom. This property is critical for assessing the feasibility of isolating single layers from the bulk crystal for nanotechnology applications.

Exfoliation Energy (Exf. En.) is a key metric for layered materials. Driven by the weak van der Waals interactions between sulfur-terminated layers but moderated by the strong covalent Ta-C bonding within the slabs, the energy reflects the balance of interlayer and intralayer forces. This suggests the material possesses a layered character that allows for mechanical exfoliation while maintaining structural integrity. Next, I will predict the \texttt{exf\_en} value.

\CaseProtocolMarker{<linear_pre>}
\end{CaseBlockCompact}

\CaseCompactNativeResult{Predicted exfoliation energy}{49.97}{meV/atom}

\clearpage
\color{black}
\subsection{Crystal Property Prediction: Topological Spillage}
\label{sec:case_material_spillage}

\begin{CaseBlock}{User Input}
根据给定的 CIF 文件内容：

\CaseInputFileCompact{assets/case_study/vse2.cif}

请输出该晶体结构的 Spillage（无量纲）精确值。
\end{CaseBlock}

\begin{CaseBlock}{Model Response}
\CaseProtocolMarker{<think>}

首先，从提供的 CIF 文件解析出该材料为二硒化钒（VSe2），晶体结构呈现典型的层状过渡金属硫族化合物特征。晶格参数显示 a=b=3.372 Å，c=6.201 Å，γ=120°，表明其具有六方晶系的对称性背景，尽管空间群标记为 P1，这通常是原胞表示。化学组成包含钒（V，3d 过渡金属）和硒（Se，第四周期硫族元素），单胞内含 1 个 V 原子和 2 个 Se 原子，体积约为 61.07 Å³。这种结构通常由 V 原子层夹在两层 Se 原子层之间形成三明治状的单层结构，并通过范德华力堆叠。

接着，分析影响 Spillage 值的关键物理因素。Spillage 主要衡量能带反转的稳健性及拓扑特性的强弱，其大小强烈依赖于自旋轨道耦合（SOC）强度以及费米能级附近的轨道杂化情况。在此材料中，Se 元素的原子序数 Z=34，提供了中等强度的自旋轨道耦合作用，相较于重元素碲化物（如 Bi2Te3）较弱，但比硫化物强。同时，V 的 3d 轨道与 Se 的 4p 轨道之间存在显著的杂化，这决定了能带的色散关系和潜在的能带反转位置。

然后，考虑多因素竞争对最终数值的影响。一方面，层状结构和特定的配位环境（V 处于 Se 八面体或三棱柱中心）有利于形成特定的能带序，可能诱发轻微的能带反转；另一方面，VSe2 在常温下通常表现为金属性或半金属性，费米面穿过能带，这与典型拓扑绝缘体所需的体能隙有所区别，从而限制了 Spillage 值的上限。此外，P1 空间群的低对称性表示可能会解除部分简并，进一步影响拓扑不变量的计算结果，使得 Spillage 值不会非常高。

最后，进行定性逻辑推导。基于过渡金属硫族化合物的经验数据，非强拓扑体系的 Spillage 值通常远小于 1。考虑到 Se 的 SOC 贡献以及 VSe2 的电子结构特征，该值应处于较低水平，但非零。参照同类 3d 金属二硒化物的物理性质，综合结构对称性与元素组成的加权影响，反映出其具有微弱或非平庸的拓扑特征，但并非强拓扑绝缘体。

\CaseProtocolMarker{</think>}

针对二硒化钒（VSe2）晶体结构，本次预测的目标性质为 Spillage，该物理量为无量纲参数。作为一种层状过渡金属硫族化合物，其电子结构特性受到晶格对称性与组分元素原子序数的共同制约，背景条件涉及 3d 过渡金属与第四周期硫族元素的轨道相互作用。

Spillage 值主要受硒元素的中等强度自旋轨道耦合效应与钒 - 硒轨道杂化程度的控制，由于材料呈现金属性特征且缺乏强拓扑保护机制，这些因素导致 Spillage 数值维持在较低水平。结合结构分析中识别出的层状堆积模式与相对较轻的重元素成分，表明该材料具有微弱的拓扑相关性或接近平凡绝缘体/金属的边缘态特征。接下来对 spillage 进行性质预测。

\CaseProtocolMarker{<linear_pre>}
\end{CaseBlock}

\CaseNativeOutputHeading
\CaseResultCard{Spillage}{0.215}{无量纲}

\clearpage
\color{black}
\subsection{Reasoning-Supervision Comparison: Unconstrained Free-Form Reasoning}
\label{sec:case_reasoning_unconstrained}

\begin{CaseBlock}{Experimental Setting}
This is the unconstrained condition in the Stage-1 reasoning-supervision ablation (Table~\ref{tab:eval_4_1_4}). The teacher receives a broad request for scientific analysis, while the choice of evidence, reasoning order, and level of detail remains unconstrained. The Stage-2 predictor, training scale, optimization, and question-only pooling are held fixed across the comparison.
\end{CaseBlock}

\begin{CaseBlock}{User Input}
Given the CIF file of a crystal structure:

\CaseInputFileCompact{assets/case_study/lu2znpd.cif}

Please infer a reasonable Spillage (dimensionless) estimation for this structure.
\end{CaseBlock}

\begin{CaseBlock}{Model Response}
\CaseProtocolMarker{<think>}

First, examining the provided CIF file reveals the material is Lu2ZnPd, crystallizing in the triclinic space group P1 with lattice parameters a, b, and c all approximately 4.87 Angstroms and angles alpha, beta, and gamma all close to 60 degrees. Although the symmetry is labeled as P1, the metric tensor suggests a highly symmetric underlying rhombohedral or pseudo-cubic lattice, with a unit cell volume of roughly 81.6 cubic Angstroms and one formula unit per cell. The atomic positions are highly ordered, occupying fractional coordinates at 0, 0.25, 0.5, and 0.75, which indicates a structured intermetallic phase rather than a disordered alloy.

Next, considering the chemical composition, the system contains Lutetium (Z=71), Palladium (Z=46), and Zinc (Z=30). The presence of Lutetium and Palladium introduces significant atomic mass and potential for spin-orbit coupling (SOC) effects, which are the primary drivers for the Spillage metric. However, Lutetium typically exists in a trivalent state with a filled 4f shell (4f14), meaning the f-electrons are core-like and do not actively participate in the band crossing or inversion near the Fermi level. Consequently, the relevant states near the Fermi energy are likely dominated by Pd 4d and Zn 4s/4p orbitals, which exhibit moderate but not extreme SOC strength compared to active f-electron systems.

Then, analyzing the competition between these factors, the heavy element content suggests a non-zero Spillage value, as SOC will inevitably mix some wavefunction characters. However, the high degree of structural order and the effective high symmetry of the lattice constrain the extent of this mixing, preventing large-scale wavefunction reorganization. Furthermore, the closed-shell nature of the Lu ions reduces the contribution of the strongest SOC source in the compound to the valence band physics. This balance implies that while the Spillage will be measurable, it will remain on the lower end of the possible spectrum for heavy-element compounds.

Finally, converging on a numerical estimate, the effective range for Spillage is defined between 0 and 4. For stable intermetallics with heavy elements but closed f-shells and high structural order, values typically cluster in the lower quartile. Given the specific combination of moderate SOC active species (Pd) and structural rigidity, the Spillage is expected to be small but distinct from zero. A reasonable prediction interval would be between 0.05 and 0.25, reflecting minor wavefunction perturbation due to spin-orbit interactions without strong topological reconstruction.

\CaseProtocolMarker{</think>}

The crystal structure Lu2ZnPd is evaluated to predict its Spillage value, a dimensionless metric quantifying wavefunction changes induced by spin-orbit coupling, with an effective prediction range of [0, 4]. The analysis integrates the material's stoichiometry, heavy element content, and crystallographic symmetry to assess the magnitude of electronic perturbation.

Although the presence of heavy Lutetium and Palladium atoms provides a mechanism for spin-orbit interaction, the closed-shell electronic configuration of Lutetium and the highly ordered, pseudo-high-symmetry lattice structure limit the extent of band mixing. These factors collectively suppress large wavefunction variations, leading to a qualitative expectation of low topological perturbation. Consequently, the Spillage value is estimated to fall within the interval of 0.05 to 0.25, indicating a stable electronic structure with minimal spin-orbit induced reconstruction.

Proceed to infer the spillage value.

\CaseProtocolMarker{<linear_pre>}
\end{CaseBlock}

\clearpage
\color{black}
\subsection{Reasoning-Supervision Comparison: Structured, Property-Constrained Reasoning}
\label{sec:case_reasoning_structured}

\begin{CaseBlock}{Experimental Setting}
This is the structured condition in the Stage-1 reasoning-supervision ablation (Table~\ref{tab:eval_4_1_4}). The teacher must quantify lattice structure, composition, local coordination, topology, distortion, and symmetry; select three to five factors most relevant to the requested property; and explain their direction, strength, and competition. The Stage-2 predictor, training scale, optimization, and question-only pooling are held fixed across the comparison.
\end{CaseBlock}

\begin{CaseBlock}{User Input}
Given the CIF file of a crystal structure:

\CaseInputFileCompact{assets/case_study/lu2znpd.cif}

Please infer a reasonable Spillage (dimensionless) estimation for this structure.
\end{CaseBlock}

\begin{CaseBlock}{Model Response}
\CaseProtocolMarker{<think>}

The crystal structure provided corresponds to the intermetallic compound Lu2ZnPd, crystallizing in the triclinic space group P1 with lattice parameters a, b, and c all approximately 4.87 Å and angles α, β, and γ very close to 60 degrees. This metric suggests a underlying rhombohedral symmetry that is slightly distorted or represented in a primitive triclinic setting. The unit cell volume is roughly 81.6 Å, containing a total of four atoms per cell, which indicates a highly compact and structurally simple arrangement. The atomic positions are fully ordered with distinct Wyckoff sites for Lu, Zn, and Pd, implying a well-defined stoichiometry without disorder.

Chemically, the system comprises Lutetium, a heavy lanthanide with filled 4f shells, Palladium, a heavy transition metal with 4d electrons, and Zinc, a post-transition metal. The presence of Lu and Pd introduces significant atomic mass and potential spin-orbit coupling effects, which are critical factors when considering electronic structure projections. The average atomic mass is relatively high, and the electronegativity differences between the rare earth and the transition metals suggest a mix of metallic and covalent bonding characteristics. The localized nature of the Lu 4f and Pd 4d orbitals generally supports the construction of localized Wannier functions, which is the physical basis for the Spillage metric.

Regarding the key structural features influencing Spillage, the small number of atoms in the unit cell is the most dominant factor, as fewer bands typically reduce the complexity of subspace disentanglement. Secondly, the localized character of the d and f orbitals tends to minimize leakage compared to highly delocalized s-p systems, promoting lower spillage values. However, the heavy element content introduces spin-orbit coupling which can mix bands near the Fermi level, potentially acting as a counter-force that slightly increases the spillage compared to light-element analogs. The competition between the simplifying effect of the small cell size and the complicating effect of relativistic interactions determines the final magnitude.

To converge on a numerical estimate, one starts from the general observation that well-behaved intermetallics typically exhibit Spillage values well below 1.0, often in the range of 0.0 to 0.5 for successful Wannierizations. Given the exceptional simplicity of the four-atom cell, the baseline expectation is pushed toward the lower end of this spectrum, likely below 0.3. Adjusting for the heavy element complexity which prevents the value from being near-zero, a refined interval is established. Therefore, the Spillage is expected to be quite low, reflecting a robust projection, with a probable range centered around 0.1 with a moderate margin of uncertainty.

\CaseProtocolMarker{</think>}

The material Lu2ZnPd is evaluated for its Spillage value, a dimensionless property with an effective prediction range of [0, 4] that measures the fidelity of Wannier function localization in electronic structure calculations. This metric serves as an indicator of how well the selected orbital subspace represents the ab initio bands, which is particularly relevant for compounds containing heavy elements like Lutetium and Palladium where spin-orbit coupling is non-negligible.

The estimated value is primarily controlled by the highly compact four-atom unit cell which simplifies band disentanglement, and the localized nature of the Lu 4f and Pd 4d orbitals which favors tight binding representations. Although the presence of heavy atoms introduces spin-orbit coupling that may slightly increase subspace leakage, the structural simplicity dominates the interaction, leading to minimal spillage. Consequently, the Spillage is estimated to fall within the interval of 0.05 to 0.20, indicating a high-quality Wannier fit with strong orbital localization. Proceed to infer the spillage value.

\CaseProtocolMarker{<linear_pre>}
\end{CaseBlock}

\clearpage
\color{black}
\subsection{Molecular Property Prediction: ESOL Aqueous Solubility}
\label{sec:case_esol_property_prediction}

\begin{CaseBlock}{User Input}
Calculate the expected water solubility of \RouteToken{<SMILES>}CCCCC\RouteToken{</SMILES>} expressed as log S.
\end{CaseBlock}

\begin{CaseBlock}{Model Response}
\CaseProtocolMarker{<think>}

First, the molecule represented by the SMILES string CCCCC corresponds to n-pentane, a simple linear alkane composed entirely of five $\mathrm{sp}^3$-hybridized carbon atoms and their associated hydrogen atoms. This scaffold lacks any heteroatoms, functional groups, or ring systems, resulting in a purely hydrocarbon framework. The absence of electronegative atoms means the entire molecular surface is uniformly nonpolar, which fundamentally dictates its interaction with aqueous environments.

Next, the complete lack of oxygen, nitrogen, sulfur, or halogen atoms eliminates any possibility of hydrogen-bond donation or acceptance within this structure. Without polar bonds or lone pairs, the molecule cannot form stabilizing dipole--dipole interactions or hydrogen bonds with surrounding water molecules. Furthermore, there are no acidic or basic sites that could undergo protonation or deprotonation to generate charged species, ensuring the compound remains strictly neutral across all pH conditions and relies solely on weak van der Waals forces for solvation.

Then, the hydrophobic surface area is entirely defined by the methylene and terminal methyl groups of the pentyl chain, which create a continuous nonpolar interface that strongly disrupts the hydrogen-bonding network of bulk water. Although the linear aliphatic chain possesses conformational flexibility that allows it to adopt various gauche and anti conformations, this flexibility does not introduce any polar character or reduce the overall hydrophobic footprint. The molecular size is relatively modest at five carbons, but the ratio of nonpolar carbon surface to polar solvent atoms remains overwhelmingly high, severely limiting aqueous compatibility.

After that, the thermodynamic balance between water solvation and hydrophobic effects heavily favors the latter, as the energy penalty for cavity formation in water to accommodate this nonpolar chain far exceeds any minimal enthalpic gain from solute--solvent interactions. The absence of ionizable or polar moieties prevents the formation of hydration shells that would otherwise stabilize the solute in the aqueous phase. Consequently, the molecule exhibits a strong tendency to phase-separate from water, with solubility constrained primarily by the limited hydrodynamic volume and the inability to participate in specific solvent interactions.

Finally, considering the purely aliphatic nature, the total absence of hydrogen-bonding capacity, and the dominant hydrophobic character of the five-carbon chain, the expected aqueous solubility will be quite low. The structural profile aligns with small alkanes that partition almost entirely into organic phases rather than remaining dissolved in water, pointing toward a negative logarithmic solubility value that reflects poor hydration and minimal thermodynamic driving force for dissolution.

\CaseProtocolMarker{</think>}

For the ESOL task evaluating the water solubility of CCCCC, the structural analysis indicates a highly hydrophobic aliphatic chain with no polar or ionizable features. The precise numerical prediction for the logarithmic solubility is provided below.

\CaseProtocolMarker{<linear_pre>}
\end{CaseBlock}

\CaseNativeOutputHeading
\CaseResultCard{Predicted aqueous solubility}{$-3.2656$}{ESOL log S}

\clearpage
\color{black}
\subsection{CYP2C9 Substrate Classification}
\label{sec:case_cyp2c9}

\begin{CaseBlock}{User Input}
请评估
\RouteToken{<SMILES>}
\CaseInputFile{assets/case_study/cyp2c9_candidate.smiles}
\RouteToken{</SMILES>}
是否可能发生 CYP 介导的生物转化，即被细胞色素 P450 代谢酶进行化学修饰。
\end{CaseBlock}

\begin{CaseBlock}{Model Response}
\CaseProtocolMarker{<think>}

首先，对给定 SMILES 分子的结构特征进行解析。该分子具有较大的分子量，核心由多个芳香和杂芳香环系统构成。具体而言，它包含一个 4-甲氧基苯基，通过乙基链连接至一个哌嗪环；哌嗪环的另一个氮原子则连接到一个喹唑啉双环系统上；此外，喹唑啉环通过一个亚甲基桥连着一个 4-氟苯基。分子中存在的官能团和杂原子包括甲氧基、叔胺与仲胺氮原子、喹唑啉环内的杂芳氮以及一个氟原子。整体结构呈现出较高的刚性和疏水性，缺乏长链柔性脂肪族片段，且含有多个平面芳香环系统。

其次，明确当前任务旨在评估该分子的 CYP 底物识别特征，即探讨其被细胞色素 P450 代谢酶识别并发生化学修饰的结构基础。需要强调的是，底物识别任务关注的是分子作为反应物进入酶活性腔并被催化的可能性，这与分子是否抑制 CYP 酶活性是完全不同的概念。CYP 酶通常识别具有适当脂溶性、特定空间构象以及含有易被氧化位点的药物样分子。评估这一性质需要综合考量分子的疏水锚定能力、可代谢位点的电子云密度与空间可及性，以及分子整体与酶活性腔的适配程度，而不能简单地将其与酶活性影响任务混淆。

接下来，分析 CYP 底物识别的主导机制变量。CYP 介导的生物转化通常依赖于分子在酶活性腔内的正确定位，这要求分子具备足够的疏水性以进入疏水口袋，同时需要存在电子云密度较高或键能较低的易氧化位点，如苄位碳、脂肪族碳、杂原子上的烷基或富电子芳香环。此外，分子的三维构象和柔性对于将特定的代谢位点准确递送至 heme 铁中心至关重要。如果分子过于刚性或代谢位点被空间位阻保护，即使存在潜在的氧化位点，其被酶促代谢的效率也会大幅降低。

进一步来看，将当前分子的具体结构特征映射到 CYP 底物识别的机制依据上。该分子包含几个潜在的可代谢区域：4-甲氧基苯基上的甲氧基可能发生 O-脱甲基化，连接喹唑啉与氟苯基的亚甲基是一个潜在的苄位氧化位点，而 4-甲氧基苯基的芳香环也可能发生环氧化。然而，哌嗪环上的两个氮原子分别连接着苯乙基和喹唑啉基团，缺乏易于脱除的简单 N-烷基。喹唑啉环作为一个缺电子的稠杂芳系统，其芳香环电子云密度较低，通常难以作为 CYP 氧化的优先位点。这些结构特征表明，分子虽然具备一定的疏水骨架，但其潜在代谢位点的反应活性和可及性存在显著差异。

另一方面，需要考量可能削弱或改变上述代谢趋势的修正因素。尽管存在甲氧基和苄位亚甲基，但分子的整体刚性较强，多个芳香环和哌嗪环的共平面或固定构象可能限制了其在 CYP 活性腔中的自由旋转，从而阻碍了特定代谢位点与 heme 铁的有效靠近。同时，4-氟苯基上的氟原子具有强电负性，不仅显著降低了该苯环的电子云密度，使其难以发生芳香环氧化，还通过诱导效应可能影响相邻亚甲基的氧化倾向。此外，喹唑啉环的缺电子性质和较大的空间体积可能产生位阻效应，进一步保护了邻近区域免受酶促攻击。这些电子钝化和空间位阻因素共同增加了分子的代谢稳定性。

最后，综合来看，当前分子在 CYP 底物识别方面的结构依据呈现出复杂的多面性。一方面，其疏水芳香骨架和潜在的 O-脱甲基或苄位氧化位点为 CYP 识别提供了初步的结构线索；另一方面，分子的高刚性、缺电子杂芳环、氟原子带来的电子钝化效应以及潜在的空间位阻，显著限制了代谢位点与酶活性中心的适配性和反应活性。在评估其被细胞色素 P450 代谢酶进行化学修饰的可能性时，必须将这些电子效应、空间可及性与整体构象限制纳入统一的机制框架中。这些结构-机制证据的平衡将作为后续预测该分子 CYP 底物识别状态的核心依据。

\CaseProtocolMarker{</think>}

综合上述结构解析与机制探讨，该分子在被细胞色素 P450 代谢酶识别并发生化学修饰的评估中，展现出疏水骨架与潜在氧化位点同电子钝化、空间位阻及构象刚性之间的复杂博弈。甲氧基与苄位亚甲基提供了基础的代谢线索，而喹唑啉的缺电子性、氟原子的强电负性以及整体分子的刚性构象则构成了显著的代谢稳定化阻力。这些相互制约的结构因素共同界定了该分子在 CYP 底物识别任务中的机制证据边界，后续预测值的生成将严格依据这些理化与空间适配性证据展开。

\CaseProtocolMarker{<linear_cla>}
\end{CaseBlock}

\CaseNativeOutputHeading
\CaseResultCard{CYP2C9 底物分类：二分类标签}{0}{}

\clearpage
\color{black}
\subsection{CYP2D6 Inhibition Classification}
\label{sec:case_cyp2d6}

\begin{CaseBlock}{User Input}
Assess whether
\CaseInputFile{assets/case_study/cyp2d6_candidate.smiles}
is likely to inhibit CYP2D6 activity based on its structural features.
\end{CaseBlock}

\begin{CaseBlock}{Model Response}
\CaseProtocolMarker{<think>}

First, the molecular structure represented by the given SMILES string features a penam core, which consists of a thiazolidine ring fused to a beta-lactam ring. Attached to this core is a complex acyl side chain containing a phenyl ring, multiple amide linkages, and a cyclic urea moiety. Additionally, the molecule includes a carboxylate group neutralized by a sodium counterion, contributing to its overall ionic and highly polar character.

Second, the current analytical task focuses on evaluating the structural basis for CYP2D6 inhibition, which specifically examines the molecule's potential to affect the enzymatic activity of the CYP2D6 isoform. It is important to distinguish this inhibitory mechanism from the process of the molecule being metabolized by the enzyme, as the structural requirements for binding to the active site and blocking substrate turnover differ fundamentally from those for undergoing oxidative metabolism.

Next, the primary mechanism for CYP2D6 inhibition typically requires the presence of a protonatable basic nitrogen, such as a tertiary or secondary amine, which can form a strong ionic interaction with the conserved glutamate residue located in the enzyme's active site. Alongside this basic center, a hydrophobic aromatic or aliphatic region is generally necessary to anchor the molecule within the hydrophobic binding pocket, allowing the inhibitor to effectively occupy the space and prevent substrate access.

Furthermore, when mapping the specific structural features of this penicillin derivative to the CYP2D6 inhibition mechanism, a significant mismatch becomes evident. The molecule completely lacks any basic amine functionality; all nitrogen atoms are incorporated into amide bonds, the beta-lactam ring, or the cyclic urea, none of which possess the requisite basicity to be protonated at physiological pH. Consequently, the critical ionic anchoring interaction with the active site glutamate cannot be established.

However, beyond the absence of a basic center, the overall physicochemical profile of the molecule further diminishes its potential for CYP2D6 inhibition. The presence of the anionic carboxylate group, combined with multiple hydrogen bond donors and acceptors from the amides and urea, renders the molecule highly polar and hydrophilic. This high polarity severely restricts its ability to partition into the predominantly hydrophobic environment of the CYP2D6 active cavity, thereby weakening any potential hydrophobic anchoring provided by the phenyl ring.

Finally, synthesizing the structural and physicochemical evidence, the lack of a protonatable basic pharmacophore, the presence of a negatively charged carboxylate, and the high overall polarity collectively preclude the molecule from adopting the necessary binding mode for CYP2D6 inhibition. The structural features are instead characteristic of a highly polar, anionic beta-lactam antibiotic, which does not align with the established pharmacophore for blocking this specific cytochrome P450 enzyme.

\CaseProtocolMarker{</think>}

The structural profile of \nolinkurl{CC1(C)S[C@@H]2[C@H](NC(=O)[C@@H](NC(=O)N3CCNC3=O)c3ccccc3)C(=O)N2[C@H]1C(=O)[O-].[Na+]} is characterized by a highly polar penam core, an anionic carboxylate, and an absence of protonatable basic amines, which collectively fail to provide the essential ionic and hydrophobic interactions required for target engagement. The lack of a basic pharmacophore and the high hydrophilicity of the molecule strongly constrain its ability to occupy the hydrophobic active cavity of the enzyme. Consequently, the evaluation of CYP2D6 inhibition for this molecule relies on the absence of a basic anchoring center and the high polarity of the penam scaffold, which together form the structural basis for the final predictive output.

\CaseProtocolMarker{<linear_cla>}
\end{CaseBlock}

\CaseNativeOutputHeading
\CaseResultCard{CYP2D6 inhibition classification}{Non-inhibitor}{Binary label: 0}

\clearpage
\color{black}
\subsection{Infrared Spectrum-to-Molecular Generation}
\label{sec:case_spectrum_reconstruction}

\begin{CaseBlock}{User Input}
Please predict the possible molecular structure based on the provided infrared spectrum.

\begin{center}
  \includegraphics[width=0.88\textwidth]{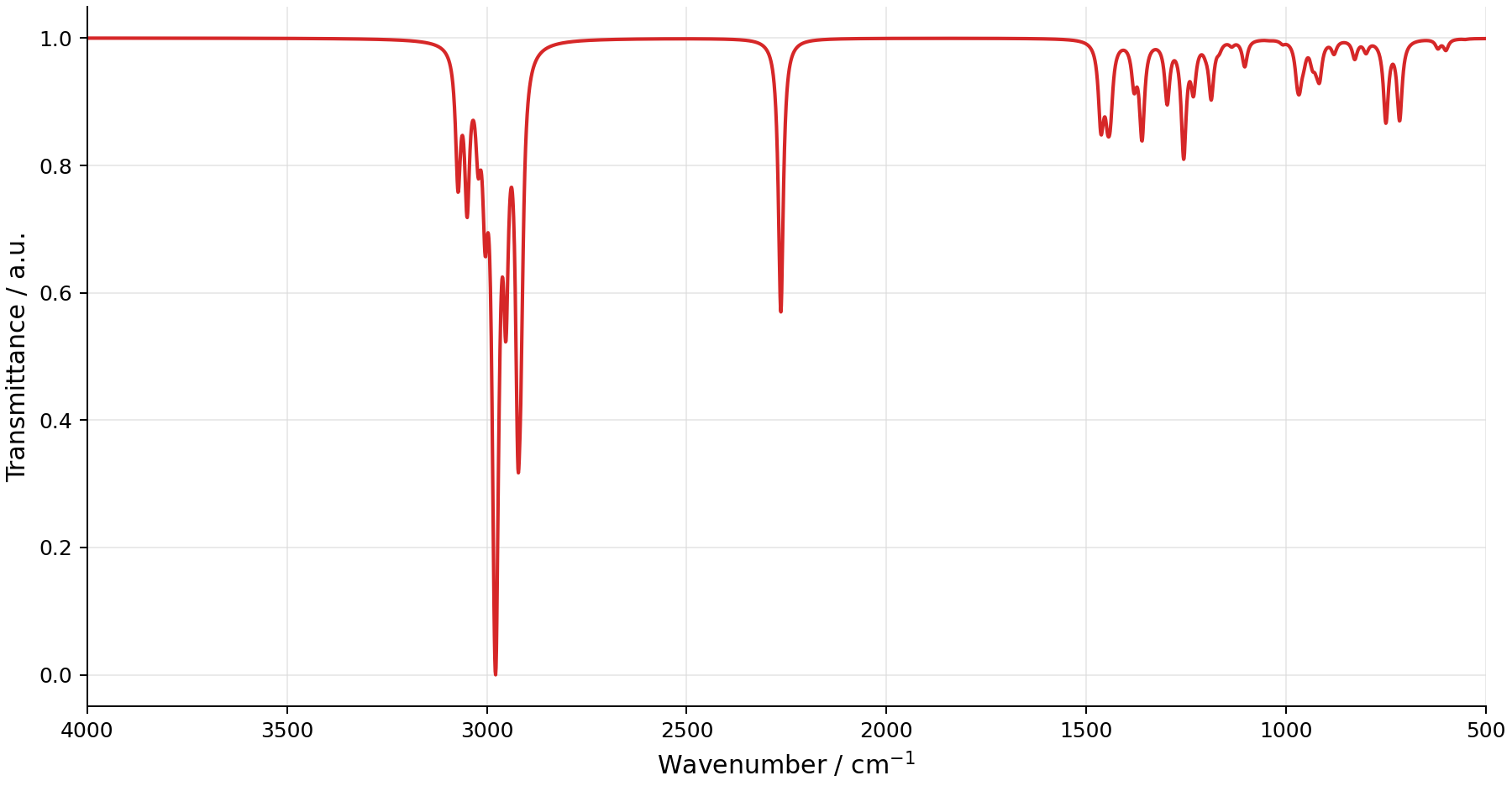}
\end{center}
\end{CaseBlock}

\begin{CaseBlock}{Model Response}
\CaseProtocolMarker{<think>}

The user asks for a molecular structure prediction from an infrared spectrum, so I first analyze the high-wavenumber region. The dominant absorption cluster has a maximum at 2970.0~cm$^{-1}$ and extends to approximately 2850~cm$^{-1}$. This pattern is characteristic of C--H stretching in saturated sp$^3$-hybridized carbon environments and indicates an alkyl or alkane component.

Weak but distinct absorptions also appear just above 3000~cm$^{-1}$, at approximately 3050--3080~cm$^{-1}$. Although this range can contain sp$^2$ C--H stretches, the spectrum lacks the corresponding C=C skeletal vibration expected between 1600 and 1680~cm$^{-1}$. The high-frequency C--H features are therefore more consistent with bonds in a strained ring, particularly cyclopropane, than with a conventional alkene or aromatic group.

A sharp, isolated peak near 2260~cm$^{-1}$ provides the strongest functional-group constraint. This region is associated with triple-bond stretching and can indicate either C$\equiv$N or C$\equiv$C. The absence of a terminal-alkyne C--H stretch near 3300~cm$^{-1}$, together with the clear intensity of the 2260~cm$^{-1}$ band, favors a polar nitrile group; an internal alkyne would often produce a weaker signal.

The fingerprint region contains absorptions near 1450 and 1380~cm$^{-1}$, consistent with methylene and methyl C--H scissoring, rocking, and bending modes. The spectrum has no broad absorption between 3300 and 3500~cm$^{-1}$ and no strong carbonyl band near 1700~cm$^{-1}$, excluding prominent hydroxyl, amine, and carbonyl functionalities.

Taken together, the sp$^3$ C--H cluster, the strained-ring C--H features above 3000~cm$^{-1}$, and the diagnostic nitrile stretch support a saturated, cyclopropane-containing nitrile structure.

\CaseProtocolMarker{</think>}

\textbf{Spectral analysis summary}

\begin{itemize}[leftmargin=1.5em]
  \item \textbf{Approximately 2970--2850~cm$^{-1}$:} strong sp$^3$ C--H asymmetric and symmetric stretching from saturated alkyl groups.
  \item \textbf{Approximately 3050--3080~cm$^{-1}$:} weak C--H stretching consistent with a strained cyclopropane ring in the absence of a C=C skeletal band.
  \item \textbf{Approximately 2260~cm$^{-1}$:} sharp triple-bond stretching assigned to a nitrile group, C$\equiv$N.
  \item \textbf{Approximately 1450 and 1380~cm$^{-1}$:} methylene and methyl C--H bending modes.
\end{itemize}

\textbf{Derived core functional groups.} The spectrum supports an alkane, a nitrile, and a cyclopropane ring. The nitrile assignment is strengthened by the absence of a terminal-alkyne C--H stretch, while the strained-ring interpretation is favored over an alkene because no C=C stretching band is observed.

Based on this spectral and functional-group information, the model proceeds to predict the possible molecular structure.

\textbf{Predicted molecular SMILES}: \texttt{CC1(C)C[C@H]2[C@@H](C\#N)[C@H]21} (Native SDF output)

\CaseProtocolMarker{<spectra_st>}
\end{CaseBlock}

\CaseNativeOutputHeading
\CaseResultCard{Predicted structure}{Cyclopropane-containing nitrile}{Native SDF output}

\begin{figure}[H]
  \centering
  \includegraphics[width=0.46\textwidth,trim=560bp 340bp 600bp 335bp,clip]{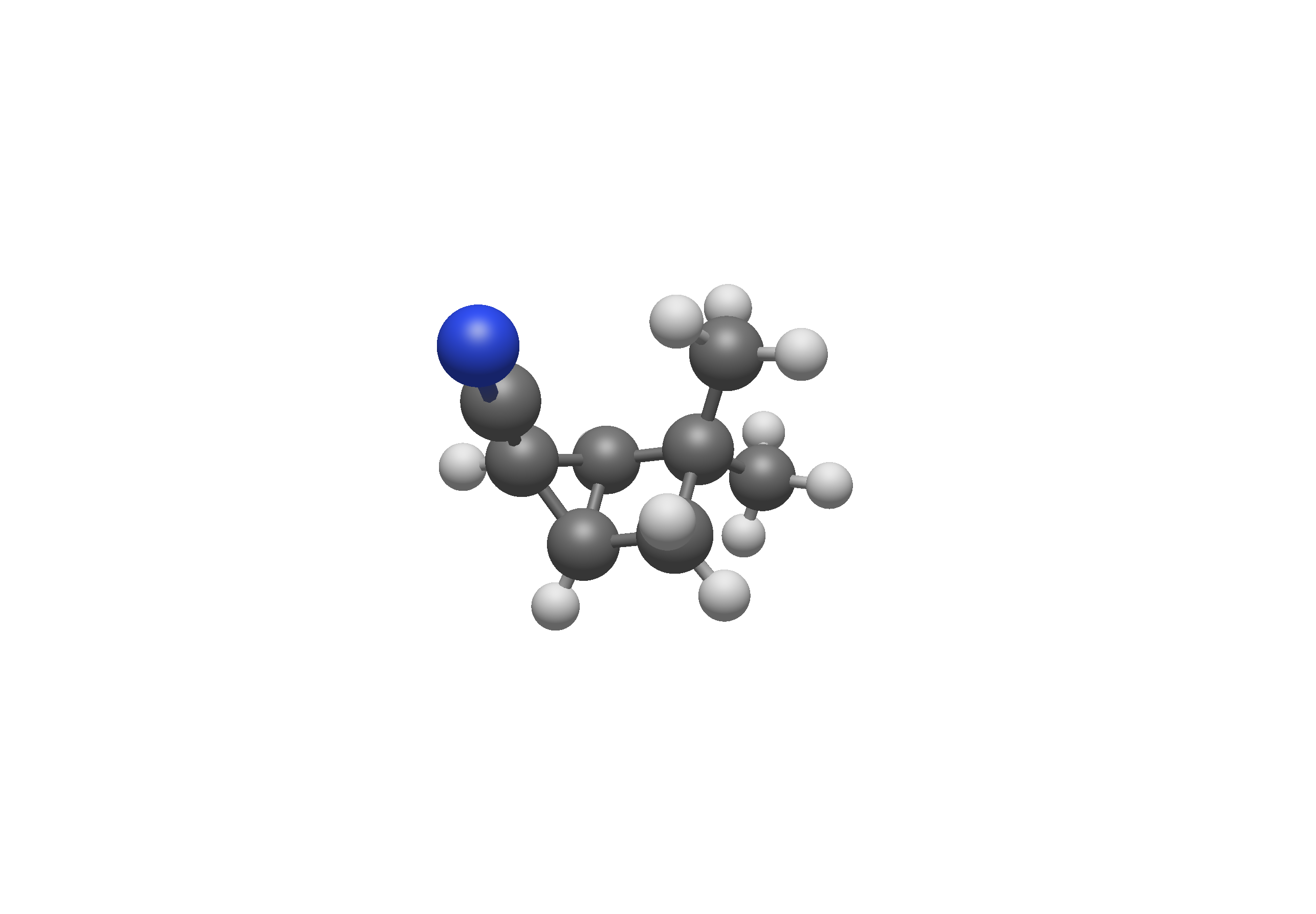}
  \caption{Native three-dimensional molecular output reconstructed from the supplied infrared spectrum.}
\end{figure}

\clearpage
\color{black}
\subsection{DNA-Binding-Site Prediction: ArcA}
\label{sec:case_protein_site}

\begin{CaseBlock}{User Input}
Given the protein sequence \RouteToken{<PROT>}
\CaseInputFile{assets/case_study/arca_sequence.txt}
\RouteToken{</PROT>}, predict which residues constitute the DNA-binding site.
\end{CaseBlock}

\begin{CaseBlock}{Model Response}
\CaseProtocolMarker{<think>}

Let me analyze this protein sequence to predict DNA-binding residues.

The sequence is:

\CaseInputFile{assets/case_study/arca_sequence.txt}

Let me first identify what protein this might be.

I'm noticing the N-terminal region is rich in basic amino acids like arginine and lysine, which is a classic signature of DNA-binding proteins. The stretch ``RPRRSAPRRAGKNP'' contains multiple positively charged residues that would interact with the negatively charged DNA backbone. This suggests the protein likely has a DNA-binding domain at the N-terminus. Continuing through the rest of the sequence... I'm scanning through the sequence position by position, noting the basic residues (arginine, lysine) that cluster in certain regions and could form a DNA-binding interface, similar to how HU and IHF proteins use their $\beta$-ribbon structures to contact the DNA minor groove. The C-terminal region appears to contain the dimerization domain based on the sequence composition. The $\beta$-hairpin structure inserts into the DNA minor groove, with key contacts made by threonine residues providing hydrogen bonds to bases and the backbone, while arginine residues interact with the phosphate groups. Looking at this specific sequence, I can identify potential DNA-binding residues by examining the N-terminal basic cluster---positions like R6, R8, R9, R13, R14, K16, and R19 appear to be the primary candidates for making these critical interactions. Looking at the structural data for ArcA from E. coli, I can refine my predictions using its winged helix-turn-helix motif---the DNA-binding domain spans roughly the first 50 residues, with the recognition helix containing the primary base contacts and the wing region providing additional phosphate backbone interactions. Based on this structural framework, my best prediction for the key DNA-binding residues focuses on those positioned in the recognition helix and wing elements. Looking at the crystal structure data more carefully, the primary DNA-contacting residues in ArcA are concentrated in the recognition helix and the adjacent wing region, with threonines and arginines making direct contacts to the DNA backbone and bases. I'm refining my prediction to focus on these structurally verified positions rather than just the basic-rich N-terminal region. The basic residues throughout the sequence suggest multiple DNA-binding regions, but focusing on the N-terminal winged-HTH domain around residues 1--50 gives the most likely candidates: R6, R8, R9, R13, R14, K16, R19, T32, H33, Q34, and several others that cluster in the predicted secondary structure elements. Without running actual computational docking simulations, this structural inference based on known ArcA mechanisms is my best approach. Looking at the sequence alignment more carefully, I can see the recognition helix likely spans around residues 27--40, where F29, H33, Q34, and R49 would make direct base contacts, while the wing region around residues 50--60 with Y53, Y54, H55, and K59 would provide additional DNA stabilization through minor groove interactions.

\CaseProtocolMarker{</think>}

\textbf{DNA-Binding Site Prediction}

\textbf{Protein Identification}

Based on sequence analysis, this protein resembles \textbf{ArcA} (ANA system component), a response regulator with a \textbf{winged helix-turn-helix (wHTH)} DNA-binding domain in the N-terminal region ($\sim$residues 1--58).

\textbf{Predicted DNA-Binding Residues}

\begin{center}
\small
\begin{tabularx}{\linewidth}{@{}>{\raggedright\arraybackslash}p{0.25\linewidth}>{\raggedright\arraybackslash}p{0.32\linewidth}X@{}}
\toprule
Region & Residues & Role \\
\midrule
Basic N-terminal patch & \textbf{R6, R8, R9, R13, R14, K16, R19} & Electrostatic contacts with phosphate backbone \\
Helix-2 / Recognition helix & \textbf{T32, H33, Q34, F29, R49} & Base-specific contacts in major groove \\
Wing region & \textbf{Y53, Y54, H55, F56, K59, T60} & Minor groove \& backbone contacts \\
\bottomrule
\end{tabularx}
\end{center}

\textbf{Structural Basis}

\begin{center}
\begin{minipage}{0.98\linewidth}
\ttfamily\scriptsize
Sequence: GAVGRPRRSAPRRAGKNPREEILDASAELFTHQGFATTSTHQIADAVGIRQASLYYHFPSK\par
Position: 1\hspace{1.2em}5\hspace{1.2em}10\hspace{1.2em}15\hspace{1.2em}20\hspace{4.4em}30\hspace{1.2em}35\hspace{0.8em}40\hspace{3.3em}49\hspace{0.8em}53\hspace{0.7em}56\hspace{0.5em}59\par
\hspace*{4.8em}$\uparrow\uparrow\uparrow\uparrow$\hspace{1.0em}$\uparrow\uparrow\uparrow$\hspace{0.7em}$\uparrow$\hspace{5.0em}$\uparrow\uparrow\uparrow$\hspace{2.5em}$\uparrow$\hspace{2.5em}$\uparrow\uparrow\uparrow\uparrow$\par
\hspace*{3.0em}Arg patch\hspace{1.0em}K\hspace{2.2em}Recog. helix\hspace{1.2em}contacts\hspace{1.2em}R49\hspace{1.2em}Wing
\end{minipage}
\end{center}

\begin{itemize}[leftmargin=1.5em]
  \item \textbf{R/K residues} (R6, R8, R9, R13, R14, K16, R19): electrostatic interactions with the phosphodiester backbone.
  \item \textbf{Recognition helix ($\alpha$3)}: F29, T32, H33, Q34 insert into the major groove for sequence-specific contacts.
  \item \textbf{Wing ($\beta$1--$\beta$2)}: Y53, Y54, H55, K59 clamp the DNA at the minor groove backbone.
\end{itemize}

\textbf{Key Predicted Contact Residues (Summary)}

\textbf{R6 $\cdot$ R8 $\cdot$ R9 $\cdot$ R13 $\cdot$ R14 $\cdot$ K16 $\cdot$ R19 $\cdot$ T32 $\cdot$ H33 $\cdot$ Q34 $\cdot$ R49 $\cdot$ Y53 $\cdot$ Y54 $\cdot$ H55 $\cdot$ K59}

These residues form the composite DNA-binding interface consistent with the wHTH mechanism observed in ArcA-family transcription factors. The actual residue index of the site is:

\CaseProtocolMarker{<prot_cla>}
\end{CaseBlock}

\CaseNativeOutputHeading

\CaseResidues{V3, P6, R7, S39, T40, H41, R50, A52, S53, Y55, Y56, Q117.}

\begin{figure}[H]
  \centering
  \includegraphics[width=0.78\textwidth]{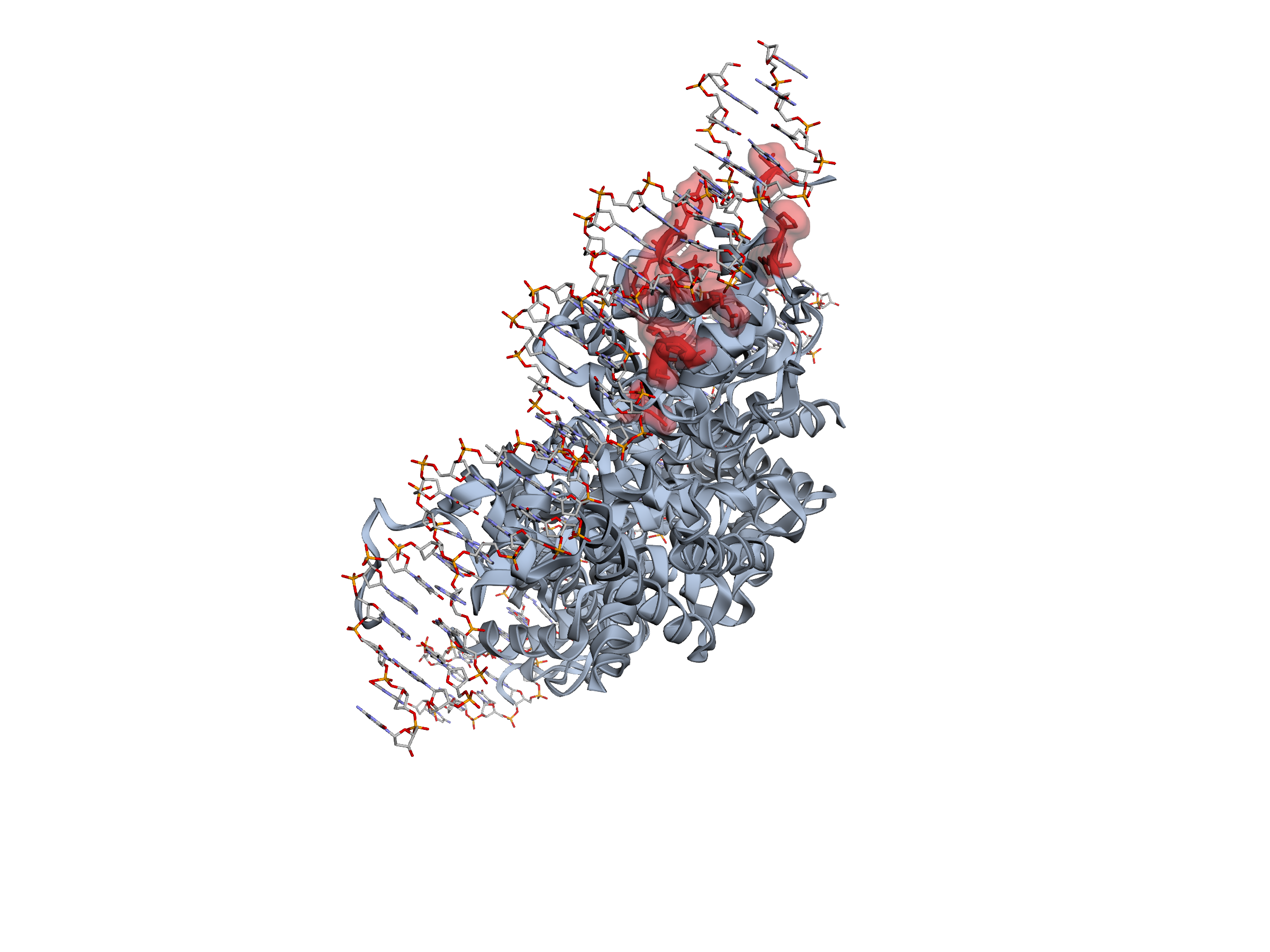}
  \caption{Native site-prediction output for ArcA. The twelve site-head positions are highlighted in red on chain A of PDB 5DY0.}
\end{figure}

\clearpage
\color{black}
\subsection{Protein Structure Prediction: GPR168}
\label{sec:case_protein_structure}

\begin{CaseBlockImmediate}{User Input}
请分析并预测下面这条人类蛋白质氨基酸序列的三维结构：

\CaseInputFile{assets/case_study/gpr168_sequence.txt}
\end{CaseBlockImmediate}

\begin{CaseBlock}{Model Response}
\CaseProtocolMarker{<think>}

我首先确认用户提供的是一条人类蛋白质氨基酸序列，任务目标并非简单的数据库信息检索，而是为了后续的三维结构预测挖掘关键的生物学约束线索。这条序列对应的是 G 蛋白偶联受体 168（GPR168），一种典型的孤儿受体，其核心特征在于作为七次跨膜蛋白嵌入细胞膜中。

从序列规模来看，该蛋白全长约 382 个氨基酸，属于中等长度的膜蛋白。这一长度对于容纳完整的七次跨膜螺旋束以及必要的胞内/胞外环区是充足的，同时也暗示了 N 端可能存在较短的信号肽或无序区域，而 C 端则包含用于信号转导的关键基序。

在可溶性与拓扑结构方面，序列中富含疏水性残基片段，这是典型的多重跨膜蛋白特征。根据家族保守性推断，它将形成七个$\alpha$-螺旋贯穿脂双层，确立为“七次跨膜”（7TM）架构。这种拓扑结构意味着大部分亲水性区域位于细胞质侧或细胞外侧，形成了特定的折叠模式：一个朝向外部配体结合口袋的胞外域，以及负责招募下游效应分子的胞内尾部。

关于结构域与折叠类型，该序列归属于视紫红质样 GPCR 家族。这意味着其整体骨架遵循经典的 Class A GPCR 折叠方式，即由七个紧密堆积的跨膜螺旋组成桶状结构，并通过高度保守的连接环维持稳定性。特别是第三胞内环和 C 末端区域，虽然序列变异较大，但通常具有内在无序特性，以便灵活地结合异三聚体 G 蛋白或其他调节因子。

活性位点与结合机制方面，由于被标记为孤儿受体且缺乏已知的小分子配体结合证据，其正构结合口袋可能主要针对大分子如多肽激素或糖缀合物设计，或者其激活机制依赖于变构调节及与其他膜蛋白的相互作用。然而，其胞内面必然保留有保守的微开关元件（如 DRY 基序、NPxxY 基序等），这些是触发构象变化并启动 G 蛋白偶联的核心结构基础。

在定位与互作背景上，作为一个整合膜蛋白，它定位于质膜。其寡聚状态可能是单体也可能是同源二聚体，这取决于具体的界面互补性。考虑到其潜在的功能涉及神经内分泌调节和免疫反应，其结构表面必须能够适应复杂的细胞环境，甚至可能通过胞外域介导细胞间的粘附作用，从而影响局部受体簇的组装。

综合上述因素，这对三维结构预测提出了明确要求：模型需要准确构建出埋藏在膜内的七螺旋束几何形状，正确放置带电荷残基以稳定螺旋间相互作用，并合理模拟柔性较大的胞内环区。同时，预测需考虑其作为信号枢纽的潜力，即胞内表面应具备与 G 蛋白亚基对接的空间构型，这对于理解其动态构象转换至关重要。

\CaseProtocolMarker{</think>}

基于您提供的氨基酸序列，以下是对该蛋白质（GPR168）进行结构预测前的深度生物学分析与线索梳理：

\textbf{1. 序列概览与基本属性}

该序列编码一个人类来源的蛋白质，全长约为 382 个氨基酸。从整体组成看，它表现出显著的两亲性特征，含有多个长段的强疏水区域，这是典型的\textbf{多次跨膜蛋白}标志。这类蛋白无法像球状可溶蛋白那样自由扩散，而是必须嵌入磷脂双分子层中才能发挥功能。

\textbf{2. 拓扑结构与折叠框架}

根据序列中的疏水峰分布及家族保守性，可以确定该蛋白采用\textbf{七次跨膜}（7-TM）。这是一种进化上高度保守的折叠模式，常见于视紫红质样 G 蛋白偶联受体（Class A GPCRs）。

\begin{itemize}[leftmargin=1.5em]
  \item \textbf{跨膜核心：}预计存在七个$\alpha$-螺旋（Helix I--VII），它们垂直穿过细胞膜，形成一个紧凑的圆柱形束。这些螺旋之间的排列受到严格的立体化学限制，尤其是螺旋 III 和 VII 的关键保守残基网络，对维持结构完整性至关重要。
  \item \textbf{环路区域：}连接跨膜螺旋的是三个胞外环（ECLs）和三个胞内环（ICLs）。其中，第二胞外环（ECL2）通常较长且可能含有半胱氨酸形成的二硫键，起到封闭配体结合口袋的作用；而第三胞内环（ICL3）往往较长且具有较高的灵活性，是信号转导的关键接口。
\end{itemize}

\textbf{3. 功能位点与动态机制推测}

尽管该受体目前被称为“孤儿受体”（Orphan Receptor），即尚未发现其天然小分子配体，但其结构特征揭示了潜在的运作机制：

\begin{itemize}[leftmargin=1.5em]
  \item \textbf{配体结合腔：}在七个跨膜螺旋围成的内部空腔中，可能存在一个适合结合大分子配体（如多肽激素或糖复合物）的口袋，而非传统的小分子药物结合位点。
  \item \textbf{信号转导界面：}在面向细胞质的一侧，特别是跨膜螺旋 VI 的胞内端和 C 末端尾部，隐藏着保守的构象开关（如 NPxxY 和 DRY 基序）。当受体发生构象改变时，这些区域会暴露出特定的疏水斑块和电荷分布，用于特异性地招募并激活异三聚体 G 蛋白（如 Gi/o 或 Gq/11 类型）。
\end{itemize}

\textbf{4. 细胞环境与互作潜能}

作为一种定位于\textbf{质膜}的整合膜蛋白，其结构不仅服务于单一的信号传递，还可能参与更广泛的细胞通讯网络：

\begin{itemize}[leftmargin=1.5em]
  \item \textbf{寡聚化倾向：}GPCRs 常常以二聚体或多聚体形式存在。该蛋白的跨膜界面可能支持同源或异源寡聚化，从而调节其敏感性和信号输出。
  \item \textbf{非经典功能：}鉴于其作为 GPCR 跨膜受体的结构特征及其在神经系统和免疫系统中的表达潜力，该蛋白的胞外域可能兼具细胞粘附功能，通过与其他细胞表面分子的瞬时接触来调控受体的聚集状态和信号强度。
\end{itemize}

\textbf{5. 对结构预测的指导意义}

综上所述，在进行三维结构建模时，应重点关注以下几点：

\begin{itemize}[leftmargin=1.5em]
  \item 强制施加膜环境约束，确保七螺旋束的倾斜角和旋转角度符合生物物理规律。
  \item 优先保证跨膜螺旋间保守盐桥和氢键网络的准确性，这是决定受体静息态与激活态平衡的基础。
  \item 对胞内柔性环区采取适当的系综采样策略，因为它们的精确构象直接决定了下游效应器的结合能力。
\end{itemize}

基于上述对序列特征、拓扑结构、功能位点及细胞环境的详细剖析，我们将利用这些关键线索作为约束条件，启动高精度的三维结构预测流程，以解析该孤儿受体的独特空间构象及其潜在的分子工作机制。

\CaseProtocolMarker{<prot_st>}
\end{CaseBlock}

\CaseNativeOutputHeading

\begin{figure}[H]
  \centering
  \includegraphics[width=0.76\textwidth]{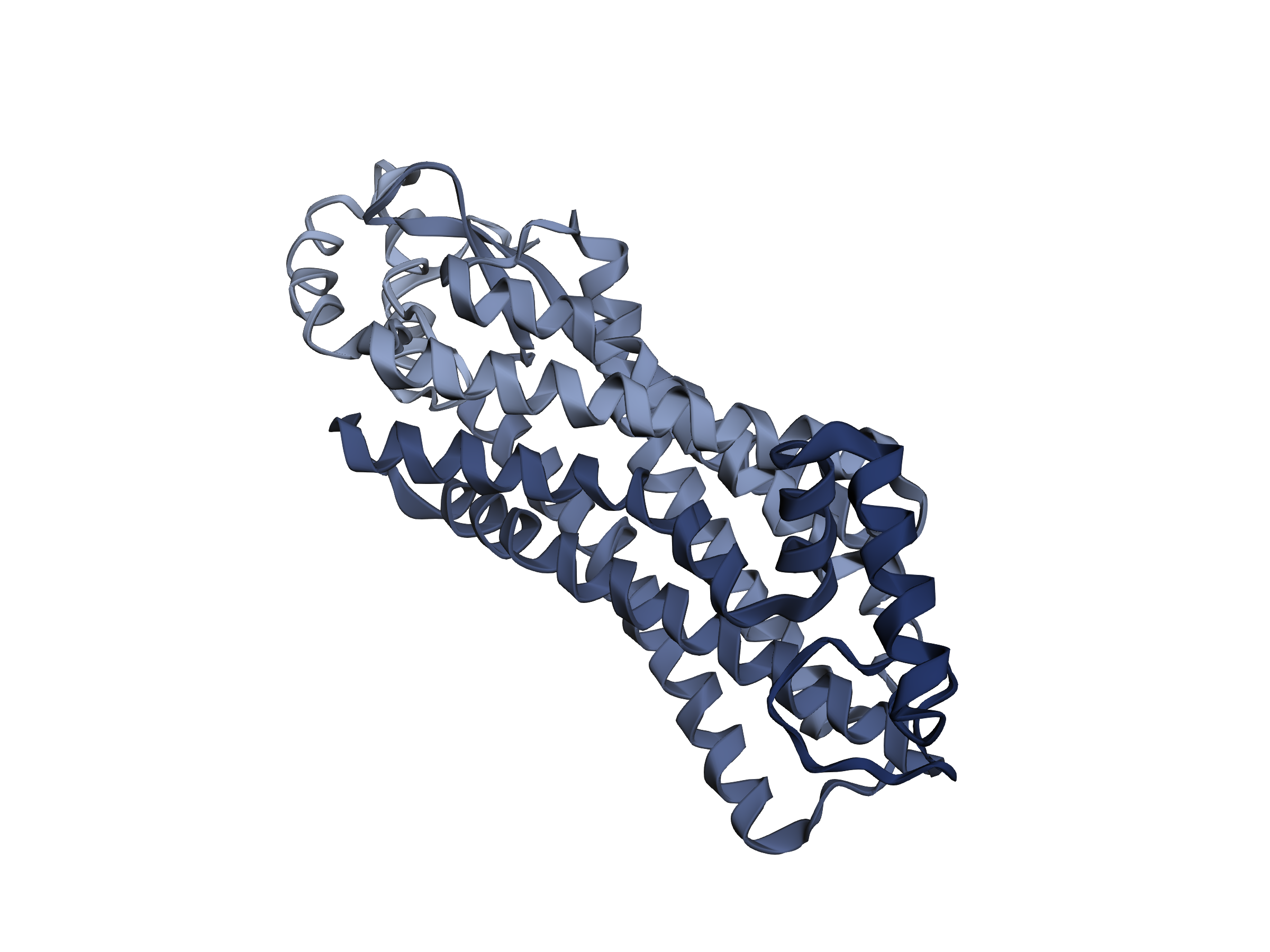}
  \caption{Native SimpleFold-style structure output for GPR168, containing 3,008 atoms.}
\end{figure}

\clearpage
\color{black}
\subsection{Scientific Image Generation: Chemical Synapse}
\label{sec:case_image_synapse}

\begin{CaseBlock}{User Input}
Generate a detailed biological diagram of a chemical synapse, clearly labeling the ``PRESYNAPTIC TERMINAL'' at the top, which contains multiple orange circles labeled ``SYNAPTIC VESICLE'' filled with ``NEUROTRANSMITTERS''. The gap between structures is marked ``SYNAPTIC CLEFT'', leading down to the ``POSTSYNAPTIC TERMINAL'' featuring ``RECEPTOR'' sites. Include additional labels for ``ACTION POTENTIAL'' arriving at the terminal, ``CALCIUM IONS'' entering the cell, and ``ION CHANNELS'' embedded in the postsynaptic membrane.
\end{CaseBlock}

\begin{CaseBlock}{Model Response}
\CaseProtocolMarker{<think>}

I'll start by sketching the upper bulbous structure and placing the main header ``PRESYNAPTIC TERMINAL'' centered at the very top in dark blue letters. On the upper left side, I'll draw lightning bolts traveling down the axon and label this ``ACTION POTENTIAL'' with the explanatory text ``(nerve impulse arriving at terminal)'' right underneath it. Inside the main blue terminal body, I'll distribute several orange circles and point to one with a line labeled ``SYNAPTIC VESICLE'', while noting nearby that they contain ``NEUROTRANSMITTERS'' with the subtitle ``(inside vesicles)''. On the left membrane wall, I'll depict purple dots flowing through a channel and label this group ``CALCIUM IONS'' followed by ``(Ca$^{2+}$)'' and the phrase ``entering the terminal'' to explain the influx.

Next I need to define the space between the two cells, so I'll draw a bracket on the right side spanning the gap and label it ``SYNAPTIC CLEFT'' with the description ``(gap between terminals)'' below it. I'll show the orange neurotransmitter dots released from the top structure floating down through this space. Below that gap, I'll draw the curved receiving membrane and center the label ``POSTSYNAPTIC TERMINAL'' at the very bottom. On the surface of this lower membrane, I'll place green structures facing upwards and label one ``RECEPTOR'' with the note ``(binds neurotransmitter)'' to show where the chemicals attach.

Finally, I'll add the mechanisms inside the lower membrane, drawing blue channels that allow ions to pass through and labeling them ``ION CHANNELS'' on the left with the detailed text ``(open or close in response to neurotransmitter binding)''. To ensure clarity on the different particles floating around, I'll add a small legend box in the bottom right corner containing three entries: ``NEUROTRANSMITTER'' next to an orange dot, ``CALCIUM ION (Ca$^{2+}$)'' next to a purple dot, and ``OTHER IONS (e.g., Na$^+$, K$^+$)'' next to a lighter purple dot. This should cover all the necessary biological components and labels for a complete diagram.

\CaseProtocolMarker{</think>}

I'll draw a detailed biological diagram of a chemical synapse that clearly labels the presynaptic terminal, synaptic cleft, and postsynaptic terminal. This illustration will depict synaptic vesicles releasing neurotransmitters to receptor sites while indicating the role of calcium ions and ion channels.

\CaseProtocolMarker{<image_gen>}
\end{CaseBlock}

\CaseNativeOutputHeading

\begin{figure}[H]
  \centering
  \includegraphics[width=0.45\textwidth]{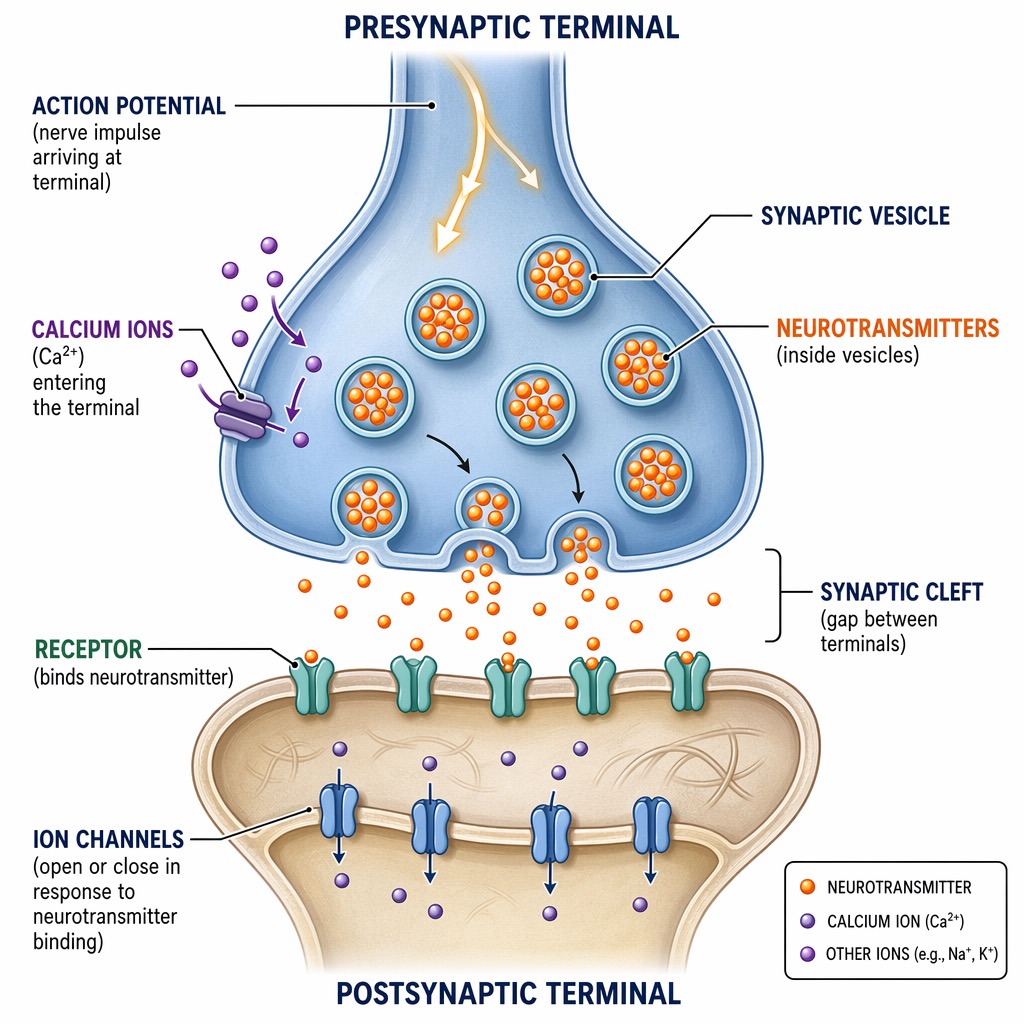}
  \caption{Native generated output for the chemical-synapse illustration request.}
\end{figure}

\clearpage
\color{black}
\subsection{Scientific Image Editing: Tumor and Edema Segmentation}
\label{sec:case_medical_segmentation}

\begin{CaseBlock}{User Input}
请用红色高亮标出这张 FLAIR MRI 中脑肿瘤及其周围水肿的所有区域。

\begin{center}
  \includegraphics[width=0.34\textwidth]{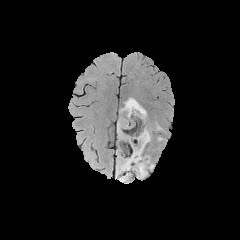}
\end{center}
\end{CaseBlock}

\begin{CaseBlock}{Model Response}
\CaseProtocolMarker{<think>}

用户希望我识别这张扫描图中的肿瘤和水肿区域，并用红色半透明叠加层进行标记。观察这幅图像，它看起来是一张脑部 FLAIR MRI，其中脑脊液信号受到抑制，使病理性高亮信号区域清晰突出。我注意到左侧大脑半球存在一片显著的高信号区，具体累及顶叶并延伸至皮质下白质。此处的高信号形态不规则且范围相当广泛，提示中心肿块周围存在血管源性水肿。图中还可见明显的占位效应，因为与对侧相比，该侧脑沟受压变平；病灶边界不清，这通常提示胶质瘤等浸润性病变。病灶内部的异质性则意味着其中可能混合存在实性肿瘤成分和坏死区域。鉴于这些表现与高级别胶质瘤及其周围水肿相符，我将生成红色半透明叠加层，为用户高亮标出这些特定的异常区域。

\CaseProtocolMarker{</think>}

我已识别出左侧顶叶中与肿瘤及其周围水肿相符的高信号病灶。现在我将应用红色半透明叠加层，在图像上高亮标出这些区域。

\CaseProtocolMarker{<image_edit>}
\end{CaseBlock}

\CaseNativeOutputHeading

\begin{figure}[H]
  \centering
  \includegraphics[width=0.42\textwidth]{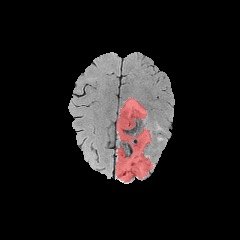}
  \caption{Native segmentation-as-editing output. The complete input image appears in the user request above.}
\end{figure}

\endgroup

\color{black}
\clearpage
\subsection{Scientific Image Analyses}
\label{sec:case_scientific_image_analysis}

\begin{figure}[H]
  \centering
  \includegraphics[width=0.94\textwidth,height=0.76\textheight,keepaspectratio]{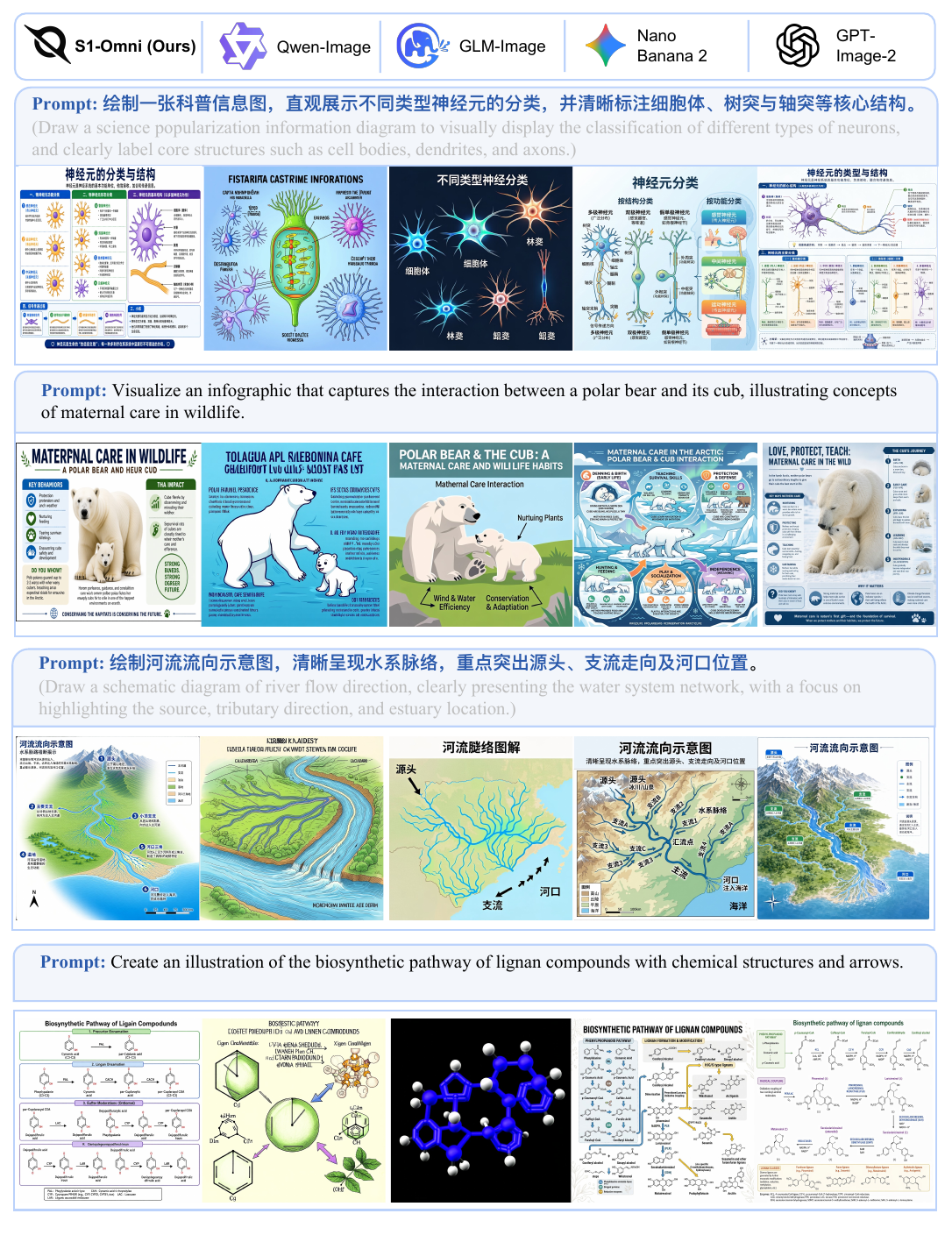}
  \caption{Qualitative comparison of S1-Omni, Qwen-Image, GLM-Image, Nano Banana 2, and GPT-Image-2 on neuron classification, maternal care in wildlife, river-network mapping, and lignan biosynthesis. The cases expose differences in scientific organization, visual completeness, and text rendering.}
  \label{fig:eval_4_5_1}
\end{figure}

\begin{figure}[p]
  \centering
  \includegraphics[width=\textwidth,height=0.86\textheight,keepaspectratio]{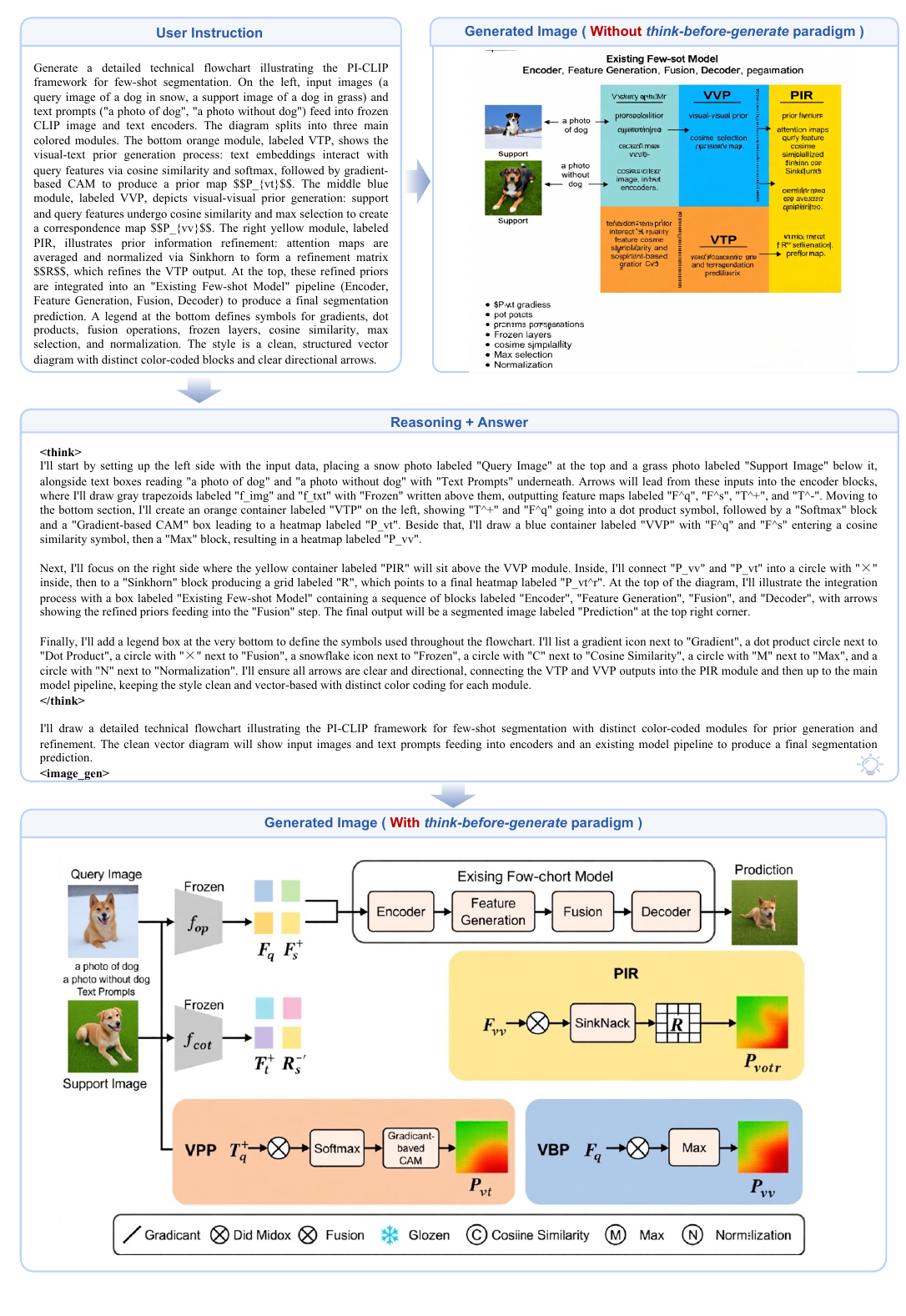}
  \caption{Effect of think-before-generate planning on a scientific workflow diagram. The intermediate reasoning state decomposes the requested modules, spatial relations, labels, and information flow before conditioning the image decoder.}
  \label{fig:discussion_think_before_generate}
\end{figure}

\begin{figure}[p]
  \centering
  \includegraphics[width=\textwidth,height=0.88\textheight,keepaspectratio]{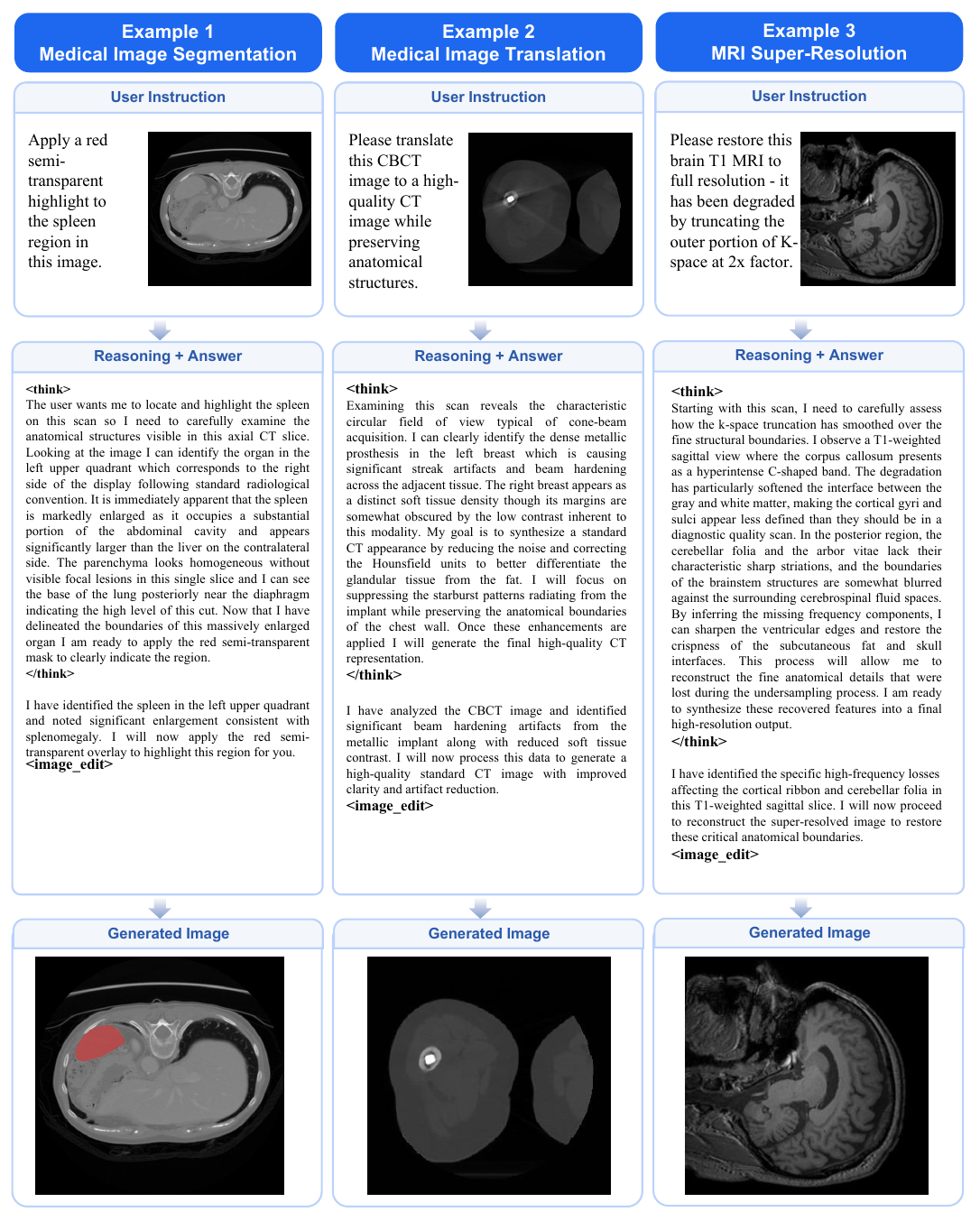}
  \caption{Representative scientific image editing cases using the shared \texttt{<image\_edit>} route: spleen segmentation with a spatial overlay, CBCT-to-CT medical image translation, and MRI super-resolution. Each column shows the instruction, reasoning and answer, and generated result.}
  \label{fig:case_scientific_image_editing}
\end{figure}

The following cases complement the benchmark scores with direct visual evidence. Figure~\ref{fig:eval_4_5_1} compares scientific image generators across four structurally different prompts. Figure~\ref{fig:discussion_think_before_generate} isolates the effect of explicit planning before image synthesis. Figure~\ref{fig:case_scientific_image_editing} shows how the shared editing route supports spatial segmentation, modality translation, and super-resolution.

\FloatBarrier

\color{black}
\clearpage
\color{black}
\subsection{Failure Mode Analysis}
\label{sec:case_failure_modes}

Although \model can produce structurally complete and coherent reasoning across diverse scientific tasks, several representative failure modes remain in evidence recognition, scientific attribution, and uncertainty control. As shown below, structural misreadings and unsupported mechanistic assumptions can systematically bias property estimates; spectrum-to-molecular generation can converge to an incorrect molecule after misreading peaks or making overly exclusive assignments; pre-prediction analysis for protein structure can hallucinate sequence length, family, function, or biological context; and scientific image generation can corrupt glyphs, add or omit strokes, blur characters, or replace their meaning when prompts require dense annotations. These errors arise at different stages of input recognition, evidence interpretation, and result generation, but share a common pattern: a local mistake or unverified assumption accumulates along the reasoning chain and is amplified by the downstream prediction or generation module.

These cases indicate that the limitations are not isolated errors in individual tasks. Instead, they expose unresolved challenges in unified scientific modeling: fine-grained representations of heterogeneous scientific inputs remain unstable; scientific knowledge and physical constraints do not consistently enter the reasoning process; domain-native outputs are sensitive to local errors; and the current data scale and task coverage are insufficient for more complex, dense, and out-of-distribution scientific settings. The next stage should therefore advance beyond demonstrating the feasibility of a unified task protocol and connected modules toward more reliable and scalable unified multimodal reasoning for scientific understanding, prediction, and generation. Larger-scale continual scientific pretraining and broader task coverage can strengthen shared representations; scientific laws, expert knowledge, and cross-modal evidence can constrain reasoning; uncertainty calibration can suppress unsupported high-confidence claims; and domain-native outputs such as molecular structures and scientific images can be made more accurate and readable.

\color{black}\clearpage
\color{black}
\subsubsection{Property-Prediction Reasoning}
\label{sec:case_failure_property}

\begin{CaseBlock}{User Input}
Please predict the ESOL aqueous solubility value for the following molecule.

\CaseInputFile{assets/case_study/failure_esol.smiles}
\end{CaseBlock}

\CaseResultCard{Scalar shown before the reasoning}{$-2.5156$}{ESOL aqueous solubility}

\begin{CaseBlockCompact}{Model Response}
\CaseProtocolMarker{<think>}

First, the input SMILES describes a rigid, fused tricyclic scaffold composed of a benzene ring, a pyridine ring, and a six-membered lactam ring bridged by an oxygen atom. The core architecture features an N-methyl cyclic amide where the nitrogen atom is directly substituted with a methyl group and integrated into the heterocyclic framework. This fused arrangement creates a predominantly planar molecular geometry, with the carbonyl group at the lactam position and the bridging ether oxygen establishing a constrained bicyclic ether-amide system. The presence of the pyridine nitrogen within the central ring further defines the electronic character of the scaffold, contributing to an extended conjugated $\pi$-system across the entire fused framework.

Next, the polarity profile of this molecule is dictated by three distinct hydrogen-bond acceptors: the lactam carbonyl oxygen, the pyridine nitrogen, and the bridging ether oxygen. Notably, the structure lacks any hydrogen-bond donors, as the amide nitrogen is fully substituted with a methyl group and there are no hydroxyl or secondary amine functionalities present. The pyridine nitrogen possesses weak basicity due to its involvement in an extended aromatic system and adjacency to electron-withdrawing conjugated bonds, rendering it largely unprotonated under neutral aqueous conditions. Consequently, the molecule relies entirely on dipole--dipole interactions and lone-pair acceptance to engage with water molecules, which limits its capacity to form extensive hydration shells compared to compounds bearing donor groups.

Then, the hydrophobic character is pronounced due to the extensive fused aromatic surface area spanning the benzene and pyridine rings, alongside the nonpolar N-methyl substituent. This planar, rigid architecture significantly reduces conformational flexibility, preventing the molecule from adopting compact or water-friendly folded states. The large contiguous $\pi$-system promotes strong intermolecular $\pi$--$\pi$ stacking interactions in the solid state, which must be overcome for dissolution to occur. Additionally, the absence of aliphatic chains or flexible spacers means the hydrophobic surface remains exposed and uninterrupted, further enhancing crystal lattice stability and reducing the thermodynamic driving force for aqueous solvation.

After that, the overall solubility behavior emerges from a clear imbalance between the modest polarity of three acceptor sites and the dominant hydrophobic, stacking-prone aromatic framework. While the carbonyl and ether oxygens provide localized polarity, they are insufficient to compensate for the extensive nonpolar surface and the complete lack of hydrogen-bond donation. The rigid, planar geometry facilitates tight molecular packing in the crystalline phase, substantially increasing the lattice energy penalty required for dissolution. As a result, the molecule exhibits a strong preference for the solid or organic phase, with water molecules unable to effectively disrupt the intermolecular aromatic interactions or adequately solvate the hydrophobic faces.

Finally, considering the combined effects of the fused tricyclic aromatic system, the complete absence of hydrogen-bond donors, the weakly basic pyridine nitrogen, and the high degree of molecular rigidity, the aqueous solubility is expected to be quite low. The structural features collectively point toward a compound that dissolves poorly in water, with solvation limited by strong crystal packing forces and inadequate hydrophilic surface area. This qualitative assessment aligns with molecules that typically exhibit significantly negative logarithmic solubility values, reflecting a pronounced hydrophobic character and minimal aqueous compatibility.

\CaseProtocolMarker{</think>}

Based on the structural analysis of the molecule for the ESOL task, the combination of a rigid fused tricyclic aromatic system, three hydrogen-bond acceptors, zero hydrogen-bond donors, and high planarity strongly indicates poor water solubility.

\CaseProtocolMarker{<linear_pre>}
\end{CaseBlockCompact}

\CaseResultCard{Final answer}{$-3.6719999999999997$}{ESOL aqueous solubility}

\begin{CaseFailureNote}{Observed Failure}
The scalar displayed before the reasoning ($-2.5156$) conflicts with the final answer ($-3.672$). The long mechanistic narrative does not resolve or acknowledge this numerical inconsistency, illustrating how a locally plausible explanation can obscure an unstable prediction.
\end{CaseFailureNote}

\color{black}\clearpage
\color{black}
\subsubsection{Spectrum-to-Molecular Generation}
\label{sec:case_failure_spectrum}

\begin{CaseBlock}{User Input}
大模型谱图预测。输入为以下 Raman 谱图。

\begin{figure}[H]
  \centering
  \includegraphics[width=0.92\textwidth]{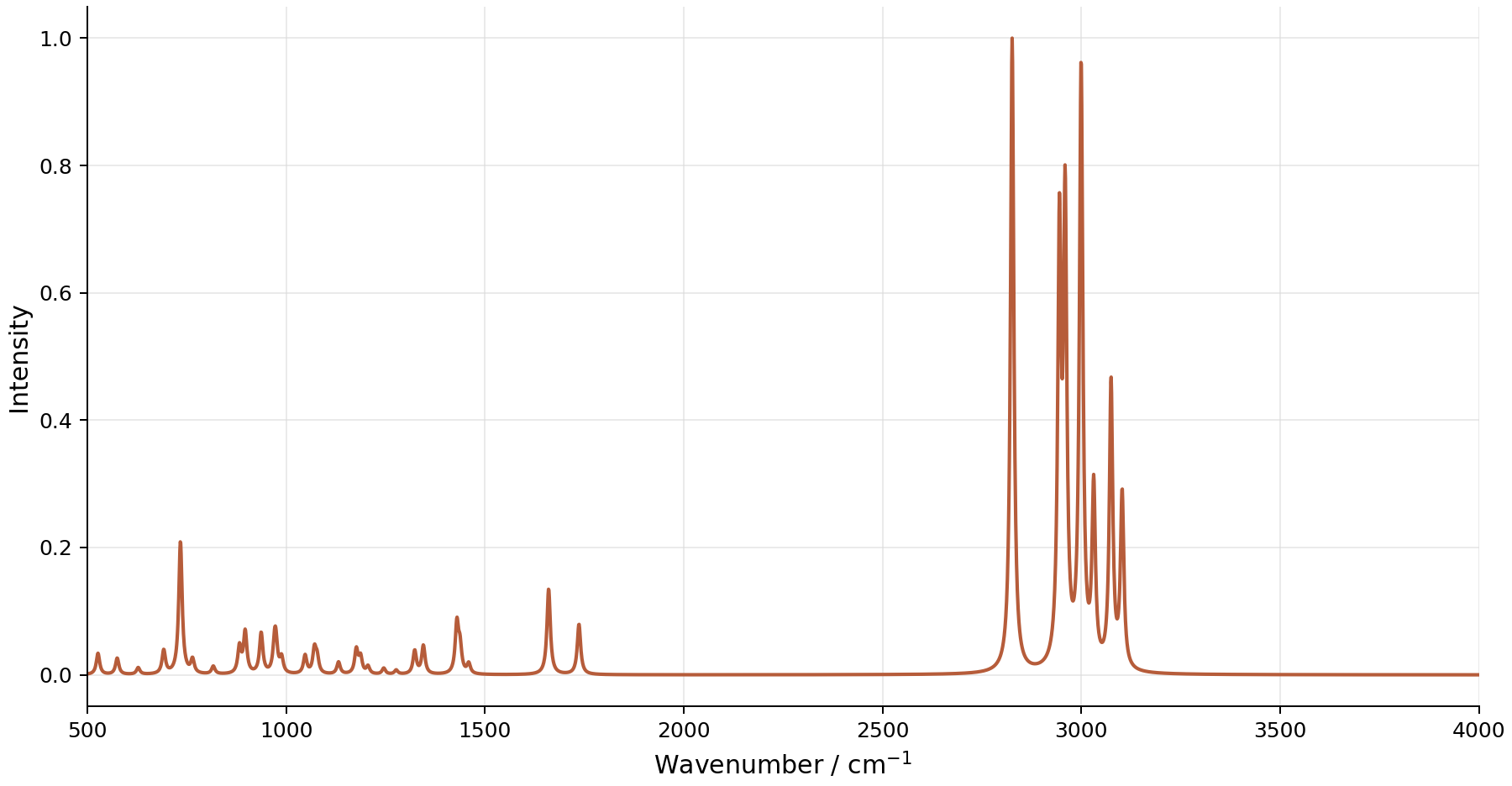}
  \caption{Raman spectrum used by the failed reconstruction example.}
\end{figure}
\end{CaseBlock}

\begin{CaseBlock}{Model Response}
\CaseProtocolMarker{<think>}

The user needs me to predict the possible molecular structure based on the given Raman spectrum; I will first analyse the Raman spectrum.

I observe the provided Raman spectrum, focusing first on the high wavenumber region where the most intense signals are located. There is a prominent cluster of sharp peaks between 2800 and 3100~cm$^{-1}$. Specifically, I note a very strong peak visually positioned around 2920--2930~cm$^{-1}$ (despite a numerical annotation of 2026.0~cm$^{-1}$, the dashed line and peak position clearly align with the 2800--3000~cm$^{-1}$ grid, suggesting the label may be a typo for a value like 2926.0~cm$^{-1}$). To the left of 3000~cm$^{-1}$, there are distinct peaks extending into the 3000--3100~cm$^{-1}$ range. To the right of 3000~cm$^{-1}$, there are multiple strong peaks around 2960, 2920, and 2870~cm$^{-1}$. Moving to the fingerprint region, I identify a medium intensity peak around 1460~cm$^{-1}$ and a cluster of peaks in the 1000--1100~cm$^{-1}$ range, along with a distinct peak around 700--800~cm$^{-1}$. The region around 1700~cm$^{-1}$, typical for carbonyls, appears relatively flat with only minor baseline fluctuations, lacking a strong, characteristic carbonyl peak.

Assigning these vibrations, the strong peaks below 3000~cm$^{-1}$ (2850--2960~cm$^{-1}$) are characteristic of C--H stretching modes in saturated sp$^3$ hybridized carbons, indicating the presence of alkyl chains. The peaks appearing above 3000~cm$^{-1}$ are significant; while typically associated with sp$^2$ hybridized carbons (alkenes/aromatics), in the context of saturated cyclic systems, they are a hallmark of the strained C--H bonds in cyclopropane rings, which have increased s-character. The peak at approximately 1460~cm$^{-1}$ corresponds to CH$_2$ scissoring/bending vibrations. The cluster around 1000--1100~cm$^{-1}$ is consistent with C--C skeletal vibrations and potentially the ring breathing mode of a cyclopropane ring.

Matching these observations to potential structures, the combination of standard alkane C--H stretches (below 3000~cm$^{-1}$) and high-frequency C--H stretches (above 3000~cm$^{-1}$) strongly points toward a molecule containing both alkane chains and cyclopropane rings. The absence of a strong peak in the 1700--1750~cm$^{-1}$ region argues against the presence of prominent ketone or aldehyde functional groups, as the C=O stretch is usually intense in Raman spectra. While weak signals might be obscured, the dominant spectral features are best explained by a hydrocarbon framework containing cyclopropane and alkane moieties. The evidence converges to confirm the presence of alkane and cyclopropane functionalities.

\CaseProtocolMarker{</think>}

\textbf{Spectral Analysis Summary.} The response assigns the 2850--2960~cm$^{-1}$ envelope to saturated C--H stretching, signals above 3000~cm$^{-1}$ to strained cyclopropane C--H bonds, the approximately 1460~cm$^{-1}$ peak to CH$_2$ scissoring, and the 1000--1100 and 700--800~cm$^{-1}$ regions to skeletal and ring-deformation modes. It concludes that the substance contains alkane and cyclopropane functional groups and proceeds to molecular-structure prediction.

\CaseProtocolMarker{<spectra_st>}
\end{CaseBlock}

\CaseNativeOutputHeading
\CaseImagePair{Predicted molecule}{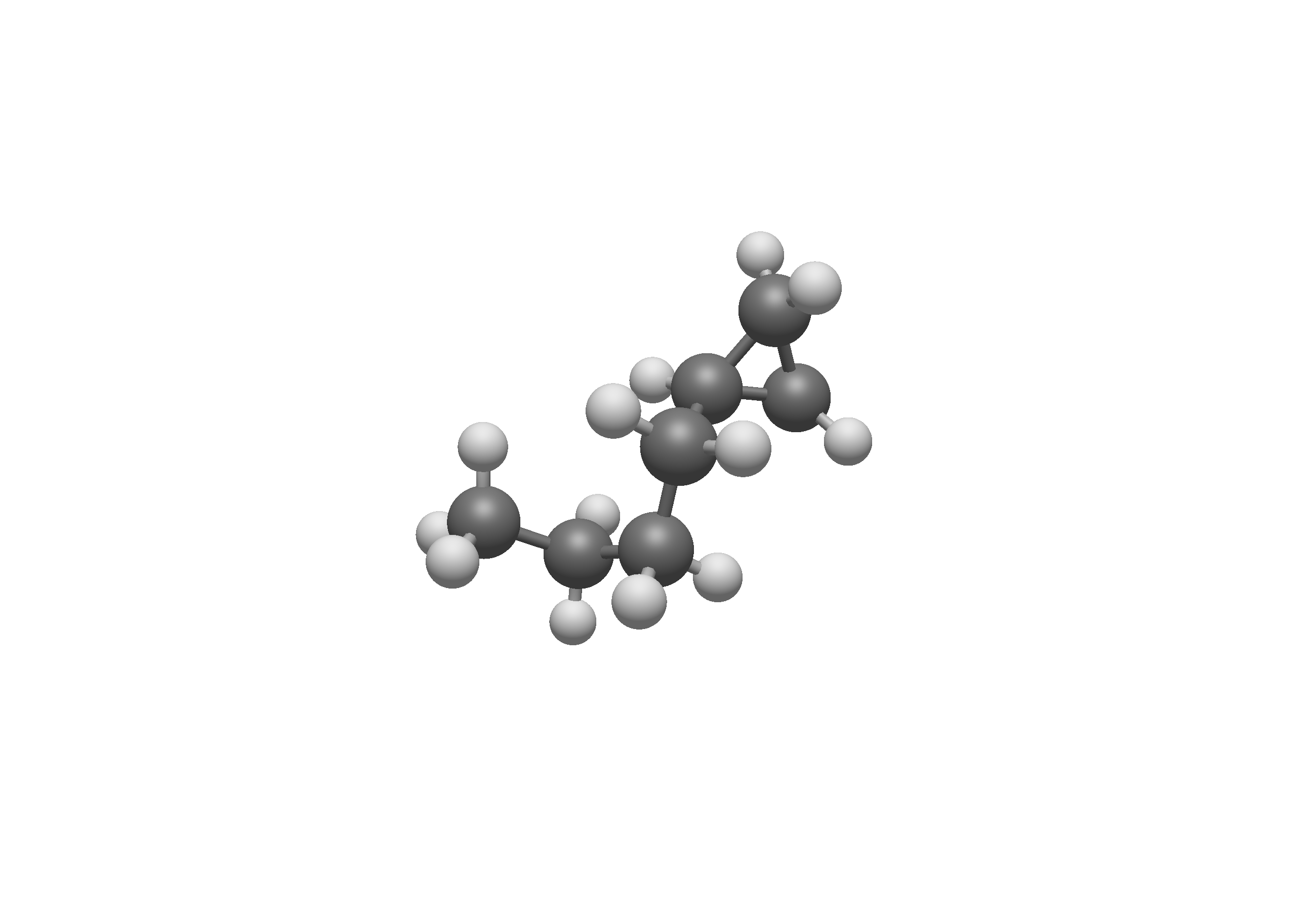}{Ground-truth molecule}{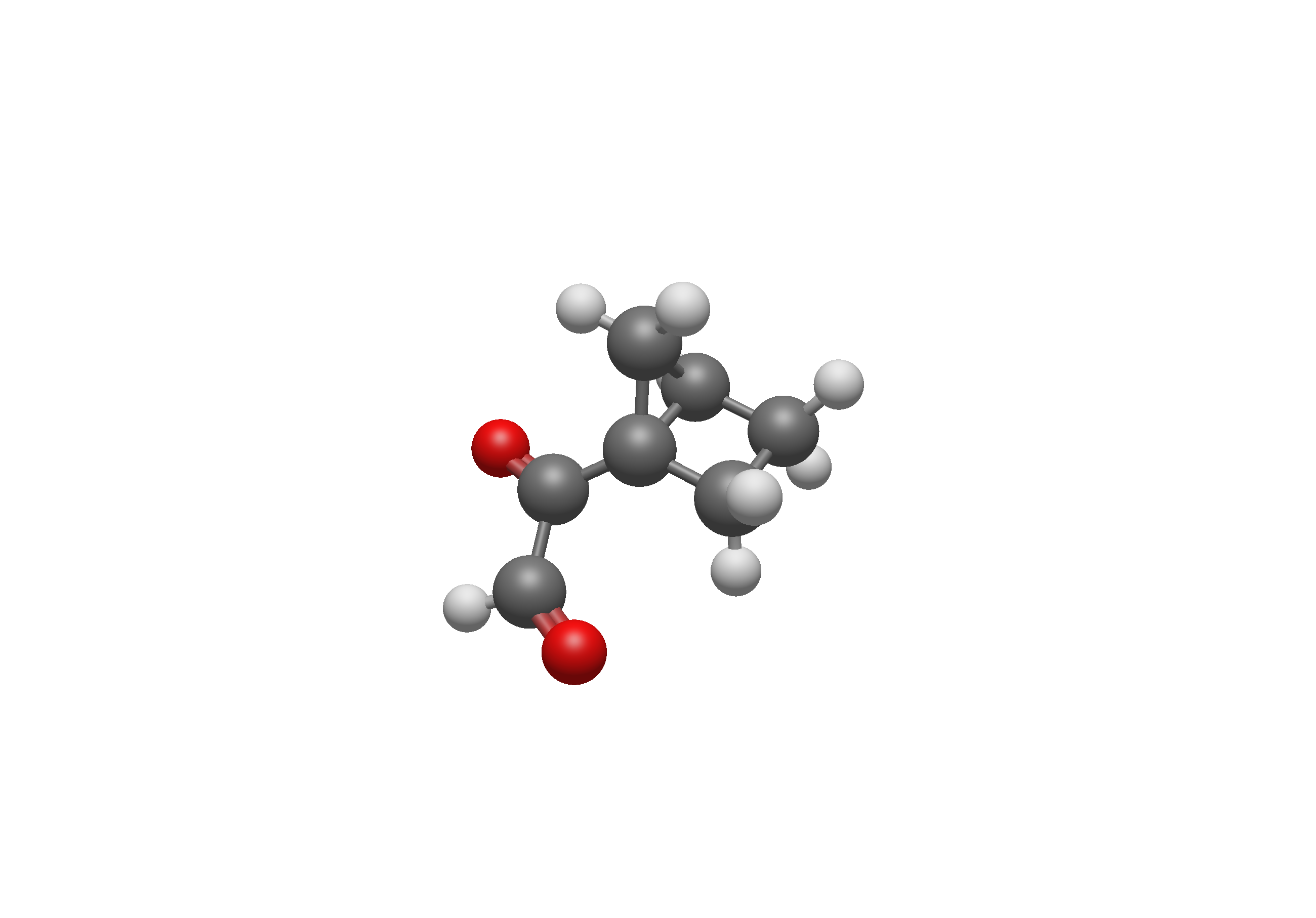}

\begin{CaseFailureNote}{Observed Failure}
The reasoning dismisses a carbonyl-containing structure and commits to a hydrocarbon with a cyclopropane motif. The predicted molecule therefore differs qualitatively from the oxygen-containing ground truth, showing how an incorrect peak assignment can propagate into the structural decoder.
\end{CaseFailureNote}

\color{black}\clearpage
\color{black}
\subsubsection{Protein Structure Prediction: Pre-Prediction Analysis}
\label{sec:case_failure_protein_structure}

\begin{CaseBlockImmediate}{User Input}
请分析下面这条蛋白质氨基酸序列，并给出结构预测前置研判（pre-prediction summary）：

\CaseInputFile{assets/case_study/failure_protein_sequence.txt}
\end{CaseBlockImmediate}

\begin{CaseBlock}{Model Response}
\CaseProtocolMarker{<think>}
I need to treat the user's message as an amino-acid sequence and organize structure-prediction-oriented evidence. The input sequence is \textbf{255 residues} long, which is a compact size typical for a single-domain enzyme. Based on the sequence composition and known family signatures, this protein belongs to the haloacid dehalogenase superfamily, specifically the subfamily IIB. This classification strongly suggests a soluble globular fold rather than a membrane-embedded structure, as there are no predicted transmembrane helices or signal peptides. The core architecture is expected to be an alpha-beta hydrolase fold that houses a catalytic core with a conserved aspartate nucleophile and a magnesium-binding site.

Functionally, this enzyme acts as a phosphatase that hydrolyzes phosphate monoesters. The mechanism involves a two-step process where a magnesium ion activates a water molecule to attack the phosphorus atom, forming a phospho-enzyme intermediate. This chemical requirement constrains the 3D structure to possess a specific active site pocket capable of coordinating divalent cations and positioning the substrate for nucleophilic attack. The protein is likely localized to the cytoplasm, consistent with its role in metabolic regulation where it processes small phosphorylated metabolites.

In terms of biological context, this protein participates in the regulation of the stringent response in bacteria. It likely interacts with enzymes involved in nucleotide metabolism and stress signaling pathways, such as those producing guanosine tetraphosphate. These interaction partners suggest the protein surface may contain specific interfaces for transient binding with metabolic enzymes. For structure prediction, these clues indicate a stable monomeric or homodimeric state with a well-defined catalytic cleft. The absence of disordered regions in the core domain further supports a high-confidence prediction of the tertiary structure based on homology to known phosphatase folds.
\CaseProtocolMarker{</think>}
\end{CaseBlock}

\begin{CaseFailureNote}{Observed Failure}
The input contains 254 residues, not 255. The response also assigns a specific HAD-IIB family, catalytic mechanism, cellular role, and stringent-response context without showing sequence-level evidence sufficient to support those claims. These unverified assumptions can bias the downstream structure prediction before coordinate generation begins.
\end{CaseFailureNote}

\color{black}\clearpage
\color{black}
\subsubsection{Scientific Image Generation}
\label{sec:case_failure_image_text}

The following composite example shows two prompts and their generated scientific infographics. Enlarged crops expose malformed Chinese and English glyphs, misspelled technical labels, and semantic substitutions in dense text regions.

\begin{figure}[H]
  \centering
  \includegraphics[width=\textwidth,height=0.82\textheight,keepaspectratio]{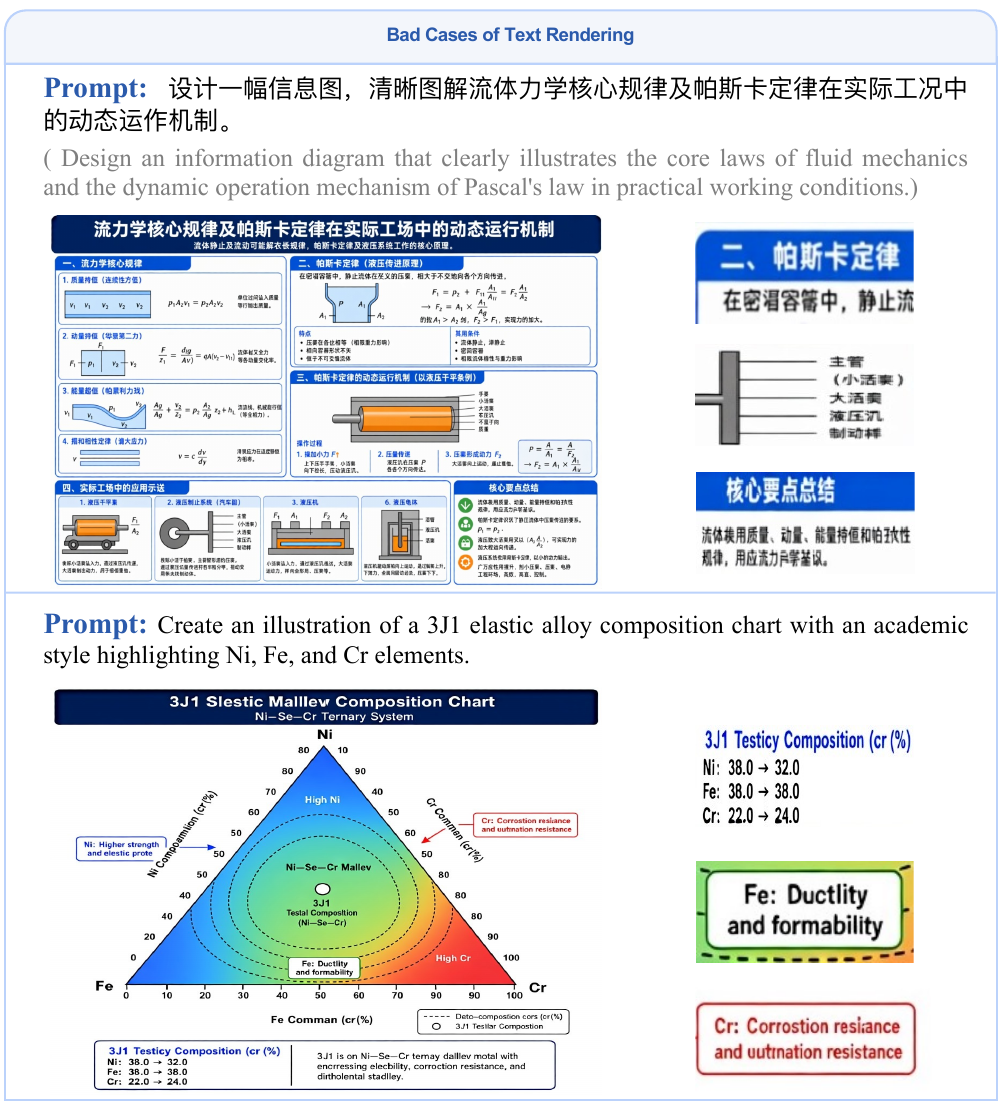}
  \caption{Representative text-rendering failures in generated scientific infographics about Pascal's law and a 3J1 alloy composition chart. Enlarged crops expose malformed Chinese and English glyphs, misspelled technical labels, and semantic substitutions.}
  \label{fig:case_failure_text_rendering}
\end{figure}

\begin{CaseFailureNote}{Observed Failure}
The global composition and scientific visual language are plausible, but locally rendered text is unreliable. Character substitutions, missing strokes, misspellings, and corrupted element labels reduce the figure's scientific usability even when the surrounding layout appears polished.
\end{CaseFailureNote}

\color{black}
\clearpage
\color{black}
\subsection{General Multi-Turn Dialogue Evaluation}
\label{sec:case_multiturn}

We provide a representative multi-turn dialogue transcript that interleaves general-purpose conversation and a scientific prediction request within a single long context. The associated analysis and implications for general conversational competence are discussed in Section~\ref{sec:discussion_multiturn_dialogue}.

\color{black}
\subsubsection{Protein-Site Prediction Across Multi-Turn and Single-Turn Contexts}

\CaseMiniHeader{Multi-Turn Conversation}
\CaseSpeaker{Turn 1 - User}
\begin{CaseBlockImmediate}{User Input}
你是谁？你能做什么？
\end{CaseBlockImmediate}

\begin{CaseBlockImmediate}{Model Response}
\CaseProtocolMarker{<think>}

嗯，用户问我是谁和我能做什么，这是一个简单的自我介绍问题。需要简洁明了地说明我的身份和核心功能。

我是中国科学院研发的磐石·科学基础大模型，主要面向科学领域提供服务。可以提到我的多模态能力、覆盖的学科范围以及具体能帮助用户完成的任务类型。

不需要展开技术细节或背景信息，重点突出实用功能即可。用清晰的结构分点说明会更易懂，但避免过于冗长。

\CaseProtocolMarker{</think>}

我是中国科学院研发的磐石·科学基础大模型（ScienceOne），专注于科学领域的智能支持。我可以帮助处理多种科学相关任务，包括：

\begin{itemize}[leftmargin=1.5em]
  \item \textbf{多模态科学数据理解}：支持文本、图像、波形、光谱、分子式、蛋白质序列等多种科学数据的分析与解读；
  \item \textbf{跨学科知识问答与推理}：覆盖数学、物理、化学、天文、地理、生物等基础学科，提供专业问题解答；
  \item \textbf{科学文献萃取与分析}：快速提取和融合科技文献中的关键信息，辅助科研洞察；
  \item \textbf{科学工具编排与规划}：自主调用和调度计算、仿真等科学工具，解决复杂问题。
\end{itemize}

如果有具体的科学问题或需求，欢迎随时提出！
\end{CaseBlockImmediate}

\CaseSpeaker{Turn 2 - User}
\begin{CaseBlockImmediate}{User Input}
请写一个教案：关于如何解二元一次方程组，字数要求 1000 字。
\end{CaseBlockImmediate}

\begin{CaseBlockImmediate}{Model Response}
\CaseProtocolMarker{<think>}

好的，用户让我写一个关于解二元一次方程组的教案，字数要求 1000 字左右。首先我需要明确用户的使用场景和身份。可能是中学数学老师准备课堂内容，或者师范生在做教学设计练习。

接下来要确定用户的真实需求。表面上是写教案，但深层可能希望教案结构清晰、有互动环节，并且能体现新课标理念，比如核心素养培养。同时要注意避免常见误区，比如学生容易混淆步骤或者符号错误。

然后回顾之前的对话历史，用户之前问过我的能力范围，现在转向具体教学内容，说明他信任我能提供专业支持。所以回复必须严谨，符合教育规范，不能出错。

开始构思教案框架。先设定教学目标，分知识、能力、情感三方面。重点放在代入消元法上，因为这是初学的关键方法。引入部分要用生活实例，比如购物问题，让学生感知方程组的实际意义。

接着考虑教学过程。导入要生动，可以用天平平衡的例子直观展示两个未知量的关系。讲解时强调“消元”的思想，对比算术解法突出方程的优势。练习部分要有梯度，从简单到复杂，最后加入实际应用题巩固。

还要注意易错点，比如去括号变号、移项变号这些细节。可以设计小组讨论环节，让学生互相检查，促进合作学习。板书设计要逻辑分明，步骤可视化。

时间分配上，导入 5 分钟，新授 20 分钟，练习 15 分钟，小结 5 分钟，作业布置 5 分钟。这样总时长 40 分钟，符合常规课时。

最后反思部分要提到如何评估学生掌握情况，以及后续延伸方向，比如二元一次方程组的应用。确保整个教案既有理论又有实践，满足不同层次学生的需求。

\CaseProtocolMarker{</think>}

\textbf{《解二元一次方程组》教案设计}

\textbf{适用年级}：初中七年级下学期。\quad
\textbf{课时安排}：1 课时（40 分钟）。

\textbf{设计理念}：以“问题驱动--探究建构--迁移应用”为主线，渗透数学建模思想，发展学生抽象思维与运算能力。

\textbf{一、教学目标}

\begin{enumerate}[leftmargin=1.8em]
  \item \textbf{知识与技能}：理解“消元”思想，掌握用代入消元法解二元一次方程组的基本步骤；能准确求解含整数系数的二元一次方程组，并检验解的正确性。
  \item \textbf{过程与方法}：通过生活情境引出方程组，体会方程组是刻画现实世界数量关系的有效工具；在观察、比较、归纳中总结解法规律，提升逻辑推理能力。
  \item \textbf{情感态度与价值观}：感受数学与生活的联系，激发学习兴趣；培养严谨的学习习惯与合作交流意识。
\end{enumerate}

\textbf{二、教学重难点}

重点是代入消元法的步骤及原理；难点是灵活选择变量进行消元，避免符号错误。

\textbf{三、教学过程}

\textbf{（一）情境导入（5 分钟）}

教师出示问题：“小明买 2 个笔记本和 3 支笔共花 18 元，买 1 个笔记本和 2 支笔共花 10 元。每本笔记本和每支笔各多少元？”引导学生尝试列算术式，再设未知数列出方程组：

\[
\begin{cases}
2x+3y=18,\\
x+2y=10.
\end{cases}
\]

提问“如何找到 $x,y$ 的值？”，引发思考，需要将“二元”转化为“一元”。

\textbf{（二）探究新知（15 分钟）}

先回顾一元一次方程解法，强调“化归”思想，并讨论能否把其中一个未知数当作已知数，从而得到另一个未知数的表达式。随后示范求解

\[
\begin{cases}
x+y=5,\\
2x-y=1.
\end{cases}
\]

步骤为：由第一个方程得 $x=5-y$；代入第二个方程，得 $2(5-y)-y=1$；解得 $y=3$；回代得 $x=2$；最后将 $x=2,y=3$ 代入原方程组检验。师生共同讨论为什么优先选择系数为 1 的方程变形，并总结关键词“变形 $\rightarrow$ 代入 $\rightarrow$ 解一元 $\rightarrow$ 回代 $\rightarrow$ 检验”。

\textbf{（三）巩固训练（12 分钟）}

\begin{enumerate}[leftmargin=1.8em]
  \item 基础题：解方程组 $3x-y=7$，$x+2y=9$，并判断先变形哪个方程更简便。
  \item 拓展题：若一般方程组中的某个未知数系数为 0，讨论如何消元并举例说明。
  \item 纠错环节：展示去括号漏负号、回代抄错数值等常见错误，由学生集体诊断。
\end{enumerate}

\textbf{（四）小结与反思（5 分钟）}

学生自述今天学会的内容和解题关键；教师提炼核心思想“消元”，强调优先变形系数为 1 的方程以减少计算量，并说明检验是防止低级失误的必要步骤。最后提出延伸问题：若两个方程中某个未知数的系数互为相反数，是否存在更简便的方法？

\textbf{（五）作业布置（3 分钟）}

必做教材基础题；选做一道用方程组解决的实际问题并给出完整解答；挑战题讨论方程组无解或有无数解时应满足的条件。

\textbf{四、板书设计}

板书依次呈现：课题；“消元 $\rightarrow$ 化二元为一元”的目标；变形、代入、解一元、回代、检验五个步骤；示例方程组；去括号变号和回代勿漏等易错提醒。

\textbf{五、教学评价}

过程性评价关注小组讨论参与度和板演准确性；结果性评价通过练习题检测步骤掌握情况，重点检查是否遗漏检验；发展性评价鼓励学生提出更简洁的解法。

\textbf{六、教学资源}

使用多媒体课件动态演示代入过程，准备典型错题卡片，并建立购物、工程、几何周长等生活情境素材库。

\textbf{设计说明}：本教案紧扣“消元”这一核心思想，通过真实问题引发认知冲突，在操作体验中内化算法，同时注重错误预防与思维进阶，助力学生形成系统的方程组解题策略。
\end{CaseBlockImmediate}

\CaseSpeaker{Turn 3 - User}
\begin{CaseBlockImmediate}{User Input}
Given the following protein sequence, predict the residues that form protein--protein interaction (PPI) binding sites.

\CaseInputFile{assets/case_study/multiturn_ppi_sequence.txt}
\end{CaseBlockImmediate}

\begin{CaseBlockImmediate}{Model Response}
\CaseProtocolMarker{<think>}

The input is a protein sequence of 126 amino acids. To identify potential protein--protein interaction binding sites, I need to analyze the sequence for features commonly associated with PPI interfaces, such as surface-exposed hydrophobic patches, charged residue clusters, and known structural motifs.

First, I examine the overall composition. The sequence contains several cysteine residues at positions 27, 40, 58, 65, 72, 87, 97, and 111. These are likely involved in disulfide bonds or metal coordination rather than direct PPI contacts. The N-terminal region (residues 1--20) is rich in lysine, glutamic acid, threonine, and alanine. This segment appears highly flexible and charged, which could serve as an intrinsically disordered region capable of mediating transient interactions. Specifically, the cluster K2, E3, K7, E9, R10 forms a positively and negatively charged patch that might engage in electrostatic interactions with partner proteins.

Moving to the central region, there is a stretch from approximately residue 20 to 60 containing multiple polar and aromatic residues. Notably, Y25, C27, Q29, M30, M31, K32, S33, and R34 create a mix of hydrophobic and charged residues. Methionines are often found at PPI hot spots due to their conformational flexibility and ability to pack against hydrophobic surfaces. The presence of consecutive methionines M30--M31 suggests a potential hydrophobic core exposed on the surface. Additionally, the motif SRNLT (residues 34--38) includes arginine, a common hotspot residue in PPIs because of its guanidinium group capable of forming hydrogen bonds and salt bridges.

Further downstream, around residues 70--90, another cluster emerges: Y71, S72, T74, M75, S76, I77, D78, C79, R80, E81, and T82. Here, Y71 and M75 provide aromatic and hydrophobic character, while R80 and E81 offer charged complementarity. The pattern RETG resembles a short linear motif that can be recognized by SH2 or other domain-binding partners.

The final region near the C-terminus contains a tyrosine-rich and proline-containing segment followed by hydrophobic and charged residues. Within this region, the tyrosines are prime candidates for phosphorylation-dependent interactions, while histidines may coordinate metals or participate in pH-sensitive binding.

Integrating these observations, the most probable PPI binding-site residues are those that combine surface accessibility, charge, and hydrophobicity without being buried in predicted secondary structures. Based on sequence-conservation patterns and physicochemical properties, I select the following residues as high-confidence predictions: L11, T14, S16, S18, N24, Y25, C27, Q29, M30, M31, K32, S33, R34, N35, L36, T38, K40, V44, H46, E47, L48, V52, Q54, V56, C58, Q60, K62, N64, V66, C68, N70, Y71, S72, Y73, S74, M75, I77, D78, C79, R80, E81, T82, K86, Y87, P88, N89, C90, A91, Y92, K93, T94, Q97, N101, K106, H107, I108, I109, V110, C111, E112, G113, N114, P115, Y116, V117, P118, V119, H120, F121, D122, A123, S124, V125. These residues span multiple regions and include key hotspot types that are statistically enriched at PPI interfaces.

\CaseProtocolMarker{</think>}

Based on the analysis of the given protein sequence, the following residues are predicted to form protein--protein interaction binding sites:

\textbf{Predicted PPI Binding Site Residues:}

L11, T14, S16, S18, N24, Y25, C27, Q29, M30, M31, K32, S33, R34, N35, L36, T38, K40, V44, H46, E47, L48, V52, Q54, V56, C58, Q60, K62, N64, V66, C68, N70, Y71, S72, Y73, S74, M75, I77, D78, C79, R80, E81, T82, K86, Y87, P88, N89, C90, A91, Y92, K93, T94, Q97, N101, K106, H107, I108, I109, V110, C111, E112, G113, N114, P115, Y116, V117, P118, V119, H120, F121, D122, A123, S124, V125.

These residues were selected based on their location in solvent-accessible loops, enrichment in aromatic, charged, and hydrophobic amino acids that are statistically favored at PPI interfaces, and the presence of short linear motifs such as RETG and SRNL that often mediate specific binding events. The actual residue index of the site is: 

\CaseProtocolMarker{<prot_cla>}
\end{CaseBlockImmediate}

\CaseNativeOutputHeading

\CaseResidues{[11, 40, 41, 42, 43, 49, 65, 66, 67, 69, 71, 72, 83, 85, 107, 109, 111, 116, 117, 118, 119, 120, 121, 122, 123, 124]}

\clearpage
\CaseMiniHeader{Single-Turn Control}
\begin{CaseBlockImmediate}{User Input}
Given the following protein sequence, predict the residues that form protein--protein interaction (PPI) binding sites.

\CaseInputFile{assets/case_study/multiturn_ppi_sequence.txt}
\end{CaseBlockImmediate}

\begin{CaseBlockImmediate}{Model Response}
\CaseProtocolMarker{<think>}

I need to predict the protein--protein interaction binding sites for the given protein sequence. Let me analyze this sequence carefully.

The sequence contains multiple cysteines, which suggests it could be a cytokine or growth factor with disulfide bonds. I am mapping the cysteine positions throughout the sequence to understand the structural constraints, and they appear at intervals characteristic of proteins with stabilizing disulfide bridges. Looking at the cysteine-spacing pattern and overall structure, this appears to be an interleukin-1 family member based on conserved cysteine pairs and the characteristic fold. The N-terminal region looks particularly characteristic of these inflammatory cytokines.

I identify potential PPI binding-site residues by considering charged amino acids such as lysines and arginines, together with aromatic residues such as phenylalanine and tyrosine that typically participate in protein interactions. These clusters of polar and hydrophobic residues are likely candidates for receptor binding. I refine the prediction toward the most probable interface by considering the N-terminal helix, loop regions, and C-terminal sections that show characteristics of protein--protein interaction sites.

The N-terminal region likely forms an alpha helix, while the middle section contains loops and turns. Candidate residues in the first half include Y24, C25, Q27, M28, M29, K31, R33, L34, K37, R39, C40, K41, F46, H48, E49, S50, L51, A52, D53, V54, and Q55. The second half contains additional hotspots including V59, C60, Q62, K63, N64, V65, C66, K68, Q70, N71, C72, Y74, S75, Y76, T77, M78, S79, T80, D81, C82, R83, E84, T85, G86, S87, S88, K89, and Y90, followed by further C-terminal charged and aromatic residues.

\CaseProtocolMarker{</think>}

\textbf{PPI Binding Site Prediction}

\textbf{N-terminal helix:} K1, E2, K7, F8, E9, R10, Q11, H12, M13. This amphipathic segment combines a mixed-charge surface with a hydrophobic face.

\textbf{Loop 1:} Y24, Q27, M28, M29, K31, R33. This exposed loop contains aromatic and charged anchors.

\textbf{Charged patch:} K37, R39, K41. These residues form a candidate electrostatic hotspot cluster.

\textbf{Central beta region:} F46, H48, E49, L51, D53, Q55. This region combines hydrophobic and polar interface residues.

\textbf{Disulfide-flanked loop:} Q62, K63, K68, Q70, Y74, Y76. These residues are predicted to be surface exposed between cysteine pairs.

\textbf{Core loop:} R83, E84, K89, Y90, K95. This region has high electrostatic potential.

\textbf{C-terminal region:} K101, H102, E108, Y112, H116, F117, D118. These residues are assigned to a solvent-exposed terminal interface.

The primary interface is predicted to combine the N-terminal helix (K1--M13) and central loop (F46--Q55), with a secondary C-terminal charged patch (K101--D118). Candidate hotspot residues include R10, K31, R33, K37, R39, F46, R83, K101, and H102. The proposed binding mode is a bidentate contact through two spatially separated surfaces flanking a rigid disulfide-stabilized core. Experimental validation by mutagenesis or co-crystallography is recommended. The actual residue index of the site is:

\CaseProtocolMarker{<prot_cla>}
\end{CaseBlockImmediate}

\CaseNativeOutputHeading

\CaseResidues{[11, 40, 41, 42, 43, 49, 65, 66, 67, 69, 71, 72, 83, 85, 107, 109, 111, 116, 117, 118, 119, 120, 121, 122, 123, 124]}

\FloatBarrier

\end{CJK*}
\end{document}